\documentclass{article} 
\usepackage[final]{neurips_2024}

\usepackage{amsmath,amsfonts,bm}

\def\eqref#1{equation~\ref{#1}}

\def\1{\bm{1}}

\DeclareMathAlphabet{\mathsfit}{\encodingdefault}{\sfdefault}{m}{sl}
\SetMathAlphabet{\mathsfit}{bold}{\encodingdefault}{\sfdefault}{bx}{n}

\usepackage{hyperref}
\usepackage{url}
\usepackage{wrapfig}
\usepackage{mathtools}
\usepackage{graphicx}
\usepackage{authblk}
\usepackage{booktabs} 
\usepackage{pifont}
\usepackage{multirow}
\usepackage{adjustbox}
\usepackage[table]{xcolor}
\usepackage{amsthm}
\usepackage{subfigure}
\usepackage{enumitem}
\usepackage{bm}
\usepackage{siunitx}
\usepackage{rotating}
\usepackage{tablefootnote}
\usepackage[dvipsnames]{xcolor}
\usepackage{etoc}  
\usepackage{hyperref} 
\usepackage[most]{tcolorbox}
\usepackage[capitalize]{cleveref}
\usepackage[ruled,vlined,linesnumbered]{algorithm2e}   
\usepackage{todonotes}
\DontPrintSemicolon
\usepackage{graphicx} 
\usepackage{float}
\DontPrintSemicolon
\makeatletter
\newcommand{\AlgCompress}{%
  \SetAlFnt{\footnotesize}%
  \SetAlgoNlRelativeSize{-1}%
  \setlength{\algomargin}{0.3em}%
  \SetInd{0.5em}{0.8em}%
  \setlength{\interspacetitleruled}{0.2ex}%
  \setlength{\algotitleheightrule}{0.4pt}%
  \LinesNumberedHidden%
}
\makeatother
\definecolor{mybackblue}{RGB}{240,248,255}
\definecolor{myframeblue}{RGB}{100,149,237}

\newtcolorbox{keyfindingbox}[1][]{%
  enhanced,
  breakable,
  colback=mybackblue,
  colframe=myframeblue,
  coltitle=white,
  fonttitle=\bfseries\small,
  attach boxed title to top left={yshift=-2mm, xshift=2mm},
  boxed title style={
    colback=myframeblue,
    sharp corners=south,
    boxrule=0pt,
    arc=3pt,
    left=6pt,
    right=6pt,
    top=2pt,
    bottom=2pt,
  },
  title=#1,
  boxrule=0.4pt,
  arc=2mm,
  left=8pt,
  right=8pt,
  top=8pt,
  bottom=8pt
}

\SetKw{Continue}{continue}
\def\pz{{\phantom{0}}}
\definecolor{lightblue}{HTML}{3498db}
\definecolor{lightred}{HTML}{E26354}
\definecolor{lightgray}{HTML}{E4E4E4}
\definecolor{myframeblue}{HTML}{3498db}   
\definecolor{mybackblue}{HTML}{D6EAF8} 
\definecolor{red}{HTML}{e74c3c}
\newtheorem{proposition}{Proposition}

\newtheorem{lemma}{Lemma}

\newcommand{\cmark}{\ding{51}}
\newcommand{\xmark}{\ding{55}}
\newlength{\SavedARW}
\newcommand{\HideVlines}{\noalign{\global\setlength{\SavedARW}{\arrayrulewidth}\global\setlength{\arrayrulewidth}{0pt}}}
\newcommand{\ShowVlines}{\noalign{\global\setlength{\arrayrulewidth}{\SavedARW}}}

\DeclarePairedDelimiter{\paren}{\lparen}{\rparen}
\title{Turning Internal Gap into Self-Improvement: Promoting the Generation-Understanding Unification in MLLMs}

\author{%
Yujin Han$^{1,8}$\thanks{Email to: Yujin Han (yujinhan@connect.hku.hk).},
Hao Chen$^{2}$,
Andi Han$^{3,4}$,
Zhiheng Wang$^{5,6}$,
Xinyu Liu$^{7}$\\
Yingya Zhang$^{8}$,
Shiwei Zhang$^{8}$\textsuperscript{\textdagger},
Difan Zou$^{1}$\thanks{Correspondence to: Shiwei Zhang (zhangjin.zsw@alibaba-inc.com) and Difan Zou (dzou@cs.hku.hk)}
}
\affil{\small{$^{1}$The University of Hong Kong, $^{2}$Carnegie Mellon University, $^{3}$University of Sydney, $^{4}$RIKEN AIP, $^{5}$Shanghai Artificial Intelligence Laboratory, $^{6}$Shanghai Jiao Tong University,$^{7}$Hong Kong University of Science and Technology, $^{8}$Alibaba Group}}

\begin{document}
\etocdepthtag.toc{mtchapter}
\etocsettagdepth{mtchapter}{none}
\etocsettagdepth{mtappendix}{none}

\maketitle

\begin{abstract}

Although unified MLLMs aim to unify generation and understanding, they are considered to exhibit an internal gap, with understanding outperforming generation. Through large‑scale evaluation across multiple MLLMs and tasks, we confirm the widespread non‑unification of MLLMs, and demonstrate that it indeed stems from weak generation rather than misunderstanding. This finding motivates us to propose a simple yet effective internal gap-based self-improvement framework, which mitigates internal gaps by leveraging stronger understanding to guide weaker generation without relying on any external signals. We validate this strategy through comprehensive experiments: scoring generations with understanding to construct image data for post-training (e.g., SFT and DPO) significantly improves generation while promoting unification. Furthermore, we empirically discover a co-improvement effect of such self-improvement, a phenomenon well known in pre-training but underexplored in post-training. Specifically, as generation improves, understanding becomes more effective at detecting false positives that were previously misclassified as prompt‑aligned. To explain this effect, we extend learning dynamic theory to the MLLM setting, showing that the shared empirical neural tangent kernel between generation and understanding encourages aligned learning dynamics, thereby driving co-improvement. This interplay between generation and understanding further motivates a curriculum learning approach for stronger self‑improvement: progressively enhanced understanding and generation revisit samples underutilized by pre‑trained MLLMs, dynamically expanding post‑training data and leading to improved performance and unification.

\end{abstract}


\section{Introduction}

\begin{figure}
  \centering
  \includegraphics[width=0.75\textwidth, height=0.09\textheight]{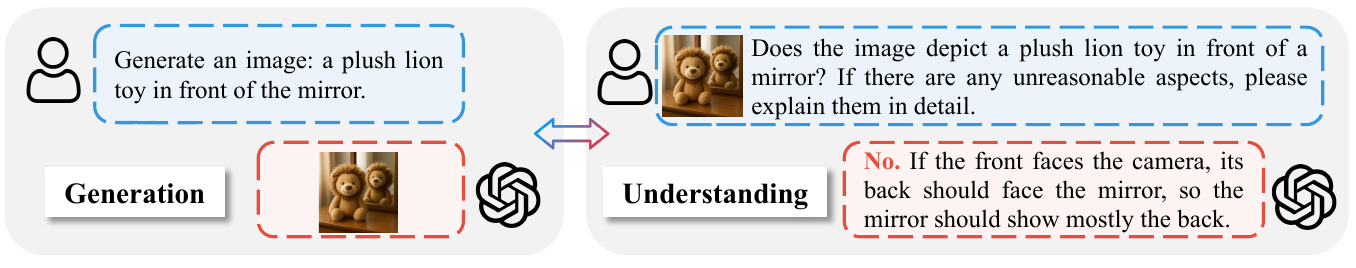}
\vspace{-0.15in}
  \caption{Illustration of MLLMs' internal gap. We examine a challenging case \citep{han2025diffusionmodelslearnhidden} involving implicit physical principles using ChatGPT o3 \citep{openai2024gpt4technicalreport} and find: images produced by generation branch are identified as incorrect by understanding branch, showing non-unification.}
\vspace{-0.1in}
  \label{fig: self-con}
\end{figure}

Unified Multimodal Large Language Models (MLLMs) have attracted growing attention for their capability to conduct both generation and understanding \citep{xie2024showosingletransformerunify, wu2024janusdecouplingvisualencoding, wang2024emu3, chameleonteam2025chameleonmixedmodalearlyfusionfoundation, zhou2024transfusionpredicttokendiffuse, chen2025blip3ofamilyfullyopen}. 
However, an emerging consensus is that, despite being designed to unify both generation and understanding, they are not truly unified in performance \citep{yang2025hermesflow,mao2025unirlselfimprovingunifiedmultimodal,hong2025reinforcing,yan2025can}, where understanding typically outperforms generation \citep{yang2025hermesflow}. 
For example, \cref{fig: self-con} shows, an MLLM’s generation may be judged as prompt-misaligned by its own understanding branch, revealing an internal generation–understanding gap. A natural question arises: 

\textit{Can the internal gap in MLLMs be leveraged as a free bonus, with the stronger branch guiding the weaker one to improve the model’s performance and mitigate non-unification?}

Prior works have discussed the internal gap in unified MLLMs, but their mitigation methods often rely on external reward models \citep{yang2025hermesflow} or additional supervised datasets \citep{mao2025unirlselfimprovingunifiedmultimodal}, or focus solely on improving a single task, e.g., generation \citep{jiang2025t2ir1reinforcingimagegeneration,yan2025can,xie2025reconstruction}, without emphasizing generation–understanding alignment. In this paper, we explore the potential of mitigating MLLMs' non-unification without any external signals, and propose a simple yet effective internal gap-based self-improvement framework. We further provide a detailed analysis of the dynamic interplay between generation and understanding during self-improvement, offering a strong complement to existing studies.

We begin by validating the generation–understanding gap across multiple MLLMs and tasks. We first introduce an internal metric, \textit{non-unification score}, defined as the proportion of cases where the understanding branch judges
the generation as prompt-misaligned. Unlike previous unification metrics that rely on an external estimator \citep{yang2025hermesflow,mao2025unirlselfimprovingunifiedmultimodal}, our metric directly quantifies the internal consistency between two branches, avoiding biases from external assessment. Comprehensive evaluation on six unified MLLMs and tasks of three difficulty levels shows that non-unification is pervasive, with  non-unification score reaching up to 60\%. Further quantitative analysis attributes most misalignments (60–100\%) to \textit{weak generation} rather than misunderstanding, consistent with prior findings on single tasks \citep{yang2025hermesflow} and single models \citep{mao2025unirlselfimprovingunifiedmultimodal}.

After confirming widespread non-unification and stronger understanding, we propose an \textbf{internal gap-based self-improvement framework} that aligns MLLMs by leveraging stronger understanding to guide the weaker generation. We validate its effectiveness on mainstream MLLMs such as Janus-Pro-7B \citep{chen2025janusprounifiedmultimodalunderstanding}: using the understanding branch to score generations and construct post-training data for generation, standard pipelines, e.g., SFT~\citep{brown2020languagemodelsfewshotlearners,radford2021learningtransferablevisualmodels} and DPO~\citep{rafailov2024directpreferenceoptimizationlanguage}, significantly boost generation (up to +20\% on T2I-CompBench++ \citep{huang2023t2i}) and reduce the internal gap (non-unification score by as much as –16\%), surpassing even baselines with multiple external reward models such as T2I-R1 \citep{jiang2025t2ir1reinforcingimagegeneration}.

Furthermore, we empirically observe a \textit{co-improvement} effect: the generation-targeted self-improvement method also enhances understanding. Specifically, self-improved MLLMs better detect false positives, i.e., samples previously misidentified as prompt-aligned. While co-improvement is well-known in pre-training \citep{tong2024metamorphmultimodalunderstandinggeneration,wu2025liquidlanguagemodelsscalable,deng2025bagel,zhang2025unifiedvisionlanguagemodelsnecessary,wu2025harmonizing}, it remains underexplored in post-training \citep{yang2025hermesflow,mao2025unirlselfimprovingunifiedmultimodal}. To explain it, we extend learning dynamic theory \citep{ren2025learningdynamicsllmfinetuning} to multimodal settings and formalize joint evolution of generation and understanding during self-improvement. Our theory reveals a shared empirical neural tangent kernel (eNTK) facilitates consistent learning dynamics across generation and understanding. Consequently, aligned dynamics reduce misaligned generations and enhance misalignment detection, thus leading to co-improvement effect observed.

Finally, motivated by the co-improvement effect, we further demonstrate that \textit{curriculum learning}~\citep{elman1993learning,bengio2009curriculum} can be incorporated into self-improvement by gradually introducing harder samples that were initially excluded due to limited capabilities in generation or understanding. Experiments show that curriculum learning enables self-improvement to dynamically expand post-training data, further enhancing both the performance and unification of MLLMs.

Through a systematic exploration of MLLMs' internal gap, our contributions are as follows:

\begin{itemize}[leftmargin=0.2in]
\item  We first introduce the non-unification score, an internal consistency metric to measure MLLMs' internal gap. Extensive evaluations across diverse models and tasks confirm pervasive non-unification phenomenon, which is primarily caused by weak generation.
\item Motivated by non-unification in MLLMs, we then propose a simple yet effective internal gap-based self-improvement framework, which leverages stronger understanding capability to guide the weaker generation. Extensive experiments show the proposed self-improvement significantly boosts both generation and unification without external signals.
\item In self-improvement, we empirically identify a co-improvement effect, where understanding better detects prompt-misaligned generations. Extending learning dynamics to MLLMs, we attribute this effect to shared eNTK between generation and understanding.
\item Finally, co-improvement effect inspires a curriculum-based self-improvement strategy: progressively strengthen understanding and generation enable reusing underutilized samples, thereby expanding post-training data and boosting both performance and unification.
\end{itemize}

\section{Related Work}
\label{sec:Related Work}

\textbf{Non-unification of MLLMs.} 
There are works showing internal gap of MLLMs, typically with understanding outperforming generation \citep{yang2025hermesflow,mao2025unirlselfimprovingunifiedmultimodal,hong2025reinforcing,yan2025can,yang2025hermesflow}. However, existing studies \textit{lack systematic quantification} of such gap across multiple MLLMs and tasks, with conclusions often confined to single models \citep{mao2025unirlselfimprovingunifiedmultimodal} or single tasks \citep{yang2025hermesflow}. Additionally, their measurements of internal gap rely on external models, e.g., ChatGPT \citep{yang2025hermesflow,mao2025unirlselfimprovingunifiedmultimodal} \textit{instead of measuring internal consistency}, which potentially makes biased estimation by external evaluators. Therefore, we first focus on introducing non-unification metric and performing large-scale verification.

\textbf{Mitigating Non-unification of MLLMs.} Several studies attempt to mitigate internal gap within MLLMs, but they rely on external models \citep{jiang2025t2ir1reinforcingimagegeneration,yang2025hermesflow} or additional data \citep{mao2025unirlselfimprovingunifiedmultimodal}. For example, Hermesflow \citep{yang2025hermesflow} leverages external Bert \citep{devlin2019bert} for understanding, self-critique and VQA \citep{antol2015vqa} models for generation, to improve both branches. Other works \citep{jiang2025t2ir1reinforcingimagegeneration,duan2025gotr1unleashingreasoningcapability} enhance weaker generation by introducing multiple external reward models, e.g., BLIP \citep{li2022blipbootstrappinglanguageimagepretraining} and HPMs \citep{wu2023humanpreferencescorev2,xu2023imagerewardlearningevaluatinghuman}. In contrast, we focus on mitigating internal gap purely through self-improvement \textit{without any external signals}. Importantly, self-improvement does not conflict with existing approaches: once achieved, external signals can be incorporated to further boost MLLMs.

\textbf{Co-improvement of MLLMs.} Co-improvement in unified MLLMs often refers to one branch improving when the other is improved, such as understanding gains from adding more generation data \citep{tong2024metamorphmultimodalunderstandinggeneration,wu2025liquidlanguagemodelsscalable}. This phenomenon has been widely observed in pre-training \citep{tong2024metamorphmultimodalunderstandinggeneration,wu2025liquidlanguagemodelsscalable,deng2025bagel,zhang2025unifiedvisionlanguagemodelsnecessary,wu2025harmonizing}, yet it has \textit{not been sufficiently highlighted} or \textit{thoroughly analyzed} in post-training \citep{yang2025hermesflow,mao2025unirlselfimprovingunifiedmultimodal,hong2025reinforcing}. Our work provides a learning-dynamics perspective on it, offering insights into interplay between understanding and generation in unified MLLMs.

\begin{figure}
  \centering
  \includegraphics[width=1\textwidth, height=0.16\textheight]{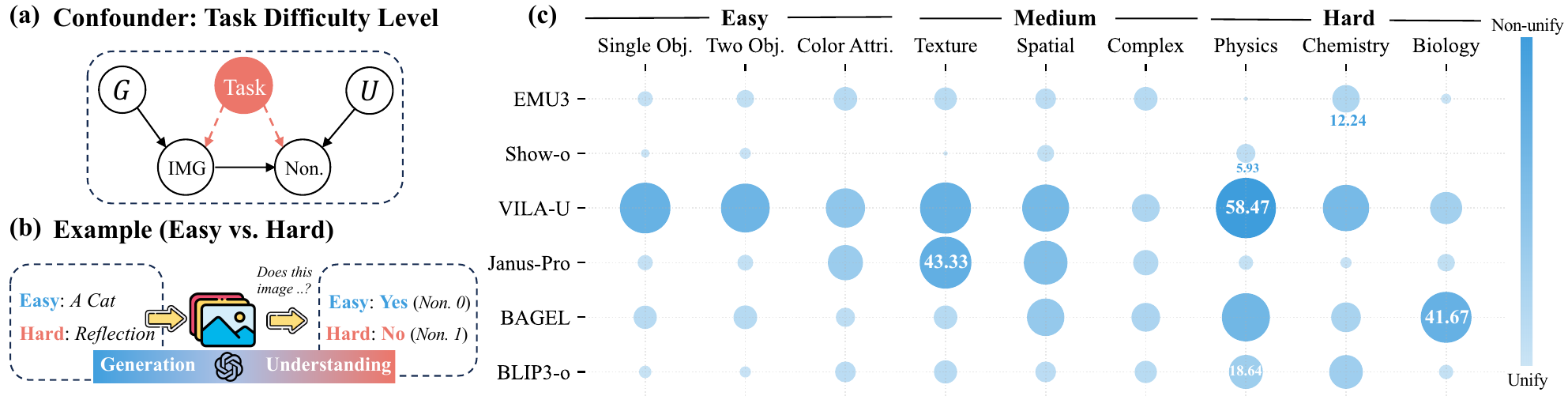}
\vspace{-0.3in}
  \caption{Verification of internal gaps. (a) and (b) identify task difficulty as a confounder in measuring non-unification score (\textit{Non.}): easy tasks may underestimate the gap, while hard tasks risk overestimation. Stratifying by task difficulty (Easy–Medium–Hard) yields a more reliable estimation. (c) Evaluation of six MLLMs across  three difficulty levels shows unified MLLMs remain non-unified, with non-unification scores approaching 60\%. More details are provided in \Cref{app:Details on Self-contradiction}.}
\vspace{-0.1in}
  \label{fig:non-unification}
\end{figure}

\section{Phenomenon Verification: The Non-unification in MLLMs}
\label{sec:Phenomenon Validation: The Non-unification in MLLMs}
While prior work suggests internal imbalances in MLLMs, this claim remains unverified through evaluation across diverse models and tasks (see \Cref{sec:Related Work}). We therefore take the large-scale empirical verification of non-unification as the starting point of our study.

We first propose a self-consistency metric to quantify the generation-understanding gap, termed the \textit{non-unification score}. Specifically, consider an MLLM $\pi_{\theta}$, a prompt $\mathbf{y}$ and the generated image $\mathbf{x} = \pi^{\text{gen}}_{\theta}(\mathbf{y})$. We form an image–question pair $(\mathbf{x}, q(\mathbf{y}))$, where $q(\mathbf{y}) \coloneqq \text{``Does this image describe } \mathbf{y} \text{?''}$. This pair is processed by the understanding branch $\pi^{\text{und}}_{\theta}(\cdot)$, yielding a binary decision: 1 if $\mathbf{x}$ is aligned with $\mathbf{y}$, and 0 otherwise. The \textit{non-unification score} is the proportion of decisions equal to 0,
\begin{equation}
\label{eq:nonunified_score}
\text{Non-unification score} := \mathbb{E}_{(\mathbf{x},\mathbf{y})} \mathbb{I} \left[ \pi^{\text{und}}_{\theta} \left( \mathbf{x},\ q(\mathbf{y}) \right) = 0 \right].
\end{equation}
Intuitively, unified MLLMs should have a near‑zero non‑unification score: generation renders the prompt as an image and understanding verifies the image matches the prompt.

\begin{wrapfigure}{r}{0.45\textwidth}
\vspace{-0.25in}
\begin{center}
    \includegraphics[width=0.43\textwidth]{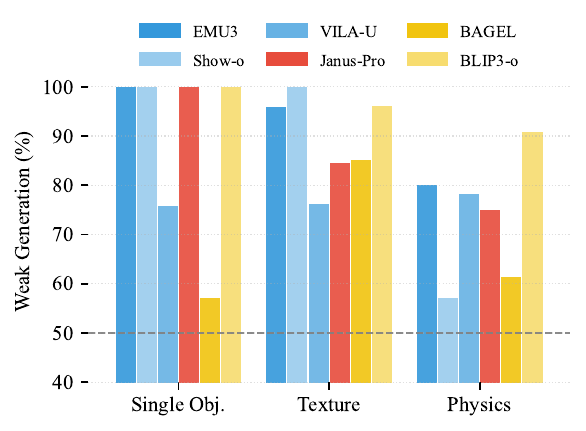}
\end{center}
\vspace{-0.3in}
\caption{Weak-generation (Qwen-checked) above 50\% (even 100\%) indicate internal gap mainly stems from weak generation. \cref{fig:human-check} provides human check, showing conclusions consistent with Qwen. \Cref{app:Full Results on Self-contradiction} reports more Weak Generation results.} 
\vspace{-0.15in}
\label{fig:nonunified_score_janus}  
\end{wrapfigure}
We evaluate multiple MLLMs \citep{wang2024emu3,xie2024showosingletransformerunify,wu2024vila,chen2025janusprounifiedmultimodalunderstanding,deng2025bagel,chen2025blip3ofamilyfullyopen}, across tasks of varying difficulty. 
We emphasize that task difficulty is a confounder affecting both generation and understanding, thereby biasing the non-unification score (see \cref{fig:non-unification}(a)). For example, as shown in \cref{fig:non-unification}(b), a simple prompt like \textit{generate a cat} makes both generation and understanding easy, so the non-unification score is close to zero and may underestimate the internal gap. In contrast, for harder tasks such as \textit{generate a mirror reflection}, where the generation branch may fail to capture latent physical rules and the score may be overestimated. Therefore, stratifying by task difficulty provides a more reliable way to estimate the internal gap. Specifically, we construct nine subtasks of increasing difficulty from three benchmarks \citep{ghosh2023genevalobjectfocusedframeworkevaluating,huang2023t2i,li2025sciencet2iaddressingscientificillusions}, ranging from simple case (e.g., \texttt{a cat}) to complex prompts with implicit rules (e.g., \texttt{ice at \SI{60}{\celsius}}). Detailed tasks and MLLMs are shown in  \Cref{app:Details on Self-contradiction}.

\textbf{Results.} \cref{fig:non-unification}(c) demonstrates non-unification is prominent across different MLLMs and shows a trend of increasing with task difficulty. On VILA-U \citep{wu2024vila}, it even reaches 58.47\%, meaning that nearly 60\% of generations are rejected (prompt misaligned) by understanding. More discussion on non‑unification is provided in \Cref{app:Full Results on Self-contradiction}.

To further distinguish whether non-unification comes from weak generation or misunderstanding, we use a stronger external model, Qwen2.5-VL-72B-Instruct~\citep{bai2025qwen25vltechnicalreport}, to check the accuracy of the understanding scores. Define Weak Generation as the probability that, when the MLLM’s understanding branch rejects an output, its judgment agrees with Qwen, i.e.,
\begin{align*}
     \text{Weak Generation}
     &\coloneqq\mathbb{P}\big( \pi^{\text{und}}_{\theta}(\mathbf{x},\ q(\mathbf{y})) = \pi^{\text{und}}_{\text{Qwen}}(\mathbf{x},\ q(\mathbf{y})) \,\big|\, \pi^{\text{und}}_{\theta}(\mathbf{x},\ q(\mathbf{y})) = 0 \big).
\end{align*}
\cref{fig:nonunified_score_janus} shows, across different task difficulties, all MLLMs achieve over 50\% and up to 100\% Weak Generation, indicating that the internal gap mainly stems from poor generation rather than misjudgments of understanding which well align with prior findings \citep{yang2025hermesflow}.

\section{Mitigating Non-Unification: A Self-Improvement Framework}
\label{sec:Mitigating Non-Unification: A Self-Improvement Framework}
\subsection{Method: Internal Gap-based Self-improvement}
\label{sec:Internal Gap-based Self-improvement}

The observation that understanding consistently outperforms generation then motivates our \textit{internal gap-based self-improvement} framework to promote unification of unified MLLMs, which leverages stronger understanding to enhance the weaker generation. Specifically, we adopt standard post-training strategies such as Direct Preference Optimization (DPO) and Supervised Fine-Tuning (SFT).  Given an image generation prompt $\mathbf{y}$, the MLLM $\pi_{\theta}$ produces $N$ candidate images, i.e., $\{\mathbf{x}_i\}_{i=1}^N = \pi_{\theta}(\mathbf{y})$. Each candidate $\mathbf{x}_i$ is paired with the question as $q(\mathbf{y}) \coloneqq \text{``Does this image describe } \mathbf{y} \text{?''}$ and processed by understanding branch $\pi^{\text{und}}_{\theta}$. Images judged (most likely) as aligned with the prompt are labeled as \textit{chosen}, while those judged (most likely) as misaligned are labeled as \textit{rejected}, forming preference data $(\mathbf{y}, \mathbf{x}_{\text{chosen}}, \mathbf{x}_{\text{rejected}})$ for DPO and supervision pairs $(\mathbf{y}, \mathbf{x}_{\text{chosen}})$ for SFT on the generation branch. \Cref{app:Additional Details on Self-Improvement} provides further details on post-training data construction, and Alg.~\ref{alg:igs-sft} outlines the SFT-based self-improvement procedure.
\begin{figure}[t]
\centering
\begin{minipage}{0.48\linewidth}
\AlgCompress\scriptsize
\begin{algorithm}[H]
\caption{Self-Improvement (SFT)}
\label{alg:igs-sft}
\KwIn{$\pi_{\theta}$, prompts $\mathcal{P}$, image candidates $N$, epochs $T$}
\KwData{$\mathcal{D}_{\text{SFT}}\!\leftarrow\!\emptyset$, discard pool $\mathcal{B}\!\leftarrow\!\emptyset$}
\For{$\mathbf{y}\in\mathcal{P}$}{
  $\{\mathbf{x}_i\}_{i=1}^{N}\!\leftarrow\!\pi^{\text{gen}}_{\theta}(\mathbf{y})$;
  $s_i\!\leftarrow\!\pi^{\text{und}}_{\theta}(\mathbf{x}_i,q(\mathbf{y}))\!\in\!\{0,1\}$;
  $\mathcal{C}\!\leftarrow\!\{\mathbf{x}_i:s_i=1\}$;
  \uIf{$|\mathcal{C}|=0$}{$\mathcal{B}\!\leftarrow\!\mathcal{B}\cup\{\mathbf{y}\}$}
  \Else{$\mathcal{D}_{\text{SFT}}\!\leftarrow\!\mathcal{D}_{\text{SFT}}\cup\{(\mathbf{y},\mathbf{x}_{\text{chosen}})\,|\,\mathbf{x}_{\text{chosen}}\!\in\!\mathcal{C}\}$}
}
\For{$t=1$ \KwTo $T$}{
     $\theta \leftarrow \theta - \eta \,\nabla_{\theta}\mathcal{L}_{\text{gen}}(\theta;\mathcal{D}_{\text{SFT}})$ ;
}
\end{algorithm}
\end{minipage}\hfill
\begin{minipage}{0.48\linewidth}
\AlgCompress\scriptsize
\begin{algorithm}[H]
\caption{Curriculum Replay}
\label{alg:curriculum}
\KwIn{$\pi_{\theta}$, discard pool $\mathcal{B}$, image candidates $N$, curriculum epochs $\mathcal{E}_{\text{cur}}$}
\KwData{$\mathcal{D}_{\text{SFT}}$ (shared with Alg.~\ref{alg:igs-sft})}
\BlankLine
\BlankLine
\For{$t\in\mathcal{E}_{\text{cur}}$}{
  \For{$\mathbf{y}\in\mathcal{B}$}{
    $\{\tilde{\mathbf{x}}_j\}_{j=1}^{N}\!\leftarrow\!\pi^{\text{gen}}_{\theta}(\mathbf{y})$;
    $\tilde{s}_j\!\leftarrow\!\pi^{\text{und}}_{\theta}(\tilde{\mathbf{x}}_j,q(\mathbf{y}))$;
    $\tilde{\mathcal{C}}\!\leftarrow\!\{\tilde{\mathbf{x}}_j:\tilde{s}_j=1\}$;
    \If{$|\tilde{\mathcal{C}}|>0$}{
      $\mathcal{D}_{\text{SFT}}\!\leftarrow\!\mathcal{D}_{\text{SFT}}\cup\{(\mathbf{y},\mathbf{x})\,|\,\mathbf{x}\!\in\!\tilde{\mathcal{C}}\}$;
      remove $\mathbf{y}$ from $\mathcal{B}$
    }
  }
}
\BlankLine
\BlankLine
\end{algorithm}
\end{minipage}
\vspace{-0.1in}
\end{figure}


\subsection{Experiment: Effectiveness of Self-Improvement on MLLMs}
\label{sec:Empirical Validation of Self-Improvement on MLLMs}
We then show effectiveness of proposed self-improvement through following experiments.
\subsubsection{Setup}
\label{sec:Setup}
\textbf{Baseline and Data.} To validate self-improvement, we apply it to two baselines: Janus-Pro-7B \citep{chen2025janusprounifiedmultimodalunderstanding} and Show-o \citep{xie2024showosingletransformerunify}. We ablate which MLLM components to optimize (e.g., the LLM and vision aligner) and find that updating only the shared LLM yields substantial gains. Further details are in \Cref{app:Fine-tuned architecture}.  Experiments are conducted on T2I-CompBench++ \citep{huang2023t2i}, which provides about 6000 text prompts as post-training candidates. After data construction, classical post-training strategies, SFT and DPO, are applied for generation-focused self-improvement. Further implementation details are in \Cref{app:Details on Self-Improvement}.

\textbf{Evaluation.} We compare self-improved and pre-trained \footnote{For clarity, we name MLLMs without self-improvement as pre-trained MLLMs, even if they may undergo post-training phases during training.} MLLMs on generation, unification and understanding. For generation, we follow T2I-CompBench++ metrics and measure unification by non-unification score. For understanding, we use win rate (excluding ties) \citep{zheng2023judging,chen2024mllm}: given validation text prompts with images generated by pre-trained MLLMs, models judge prompt–image alignment. Win rate is the proportion of cases where the self-improved MLLM disagrees with the pre-trained one but agrees with the stronger external judge, e.g., Qwen2.5-VL-72B-Instruct. For example, if the models disagree on three samples and the self-improved model matches Qwen on two, win rate is $2/3$. Pre- and post-trained models with comparable understanding achieve a win rate of $0.5$. Win rate enables tracking changes in understanding and generation on the same task, facilitating analysis of two branches. See \Cref{app:Additional Details on Self-Improvement} for more metric details.


\subsubsection{Results}
\label{sec:Results}
We summarize key findings under SFT as follows. The corresponding DPO results, largely consistent with SFT, are provided in \Cref{app:Full Results on Self-Improvement}.

\begin{figure}[t!]
\centering
  \includegraphics[width=1\textwidth, height=0.18\textheight]{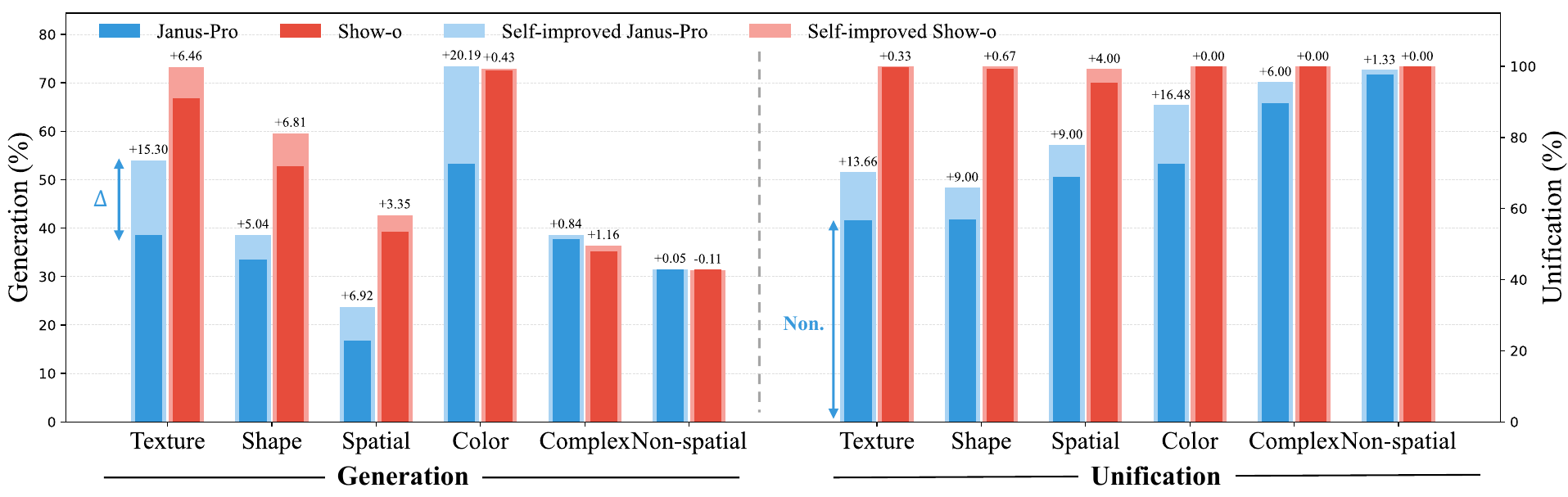}
\vspace{-0.3in}
\caption{Self-improvement enhances generation and unification, with gains up to 20\% and 16\% (1–non-unification score). Furthermore, improvements correlate with the internal gap (correlation coefficient $\rho_{\Delta,\text{Non.}}=0.53$): models and subtasks with larger gaps benefit more.}
\vspace{-0.1in}
\label{fig:sft_improve}
\end{figure}

\textbf{Finding 1: Internal gap-based self-improvement effectively improves generation and promotes MLLM unification.} \cref{fig:sft_improve} shows self-improved MLLMs can achieve up to 20\% gains in generation and up to 16\% in unification, validating effectiveness of proposed method. Moreover, we find improvements in generation are significantly correlated with unification ($\rho_{\Delta,\text{Non.}}=0.53$). Specially, for model level, Janus-Pro, with a larger internal gap (see \cref{fig:non-unification}), achieves greater gains than Show-o with a smaller gap. For task level, subtasks with lower unification (e.g., \texttt{Texture}) benefit more. We attribute this to internal gap–based method encouraging more post-training samples from larger-gap subtasks, thereby enabling greater improvements. \cref{fig:training_distribution} further confirms this by showing post-training data contain a higher proportion of samples from larger-gap subtasks.

\begin{figure}
  \centering
  \includegraphics[width=1\textwidth, height=0.165\textheight]{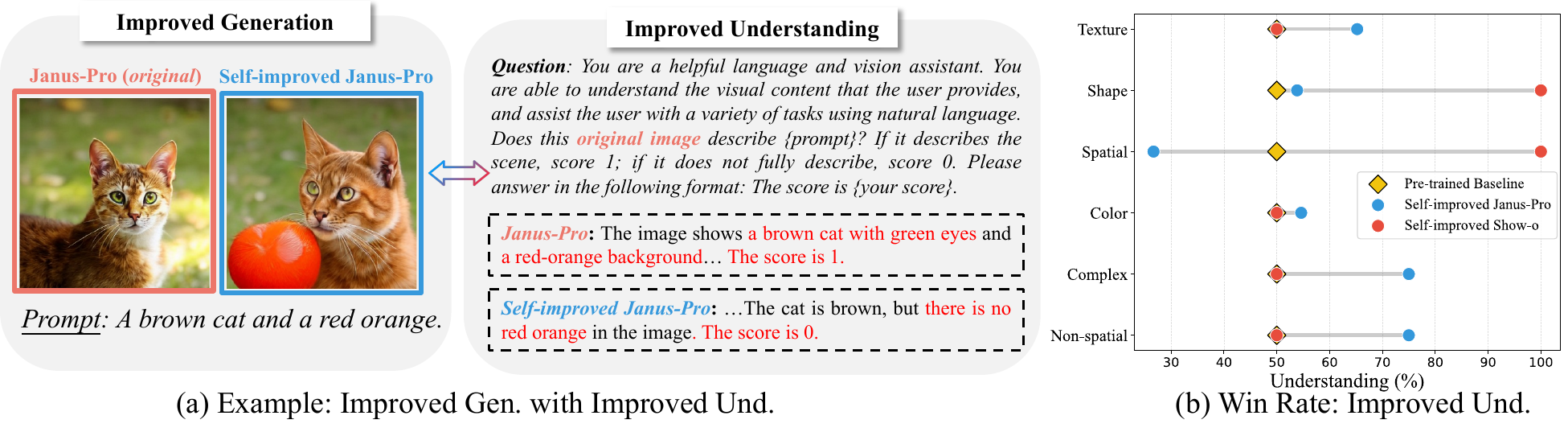}
\vspace{-0.3in}
  \caption{The Co-improvement Effect. (a) illustrates an example where self-improved Janus-Pro generates prompt-aligned images and correctly scores the original as mismatched (see more cases in \Cref{app:Details on Self-Improvement}); (b) reports win rates mostly above 50\%, showing  self-improved MLLMs judge prompt–image alignment more accurately than pre-trained ones.}
\vspace{-0.2in}
  \label{fig:improved_und}
\end{figure}

\textbf{Finding 2: Generation-targeted self-improvement also enhances understanding, showing a co-improvement effect.} \cref{fig:improved_und}(a) shows an example that, in addition to generating more prompt-aligned images, the self-improved MLLM also better detects mismatches between the original image and the prompt. \cref{fig:improved_und}(b) further reports high win rates for Janus-Pro and Show-o across six subtasks. For instance, the self-improved Janus-Pro achieves a win rate above 50\% on 5 of 6 subtasks, indicating higher accuracy than its pre-trained counterpart in judging prompt–image alignment. Additionally, we also provide results on standard understanding benchmarks in \Cref{tab:janus_und}, where self-improved MLLM consistently outperforms the pre-trained model.

\section{Understanding Co-improvement in Self-improvement}
\label{sec:Understanding Co-improvement in Self-improvement}
\Cref{sec:Results} reveals a co-improvement effect in self-improvement, an underexplored phenomenon in unified MLLMs (see \Cref{sec:Related Work}). Understanding this effect is crucial, as it highlights the unique interplay between generation and understanding and may inspire more effective self-improvement.


\subsection{Learning Dynamics of Generation and Understanding}
\label{sec:Learning Dynamics of Generation and Understanding}
We extend the learning dynamics  framework \citep{ren2025learningdynamicsllmfinetuning} to the multimodal setting, as it provides a principled way to analyze how MLLMs $\pi_{\theta}$ evolve after self-improvement on post-training data $(\mathbf{y}_u,\mathbf{x}_u)$. Specifically, the framework helps to answer: (1) \textit{Generation}:  given a text input $\mathbf{y}_0$, how generated images of the self-improved model differs from that of the base model; (2) \textit{Understanding}: given an image input $\mathbf{x}_0$, how understanding output  of the self-improved model differs from that of the base model.


Suppose $\mathbf{x}_0$ (from the pre-trained MLLM) and $\mathbf{y}_0$ are misaligned. If generation and understanding share aligned learning dynamics, e.g., jointly decreasing incorrect generation $\pi_{\theta}(\mathbf{x}_0|\mathbf{y}_0)$ and misunderstanding $\pi_{\theta}(\mathbf{y}_0|\mathbf{x}_0)$, the co-improvement occurs.

\textbf{Settings.} We first consider the setting where generation and understanding \textit{share the same tokenizer}, as in Show-o and EMU3 \citep{wang2024emu3}. This contrasts with decoupled designs (e.g., Janus-Pro) that use separate tokenizers. Nevertheless, our later analysis in \Cref{sec:Empirical Verification} indicates that the conclusions drawn under the shared-tokenizer assumption also apply to decoupled architectures. Additionally, our theoretical framework can also be extended to MLLMs that employ diffusion models for modeling continuous image tokens \citep{xie2024showosingletransformerunify,zhou2024transfusionpredicttokendiffuse}. Then, we denote $\mathcal{V}$ as the unified vocabulary of text and image tokens with size $V=|\mathcal{V}|$. Given a validation example $(\mathbf{y}_{0}, \mathbf{x}_{0})$, with image token sequence $\mathbf{x}_{0}=(x_{0,1},\ldots,x_{0,M})$ of length $M$ and text token sequence $\mathbf{y}_{0}=(y_{0,1},\ldots,y_{0,L})$ of length $L$, our goal is to analyze how MLLM's generation and understanding outputs on $(\mathbf{x}_{0}, \mathbf{y}_{0})$ change after self-improvement on the post-training sample $(\mathbf{y}_{u},\mathbf{x}_{u})$\footnote{\Cref{app:proof details} provides more detailed preliminaries.}.


Following \cite{ren2025learningdynamicsllmfinetuning}, we adopt standard causal masking in MLLMs \citep{wu2024janusdecouplingvisualencoding,wang2024emu3,wu2025harmonizing} and define the input to generation branch as $\mathcal{Y}_0 = [\,\mathbf{y}_0 \mid \mathbf{x}_0\,] \in \mathbb{R}^{d \times (M+L)}$, and input to understanding branch as
$\mathcal{X}_0 = [\,\mathbf{x}_0 \mid \mathbf{y}_0\,] \in \mathbb{R}^{d \times (M+L)}$\footnote{We omit potential special tokens (e.g., \texttt{[SOI]}) for simplicity.}. We denote the logit network as $h_\theta$, which outputs understanding and generation logits  $\mathbf{z}^0_{\mathrm{und}}\coloneqq h_{\theta}(\mathcal{X}_0)_{[:,\,M+1:M+L]}$ and $\mathbf{z}^0_{\mathrm{gen}}\coloneqq h_{\theta}(\mathcal{Y}_0)_{[:,\,L+1:L+M]}$ respectively.
We define the likelihood of sample $(\mathbf y_0, \mathbf x_0)$ under generation and understanding branch  as 
\begin{align}
    &\pi_\theta( \mathbf x_0 \mid \mathcal{Y}_0) = \prod^{M}_{k=1} \pi_\theta(x_{0, k} \mid  \mathbf y_0, \mathbf x_{0, < k}) = \prod^{M}_{k=1}  \left[{\rm softmax}(\mathbf{z}^0_{\mathrm{und}})\right]_{x_{0, k}, k} \tag{Generation} \\
    &\pi_\theta(\mathbf y_0 \mid \mathcal{X}_0) =  \prod^{L}_{\ell=1} \pi_\theta(y_{0, \ell} \mid \mathbf x_0, \mathbf y_{0, < \ell}) = \prod^{L}_{\ell=1} \left[{\rm softmax} (\mathbf{z}^0_{\mathrm{gen}}) \right]_{y_{0,\ell}, \ell} \tag{Understanding}
\end{align}
where the softmax is applied column-wise.



\textbf{One-step learning dynamics.}
At epoch $t$, we define the {one-step learning dynamics} of evaluation data pair $(\mathbf{y}_0,\mathbf{x}_0)$ likelihood after training one-step on post-training data $(\mathbf{y}_u,\mathbf{x}_u)$ as $\Delta  G_t \paren*{\mathbf{x}_0\mid\mathcal{Y}_0} \coloneqq \log \pi_{\theta_{t+1}}\!\paren*{\mathbf{x}_0\mid\mathcal{Y}_0}
- \log \pi_{\theta_{t}}\!\paren*{\mathbf{x}_0\mid\mathcal{Y}_0}$ for generation branch and $\Delta U_t \paren*{\mathbf{y}_0\mid\mathcal{X}_0}
\coloneqq \log \pi_{\theta_{t+1}}\!\paren*{\mathbf{y}_0\mid\mathcal{X}_0} - \log \pi_{\theta_{t}}\!\paren*{\mathbf{y}_0\mid\mathcal{X}_0}$ for understanding branch. 
We consider self-improvement with SFT and relate the dynamics of understanding and generation in the following proposition. Self-improvement with DPO are analyzed in \Cref{app:dpo}.
\begin{proposition}
[Learning Dynamics of Generation and Understanding under SFT]
\label{theorem:interpaly}
    Consider self-improvement proposed in \Cref{sec:Mitigating Non-Unification: A Self-Improvement Framework} with SFT and at epoch $t$.
    
    The one-step learning dynamics of \textbf{generation} is
\begin{align}
  \label{eq:gen-main}
  \Delta G_t(\mathbf{x}_0\mid\mathcal{Y}_0) =
  -\eta\sum_{k=1}^{M}\sum_{r=1}^{M}
  (\mathbf{e}_{x_{0,k}}-{\pi}^{0}_{k})^{\top}
  \textcolor{lightblue}{{\mathcal{K}^t_{k,r}(\mathcal{Y}_0,\mathcal{Y}_u)}}
  ({\pi}^{u}_{r}-\mathbf{e}_{x_{u,r}})
  + \mathcal{O}(\eta^2),
\end{align}
   where ${\pi}^{u}_{r}=\mathrm{softmax}(\mathbf{z}^{u}_{r})$ and $\mathbf{z}^{u}_{r}=[h_{\theta}(\mathcal{Y}_u)]_r$ are the logits at 
   position $r$ obtained by running $h_{\theta}$ on $\mathcal{Y}_u$ and $\mathcal{K}^t_{k,r}(\mathcal{Y}_0,\mathcal{Y}_u)\coloneqq(\nabla_{\theta_t}\mathbf{z}^{0}_{k})(\nabla_{\theta_t}\mathbf{z}^{u}_{r})^{\!\top} \in \mathbb{R}^{V \times V}$ is empirical neural tangent kernel (eNTK). 

 The one-step learning dynamics of \textbf{understanding} is
\begin{equation}
\label{eq:und-main}
\resizebox{\linewidth}{!}{$
\begin{aligned}
\Delta U_t(\mathbf{y}_0\mid\mathcal{X}_0)&= -\eta \sum^M_{k=1}\sum_{r=1}^{M}\sum_{\mathbf y_i\neq \mathbf y_0} w_{\theta_t}(\mathbf y_i\mid \mathbf x_0)\,
\Big(
(\mathbf{e}_{x_{0,k}}-{\pi}^{0}_{k})^{\!\top}
 \textcolor{lightblue}{{\mathcal{K}^{\,t}_{k,r}(\mathcal{Y}_0,\mathcal{Y}_u)}}
-
(\mathbf{e}_{x_{0,k}}-{\pi}^{i}_{k})^{\!\top}
\textcolor{lightred}{{\mathcal{K}^{\,t}_{k,r}(\mathcal{Y}_i,\mathcal{Y}_u)}}
\Big)
({\pi}^{u}_{r}-\mathbf{e}_{x_{u,r}})\\
&\quad + \mathcal{O}(\eta^2)
\end{aligned}
$}
\end{equation}
   where   $w_{\theta_t}(\mathbf y \mid \mathbf x_0)\;\coloneqq\;
\frac{\pi_{\theta_t}(\mathbf x_0\mid \mathbf y)}{\sum_{\mathbf y'} \pi_{\theta_t}(\mathbf x_0\mid \mathbf y')}$ and $\mathcal{Y}_i$ denotes the concatenation of prompt $\mathbf y_i \neq \mathbf y_0$ and $ \mathbf{x}_0$.
\end{proposition}

\Cref{theorem:interpaly} shows the learning dynamics of generation ($\Delta G_t$ in \Cref{eq:gen-main}) and understanding ($\Delta U_t$ in \Cref{eq:und-main}) are similar. The key difference is that $\Delta U_t$ includes an additional eNTK term, \textcolor{lightred}{\(\mathcal{K}^{\,t}_{k,r}(\mathcal{Y}_i,\mathcal{Y}_u)\)}, which measures alignment between \(\mathcal{Y}_i\) ($i\neq 0$) and the post-training data \(\mathcal{Y}_u\).  

\textit{We therefore hypothesize}: for co-improved pair \((\mathbf{y}_0,\mathbf{x}_0)\), there likely exist post-training samples \((\mathbf{y}_u,\mathbf{x}_u)\) that are highly similar, leading to $\|\textcolor{lightblue}{\mathcal{K}^{\,t}_{k,r}(\mathcal{Y}_0,\mathcal{Y}_u)}\|_{F} \geq \| \textcolor{lightred}{\mathcal{K}^{\,t}_{k,r}(\mathcal{Y}_i,\mathcal{Y}_u)}\|_F$. Hence, understanding update $\Delta U_t$ in \Cref{eq:und-main} is dominated by \textcolor{lightblue}{\(\mathcal{K}^{\,t}_{k,r}(\mathcal{Y}_0,\mathcal{Y}_u)\)}, which is a \emph{shared eNTK} term with the generation update \(\Delta G_t\) in \Cref{eq:gen-main}. Aligned updates between generation and understanding, i.e., aligned $\Delta G_t$ and $\Delta U_t$, can jointly reduce the probabilities of mis-generation $\pi_{\theta}(\mathbf{x}_0\mid\mathbf{y}_0)$ and misunderstanding $\pi_{\theta}(\mathbf{y}_0\mid\mathbf{x}_0)$, thereby yielding co-improvement.

To test this hypothesis, we combine empirical evidences from \Cref{sec:Empirical Validation of Self-Improvement on MLLMs} with theoretical results in \Cref{theorem:interpaly}, and empirically examine:  whether understanding-improving samples $(\mathbf{y}_0,\mathbf{x}_0)$ admit highly similar post-training sample $(\mathbf{y}_u,\mathbf{x}_u)$. Such similarity renders $\Delta U_t$ dominated by eNTK $\textcolor{lightblue}{\mathcal{K}^{\,t}_{k,r}(\mathcal{Y}_0,\mathcal{Y}_u)}$, which is shared by both $\Delta G_t$ and $\Delta U_t$, thereby aligning updates of both branches.

\subsection{Empirical Evidence}
\label{sec:Empirical Verification}
\begin{wrapfigure}{r}{0.45\textwidth}
\vspace{-0.05in}
\begin{center}
    \includegraphics[width=0.43\textwidth]{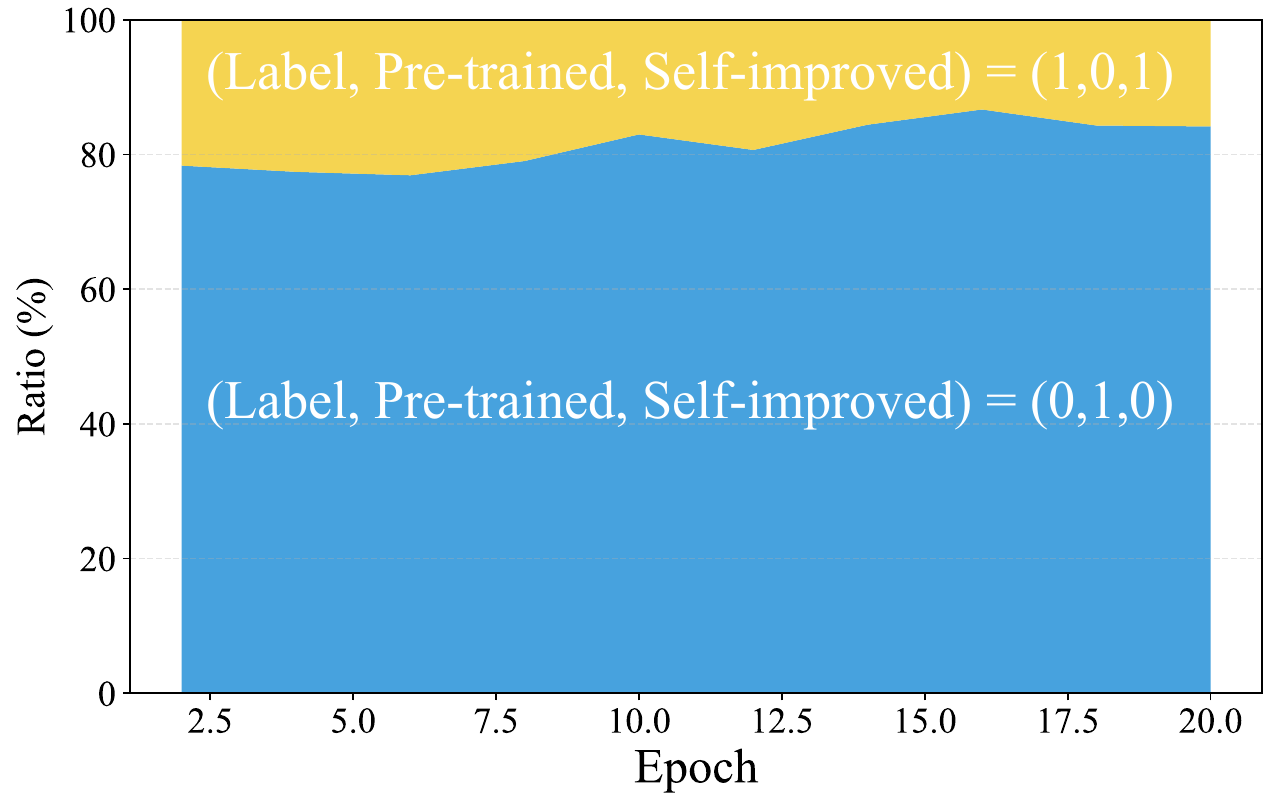}
\end{center}
\vspace{-0.25in}
\caption{On T2I-CompBench++, understanding gains primarily (80\%) arise from false positive correction . See \Cref{app:Learning Dynamics of Generation and Understanding} for results on additional MLLMs.} 
\vspace{-0.35in}
\label{fig:janus_und_source_sft}  
\end{wrapfigure}
First, understanding-improving samples can be classified into two cases: (1) False Positive Correction: when image $\mathbf{x}_0$ and text $\mathbf{y}_0$ are actually misaligned (Qwen label = 0), pre-trained MLLMs incorrectly judge them as aligned (score = 1), while self-improved MLLMs correctly predict misalignment (score = 0);  (2) False Negative Correction: when $\mathbf{x}_0$ and $\mathbf{y}_0$ are aligned (Qwen label = 1), pre-trained MLLMs incorrectly predict misalignment (score = 0), while self-improved MLLMs correctly judge alignment (score = 1). Using self‑improvement with SFT on Janus‑Pro as an example, \Cref{fig:janus_und_source_sft} shows approximately 80\% of the understanding improvement originates from case 1, i.e., false positive correction.
\begin{figure}[t!]
\centering
  \includegraphics[width=1\textwidth, height=0.15\textheight]{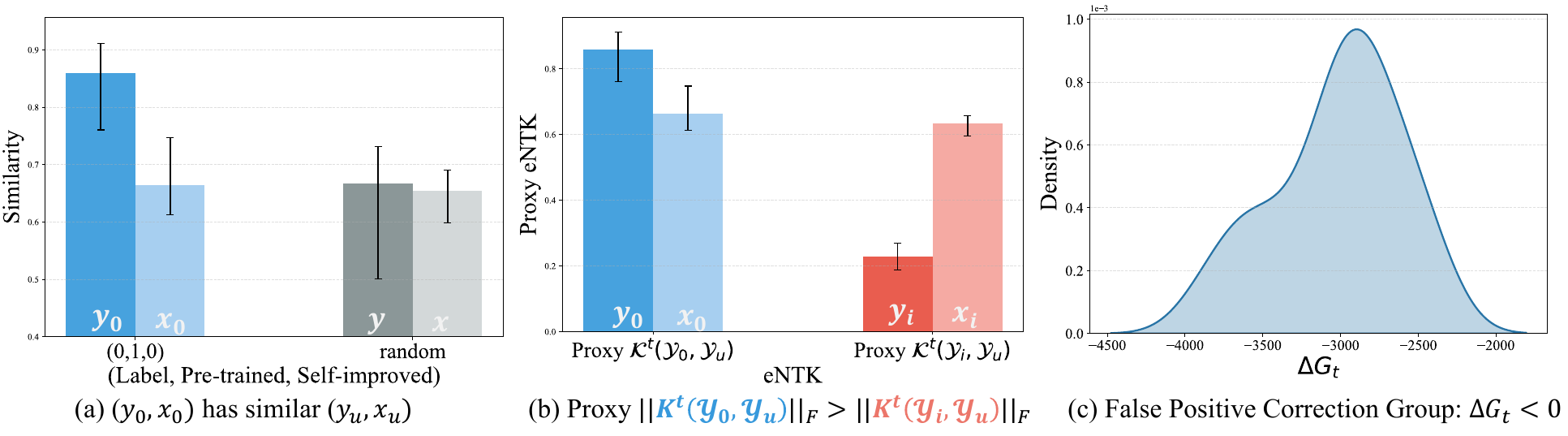}
\vspace{-0.25in}
\caption{Empirical Evidence from Self‑Improvement with Janus‑Pro and SFT. (a) Compared to random samples, those in the false positive correction group are more likely to be matched with highly similar post-training pairs $(\mathbf{y}_u,\mathbf{x}_u)$. (b) Such high similarity makes $\textcolor{lightblue}{\mathcal{K}^{\,t}_{k,r}(\mathcal{Y}_0,\mathcal{Y}_u)}$ be the dominant term in \Cref{eq:und-main}, thereby promoting aligned learning dynamics $\Delta G_t$ and $\Delta U_t$. (c) With aligned dynamics, $\Delta G_t < 0$ implies $\Delta U_t < 0$: both the probability of misaligned generation $\pi_{\theta}(\mathbf{x}_0 \mid \mathbf{y}_0)$ and misjudging  $\pi_{\theta}(\mathbf{y}_0 \mid \mathbf{x}_0)$, are reduced, i.e., co-improvement occurs.}
\vspace{-0.2in}
\label{fig:evidence}
\end{figure}

Therefore, our verification mainly focuses on \textit{false positive correction}. Specifically, consider $\mathbf{y}_0$ and misaligned image $\mathbf{x}_0$ (generated by pre-trained MLLMs), we find the following empirical evidence:
\begin{itemize}[leftmargin=0.2in]
  \item \cref{fig:evidence}(a) shows that samples from the false positive correction group \((\mathbf{y}_0,\mathbf{x}_0)\) typically have higher-similarity post-training counterparts \((\mathbf{y}_u,\mathbf{x}_u)\). In particular, the prompt \(\mathbf{y}_0\) achieves an average similarity of about \(0.8\), significantly higher than the randomly sampled reference.
  \item \cref{fig:evidence}(b) supports that, the understanding branch of data in false positive correction group is dominated by the eNTK term \textcolor{lightblue}{\(\mathcal{K}^{\,t}_{k,r}(\mathcal{Y}_0,\mathcal{Y}_u)\)}.  Using sample similarities (prompt- and image-level) as proxies for eNTK, we consistently observe $\|\textcolor{lightblue}{\mathcal{K}^{\,t}_{k,r}(\mathcal{Y}_0,\mathcal{Y}_u)}\|_{F} \geq \| \textcolor{lightred}{\mathcal{K}^{\,t}_{k,r}(\mathcal{Y}_i,\mathcal{Y}_u)}\|_F$.
  \item \cref{fig:evidence}(c) shows, for false positive correction samples, the generation update satisfies \(\Delta G_t<0\), i.e., the mis-generation probability \(\pi_{\theta}(\mathbf{x}_0\mid\mathbf{y}_0)\) decreases. Combined with \cref{fig:evidence}(a)(b), this further implies \(\Delta U_t<0\), meaning the misunderstanding probability \(\pi_{\theta}(\mathbf{y}_0\mid\mathbf{x}_0)\) also decreases. 
\end{itemize}
The above empirical evidence supports the hypothesis derived from \Cref{theorem:interpaly}, explaining both the mechanism of false positive correction and the emergence of co-improvement. We provide details on how each empirical result was obtained and interpreted in \Cref{app:Learning Dynamics of Generation and Understanding}.

\begin{table*}[t]
\centering
\setlength{\tabcolsep}{2.8pt}
\renewcommand{\arraystretch}{1.05}

\begin{adjustbox}{max width=\textwidth, keepaspectratio}
\begin{tabular}{l c c c *{6}{|ccc}}
\toprule
\HideVlines
\multirow{2}{*}{Model} & \multirow{2}{*}{ES} & \multirow{2}{*}{IS} & \multirow{2}{*}{CL}\tablefootnote{Notation: external signals (ES), internal signals (IS) and curriculum learning (CL).}
  & \multicolumn{3}{c}{Texture}
  & \multicolumn{3}{c}{Shape}
  & \multicolumn{3}{c}{Spatial}
  & \multicolumn{3}{c}{Color}
  & \multicolumn{3}{c}{Complex}
  & \multicolumn{3}{c}{Non-spatial} \\
\ShowVlines
\cmidrule(lr){5-7}\cmidrule(lr){8-10}\cmidrule(lr){11-13}\cmidrule(lr){14-16}\cmidrule(lr){17-19}\cmidrule(lr){20-22}
\HideVlines
& & & &
  Gen.$\uparrow$ & Und.$\uparrow$ & Non.$\downarrow$
& Gen.$\uparrow$ & Und.$\uparrow$ & Non.$\downarrow$
& Gen.$\uparrow$ & Und.$\uparrow$ & Non.$\downarrow$
& Gen.$\uparrow$ & Und.$\uparrow$ & Non.$\downarrow$
& Gen.$\uparrow$ & Und.$\uparrow$ & Non.$\downarrow$
& Gen.$\uparrow$ & Und.$\uparrow$ & Non.$\downarrow$ \\
\ShowVlines
\midrule

\multicolumn{22}{l}{\textit{Gen. only}}\\[2pt]
StrucDiffusion~\citep{feng2022training}
  & \xmark & \xmark & \xmark
  & 49.00   & -- & --
  & 42.18 & -- & --
  & 13.86 & -- & --
  & 49.90   & -- & --
  & 33.55   & -- & --
  & 31.11  & -- & --
\\
CompDiffusion~\citep{liu2022compositional}
  & \xmark & \xmark & \xmark
   & 36.45   & -- & --
  & 32.99 & -- & --
  & 8.00 & -- & --
  & 40.63   & -- & --
  & 28.98   & -- & --
  & 29.80  & -- & --
\\
Attend\&Excite~\citep{chefer2023attend}
  & \xmark & \xmark & \xmark
  & 59.63   & -- & --
  & 45.17 & -- & --
  & 14.55 & -- & --
  & 64.00   & -- & --
  & 34.01   & -- & --
  & 31.09  & -- & --
\\
PixArt-$\alpha$~\citep{chen2023pixart}
  & \xmark & \xmark & \xmark
   & 64.77 & -- & --
  & 49.27 & -- & --
  & 20.64   & -- & --
  & 66.90   & -- & --
  & 34.33   & -- & --
  & \textbf{31.97}   & -- & --
\\
CoMat~\citep{jiang2024comat}
  & \xmark & \xmark & \xmark
   & 64.68   & -- & --
  & 53.29 & -- & --
  & 24.28 & -- & --
  & \textbf{78.27}  & -- & --
  & 36.80   & -- & --
  & \underline{31.87}  & -- & --
\\
SDv1.5~\citep{rombach2022high}
  & \xmark & \xmark & \xmark
    & 41.86 & --   & --   
  & 37.13 & --   & --   
  & 11.65 & --   & --   
  & 37.58   & --   & --   
  & 30.47   & --   & --   
  & 31.12   & --   & --   
\\
SD-XL-base-1.0~\citep{podell2023sdxl}
  & \xmark & \xmark & \xmark
   & 52.99 & -- & --
  & 46.87 & -- & --
  & 21.31 & -- & --
  & 58.79   & -- & --
  & 32.37   & -- & --
  & 31.19  & -- & --
\\
FLUX.1~\citep{flux2024}
  & \xmark & \xmark & \xmark
  & 69.22   & -- & --
  & 57.18 & -- & --
  & 28.63 & -- & --
  & \underline{74.07}   & -- & --
  & 37.03   & -- & --
  & 31.27  & -- & --
\\
\midrule

\multicolumn{22}{l}{\textit{Gen. and Und.}}\\[2pt]
Janus-Pro-7B~\citep{chen2025janusprounifiedmultimodalunderstanding}
  & \xmark & \xmark & \xmark
   & 38.63 &50.00& 43.33 
  & 33.49& 50.00& 43.00 
  & 16.81& 50.00& 31.00
  & 53.22&  50.00&  27.33 
  & 37.73 &50.00& 10.33 
  & 31.40 &50.00& 2.33
\\
T2I\text{-}R1~\citep{jiang2025t2ir1reinforcingimagegeneration}\tablefootnote{For fair comparison, we generate images for T2I-R1 directly from original prompts, without using the understanding branch for prompt expansion.}
  & \cmark & \xmark & \xmark
   &50.91 & 52.50  &34.67 
  &37.80& 53.49 &36.00 
  & 24.22 & 45.00 & 23.67
  &70.47& 35.29 &11.33 
  &38.53& 72.73 &\underline{3.33} 
  & 31.38 &75.00 &\underline{1.00}
\\
\multicolumn{1}{>{\columncolor{lightblue}}l}{\textcolor{white}{{Self-improved Janus-Pro-7B}}}
& & & 
& \multicolumn{18}{l}{} \\ 
\quad + \textit{SFT}
  & \xmark & \cmark & \xmark
   &53.93& 65.22& 29.67 
  & 38.63& 53.85& 34.00 
  & 23.73 &26.67& 22.00
  &73.41&  54.62&  10.85
  &\underline{38.57} &75.00 &{4.33}
  & 31.45& 75.00& \underline{1.00}
\\
\quad  + \textit{C\text{-}SFT}
  & \xmark & \cmark & \cmark
    &56.38& 66.67 &28.33  
  & 39.86&  64.52 & 33.67 
  & 24.87 &38.46 &21.67
  &73.77 &52.14 &12.20  
  & \textbf{38.78}& 70.00&\underline{3.33}
  & 31.44& 75.00 &2.33
\\
\midrule
\multicolumn{22}{l}{\textit{Gen. and Und.}}\\[2pt]
Show-o~\citep{xie2024showosingletransformerunify}
  & \xmark & \xmark & \xmark
   & 66.80 &50.00 &\underline{0.33}
  & 52.72& 50.00 &0.67 
  & 39.31& 50.00 &4.67
  & 72.50 & 50.00&\textbf{0.00}
  & 35.17 & 50.00 &\textbf{0.00}
  & 31.43&  50.00&  \textbf{0.00}
\\
Hermesflow~\citep{yang2025hermesflow}
  & \cmark & \cmark & \xmark
   & 67.96& 50.00 &\underline{0.33}
  &51.81& 50.00 &\underline{0.33}
  & 38.45 & 0.00 & 4.00
  &72.96& 50.00& \underline{0.34}
  & 35.28&  50.00&  \textbf{0.00}
  & 31.42& 50.00 &\textbf{0.00}
\\
\multicolumn{1}{>{\columncolor{lightblue}}l}{\textcolor{white}{{Self-improved Show-o}}}
& & & 
& \multicolumn{18}{l}{}  \\
\quad  + \textit{SFT}
  & \xmark & \cmark & \xmark
   & \underline{73.26}& 50.00 &\textbf{0.00}
  & \underline{59.53}& 100.00& \textbf{0.00}
  & \underline{42.66}&  100.00 &\underline{0.67}
  & 72.93 &50.00 &\textbf{0.00}
  &36.33& 50.00& \textbf{0.00}
  & 31.32&  50.00&  \textbf{0.00}
\\
\quad  + \textit{C\text{-}SFT}
  & \xmark & \cmark & \cmark
    & \textbf{74.11}& 50.00& \textbf{0.00}
  & \textbf{59.75} &100.00 &\textbf{0.00}
  & \textbf{42.70}& 100.00& \textbf{0.33}
  & 72.38&  50.00 & \textbf{0.00}
  & 36.42&  50.00&  \textbf{0.00}
  &{31.53}&   50.00&   \textbf{0.00}
\\
\bottomrule
\end{tabular}
\end{adjustbox}
\vspace{-0.1in}
\caption{Curriculum learning‑based self‑improvement (\textit{C‑SFT}) yields better generation (higher {Gen.}) and understanding (higher {Und.}), and alleviates non‑unification (lower {Non.}). which even surpasses baselines rely on external reward models, such as T2I-R1 (built on Janus-Pro-7B) and HermesFlow (built on Show-o). Additional post-training strategy, e.g., DPO, and evaluations on more benchmarks are provided in \Cref{app:Full Results on Curriculum-Learning-Based Self-Improvement}.}
\label{tab:main_results}
\vspace{-0.3in}
\end{table*}

\section{Curriculum Learning for Stronger Self-Improvement}
\label{sec:Curriculum Learning for Stronger Self-Improvement}
Co-improvement effect motivates a curriculum learning~\citep{elman1993learning,bengio2009curriculum} approach for stronger self-improvement: as generation and understanding improve together, difficult samples that pre-trained MLLMs could not previously utilize  (due to weak generation or inaccurate understanding) can be incorporated later, forming an adaptive data expansion process based on prompt complexity \citep{li20252dcurridpotwodimensionalcurriculumlearning}. To demonstrate co-improvement incorporates more unused prompts, we compare two settings: (1) jointly improving generation and understanding, and (2) enhancing only a single branch (e.g., generation). As shown in \Cref{fig:cl-motivation}, co-improvement contributes about 1000 additional samples from discard pool $\mathcal{B}$ (defined in Alg.~\ref{alg:igs-sft}) versus roughly 600 for single-branch enhancement, supporting our motivation. Alg.~\ref{alg:curriculum} shows details of curriculum learning.

\begin{wraptable}{r}{0.45\textwidth}
\vspace{-0.15in}
\centering
\resizebox{\linewidth}{!}{%
\begin{tabular}{c c c}
\toprule
 & \textbf{Und.} & \textbf{Self-improved Und.} \\
\midrule
\textbf{Gen.} & 0 & 649 \\
\textbf{Self-improved Gen.} & 603 & \textbf{1091} \\
\bottomrule
\end{tabular}
} 
\vspace{-0.15in}
\caption{Co-improvement (self-improved both Und. and Gen.) adds 1091 samples from discard pool $\mathcal{B}$, compared to roughly 600 when improving only a single branch.}
\label{fig:cl-motivation}  
\vspace{-0.1in}
\end{wraptable}
\textbf{Setup.} Following the experimental setup in \Cref{sec:Setup}, we adopt self‑improvement with curriculum learning strategy. For Janus‑Pro and Show‑o, curriculum learning is introduced at epoch 10, during which the models regenerate and rescore previously unused prompts to produce additional post‑training samples.  Evaluation follows the same metrics in \Cref{sec:Setup}.  We provide more implementation details in \Cref{app:Details on Curriculum-Learning-Based Self-Improvement} and ablation study in \Cref{app:Curriculum Learning Parameters}.



\textbf{Baseline.} Apart from generation-only models, e.g., SDv1.5 \citep{rombach2022high}, we consider two unified MLLM baselines: T2I-R1 \citep{jiang2025t2ir1reinforcingimagegeneration} improves generation of Janus-Pro-7B by using multiple external reward models and provides comparison for Janus-Pro-7B-based self-improvement. And Hermesflow \citep{yang2025hermesflow} similarly employs external reward models, e.g., Bert \citep{devlin2019bert}, to enhance Show‑o, serving as a reference for Show‑o‑based approach.

\textbf{Results.} We report only SFT-based self-improvement with curriculum learning (denoted as \textit{C-SFT}) on T2I-CompBench++ evaluation set. Results for DPO-based method and additional benchmarks, such as GenEval \citep{ghosh2023genevalobjectfocusedframeworkevaluating} and Science-T2I \citep{li2025sciencet2iaddressingscientificillusions}, are provided in \Cref{app:Full Results on Curriculum-Learning-Based Self-Improvement}. As shown in \Cref{tab:main_results}, incorporating curriculum learning enables unified MLLMs to achieve stronger self‑improvement: compared with standard self‑improvement, \textit{C‑SFT} delivers consistent gains in generation, understanding, and unification across most tasks, even surpassing baselines that rely on external rewards, such as T2I‑R1 and Hermesflow. These results confirm the effectiveness of incorporating curriculum learning into the self‑improvement process.


\section{Conclusion and Limitation}

In this paper, we systematically study the internal generation–understanding gap in MLLMs, covering empirical validation, mitigation method, mechanistic analysis, and the design of improved methods. We demonstrate that internal gap-based self-improvement can effectively mitigate non-unification in MLLMs and further induce co-improvement between generation and understanding.

This work has the following limitations.  First, our exploration of self-improvement is restricted to limited MLLMs, such as Janus-Pro and Show-o. We leave validation on more models, e.g., Bagel \citep{deng2025bagel}, to future work. Second, we attribute the observed co-improvement to shared eNTK between generation and understanding. A deeper question, however, is why such NTK sharing arises in unified MLLMs, which calls for further investigation into model's mechanisms.




\bibliography{iclr2026_conference}

\begin{thebibliography}{64}
\providecommand{\natexlab}[1]{#1}
\providecommand{\url}[1]{\texttt{#1}}
\expandafter\ifx\csname urlstyle\endcsname\relax
  \providecommand{\doi}[1]{doi: #1}\else
  \providecommand{\doi}{doi: \begingroup \urlstyle{rm}\Url}\fi

\bibitem[Antol et~al.(2015)Antol, Agrawal, Lu, Mitchell, Batra, Zitnick, and Parikh]{antol2015vqa}
Stanislaw Antol, Aishwarya Agrawal, Jiasen Lu, Margaret Mitchell, Dhruv Batra, C~Lawrence Zitnick, and Devi Parikh.
\newblock Vqa: Visual question answering.
\newblock In \emph{Proceedings of the IEEE international conference on computer vision}, pp.\  2425--2433, 2015.

\bibitem[Bai et~al.(2025)Bai, Chen, Liu, Wang, Ge, Song, Dang, Wang, Wang, Tang, Zhong, Zhu, Yang, Li, Wan, Wang, Ding, Fu, Xu, Ye, Zhang, Xie, Cheng, Zhang, Yang, Xu, and Lin]{bai2025qwen25vltechnicalreport}
Shuai Bai, Keqin Chen, Xuejing Liu, Jialin Wang, Wenbin Ge, Sibo Song, Kai Dang, Peng Wang, Shijie Wang, Jun Tang, Humen Zhong, Yuanzhi Zhu, Mingkun Yang, Zhaohai Li, Jianqiang Wan, Pengfei Wang, Wei Ding, Zheren Fu, Yiheng Xu, Jiabo Ye, Xi~Zhang, Tianbao Xie, Zesen Cheng, Hang Zhang, Zhibo Yang, Haiyang Xu, and Junyang Lin.
\newblock Qwen2.5-vl technical report, 2025.
\newblock URL \url{https://arxiv.org/abs/2502.13923}.

\bibitem[Bengio et~al.(2009)Bengio, Louradour, Collobert, and Weston]{bengio2009curriculum}
Yoshua Bengio, J{\'e}r{\^o}me Louradour, Ronan Collobert, and Jason Weston.
\newblock Curriculum learning.
\newblock In \emph{Proceedings of the 26th annual international conference on machine learning}, pp.\  41--48, 2009.

\bibitem[Brown et~al.(2020)Brown, Mann, Ryder, Subbiah, Kaplan, Dhariwal, Neelakantan, Shyam, Sastry, Askell, Agarwal, Herbert-Voss, Krueger, Henighan, Child, Ramesh, Ziegler, Wu, Winter, Hesse, Chen, Sigler, Litwin, Gray, Chess, Clark, Berner, McCandlish, Radford, Sutskever, and Amodei]{brown2020languagemodelsfewshotlearners}
Tom~B. Brown, Benjamin Mann, Nick Ryder, Melanie Subbiah, Jared Kaplan, Prafulla Dhariwal, Arvind Neelakantan, Pranav Shyam, Girish Sastry, Amanda Askell, Sandhini Agarwal, Ariel Herbert-Voss, Gretchen Krueger, Tom Henighan, Rewon Child, Aditya Ramesh, Daniel~M. Ziegler, Jeffrey Wu, Clemens Winter, Christopher Hesse, Mark Chen, Eric Sigler, Mateusz Litwin, Scott Gray, Benjamin Chess, Jack Clark, Christopher Berner, Sam McCandlish, Alec Radford, Ilya Sutskever, and Dario Amodei.
\newblock Language models are few-shot learners, 2020.
\newblock URL \url{https://arxiv.org/abs/2005.14165}.

\bibitem[Chang et~al.(2022)Chang, Zhang, Jiang, Liu, and Freeman]{chang2022maskgitmaskedgenerativeimage}
Huiwen Chang, Han Zhang, Lu~Jiang, Ce~Liu, and William~T. Freeman.
\newblock Maskgit: Masked generative image transformer, 2022.
\newblock URL \url{https://arxiv.org/abs/2202.04200}.

\bibitem[Chefer et~al.(2023)Chefer, Alaluf, Vinker, Wolf, and Cohen-Or]{chefer2023attend}
Hila Chefer, Yuval Alaluf, Yael Vinker, Lior Wolf, and Daniel Cohen-Or.
\newblock Attend-and-excite: Attention-based semantic guidance for text-to-image diffusion models.
\newblock \emph{ACM transactions on Graphics (TOG)}, 42\penalty0 (4):\penalty0 1--10, 2023.

\bibitem[Chen et~al.(2024)Chen, Chen, Zhang, Wang, Liu, Zhou, Zhang, Wan, Zhou, and Sun]{chen2024mllm}
Dongping Chen, Ruoxi Chen, Shilin Zhang, Yaochen Wang, Yinuo Liu, Huichi Zhou, Qihui Zhang, Yao Wan, Pan Zhou, and Lichao Sun.
\newblock Mllm-as-a-judge: Assessing multimodal llm-as-a-judge with vision-language benchmark.
\newblock In \emph{Forty-first International Conference on Machine Learning}, 2024.

\bibitem[Chen et~al.(2025{\natexlab{a}})Chen, Xu, Pan, Hu, Qin, Goldstein, Huang, Zhou, Xie, Savarese, Xue, Xiong, and Xu]{chen2025blip3ofamilyfullyopen}
Jiuhai Chen, Zhiyang Xu, Xichen Pan, Yushi Hu, Can Qin, Tom Goldstein, Lifu Huang, Tianyi Zhou, Saining Xie, Silvio Savarese, Le~Xue, Caiming Xiong, and Ran Xu.
\newblock Blip3-o: A family of fully open unified multimodal models-architecture, training and dataset, 2025{\natexlab{a}}.
\newblock URL \url{https://arxiv.org/abs/2505.09568}.

\bibitem[Chen et~al.(2023)Chen, Yu, Ge, Yao, Xie, Wu, Wang, Kwok, Luo, Lu, et~al.]{chen2023pixart}
Junsong Chen, Jincheng Yu, Chongjian Ge, Lewei Yao, Enze Xie, Yue Wu, Zhongdao Wang, James Kwok, Ping Luo, Huchuan Lu, et~al.
\newblock Pixart-$alpha$: Fast training of diffusion transformer for photorealistic text-to-image synthesis.
\newblock \emph{arXiv preprint arXiv:2310.00426}, 2023.

\bibitem[Chen et~al.(2025{\natexlab{b}})Chen, Wu, Liu, Pan, Liu, Xie, Yu, and Ruan]{chen2025janusprounifiedmultimodalunderstanding}
Xiaokang Chen, Zhiyu Wu, Xingchao Liu, Zizheng Pan, Wen Liu, Zhenda Xie, Xingkai Yu, and Chong Ruan.
\newblock Janus-pro: Unified multimodal understanding and generation with data and model scaling, 2025{\natexlab{b}}.
\newblock URL \url{https://arxiv.org/abs/2501.17811}.

\bibitem[Deng et~al.(2025)Deng, Zhu, Li, Gou, Li, Wang, Zhong, Yu, Nie, Song, Shi, and Fan]{deng2025bagel}
Chaorui Deng, Deyao Zhu, Kunchang Li, Chenhui Gou, Feng Li, Zeyu Wang, Shu Zhong, Weihao Yu, Xiaonan Nie, Ziang Song, Guang Shi, and Haoqi Fan.
\newblock Emerging properties in unified multimodal pretraining.
\newblock \emph{arXiv preprint arXiv:2505.14683}, 2025.

\bibitem[Devlin et~al.(2019)Devlin, Chang, Lee, and Toutanova]{devlin2019bert}
Jacob Devlin, Ming-Wei Chang, Kenton Lee, and Kristina Toutanova.
\newblock Bert: Pre-training of deep bidirectional transformers for language understanding.
\newblock In \emph{Proceedings of the 2019 conference of the North American chapter of the association for computational linguistics: human language technologies, volume 1 (long and short papers)}, pp.\  4171--4186, 2019.

\bibitem[Dong et~al.(2023)Dong, Han, Peng, Qi, Ge, Yang, Zhao, Sun, Zhou, Wei, et~al.]{dong2023dreamllm}
Runpei Dong, Chunrui Han, Yuang Peng, Zekun Qi, Zheng Ge, Jinrong Yang, Liang Zhao, Jianjian Sun, Hongyu Zhou, Haoran Wei, et~al.
\newblock Dreamllm: Synergistic multimodal comprehension and creation.
\newblock \emph{arXiv preprint arXiv:2309.11499}, 2023.

\bibitem[Duan et~al.(2025)Duan, Fang, Wang, Wang, Huang, Zeng, Li, and Liu]{duan2025gotr1unleashingreasoningcapability}
Chengqi Duan, Rongyao Fang, Yuqing Wang, Kun Wang, Linjiang Huang, Xingyu Zeng, Hongsheng Li, and Xihui Liu.
\newblock Got-r1: Unleashing reasoning capability of mllm for visual generation with reinforcement learning, 2025.
\newblock URL \url{https://arxiv.org/abs/2505.17022}.

\bibitem[Dubey et~al.(2024)Dubey, Abhimanyu, Jauhri, Pandey, Kadian, Al-Dahle, Letman, and et~al.]{grattafiori2024llama3herdmodels}
Dubey, Abhimanyu, Abhinav Jauhri, Abhinav Pandey, Abhishek Kadian, Ahmad Al-Dahle, Aiesha Letman, and Akhil~Mathur et~al.
\newblock The llama 3 herd of models, 2024.
\newblock URL \url{https://arxiv.org/abs/2407.21783}.

\bibitem[Elman(1993)]{elman1993learning}
Jeffrey~L Elman.
\newblock Learning and development in neural networks: The importance of starting small.
\newblock \emph{Cognition}, 48\penalty0 (1):\penalty0 71--99, 1993.

\bibitem[Feng et~al.(2022)Feng, He, Fu, Jampani, Akula, Narayana, Basu, Wang, and Wang]{feng2022training}
Weixi Feng, Xuehai He, Tsu-Jui Fu, Varun Jampani, Arjun Akula, Pradyumna Narayana, Sugato Basu, Xin~Eric Wang, and William~Yang Wang.
\newblock Training-free structured diffusion guidance for compositional text-to-image synthesis.
\newblock \emph{arXiv preprint arXiv:2212.05032}, 2022.

\bibitem[Ge et~al.(2024)Ge, Zhao, Zhu, Ge, Yi, Song, Li, Ding, and Shan]{ge2024seed}
Yuying Ge, Sijie Zhao, Jinguo Zhu, Yixiao Ge, Kun Yi, Lin Song, Chen Li, Xiaohan Ding, and Ying Shan.
\newblock Seed-x: Multimodal models with unified multi-granularity comprehension and generation.
\newblock \emph{arXiv preprint arXiv:2404.14396}, 2024.

\bibitem[Ghosh et~al.(2023)Ghosh, Hajishirzi, and Schmidt]{ghosh2023genevalobjectfocusedframeworkevaluating}
Dhruba Ghosh, Hanna Hajishirzi, and Ludwig Schmidt.
\newblock Geneval: An object-focused framework for evaluating text-to-image alignment, 2023.
\newblock URL \url{https://arxiv.org/abs/2310.11513}.

\bibitem[Han et~al.(2025)Han, Han, Huang, Lu, and Zou]{han2025diffusionmodelslearnhidden}
Yujin Han, Andi Han, Wei Huang, Chaochao Lu, and Difan Zou.
\newblock Can diffusion models learn hidden inter-feature rules behind images?, 2025.
\newblock URL \url{https://arxiv.org/abs/2502.04725}.

\bibitem[Hong et~al.(2025)Hong, Zhang, Wang, Liu, Wen, and Yan]{hong2025reinforcing}
Jixiang Hong, Yiran Zhang, Guanzhong Wang, Yi~Liu, Ji-Rong Wen, and Rui Yan.
\newblock Reinforcing multimodal understanding and generation with dual self-rewards.
\newblock \emph{arXiv preprint arXiv:2506.07963}, 2025.

\bibitem[Huang et~al.(2023)Huang, Sun, Xie, Li, and Liu]{huang2023t2i}
Kaiyi Huang, Kaiyue Sun, Enze Xie, Zhenguo Li, and Xihui Liu.
\newblock T2i-compbench: A comprehensive benchmark for open-world compositional text-to-image generation.
\newblock \emph{Advances in Neural Information Processing Systems}, 36:\penalty0 78723--78747, 2023.

\bibitem[Hudson \& Manning(2019)Hudson and Manning]{hudson2019gqanewdatasetrealworld}
Drew~A. Hudson and Christopher~D. Manning.
\newblock Gqa: A new dataset for real-world visual reasoning and compositional question answering, 2019.
\newblock URL \url{https://arxiv.org/abs/1902.09506}.

\bibitem[Jiang et~al.(2024)Jiang, Song, Wu, Zhang, Shen, Zong, Liu, and Li]{jiang2024comat}
Dongzhi Jiang, Guanglu Song, Xiaoshi Wu, Renrui Zhang, Dazhong Shen, Zhuofan Zong, Yu~Liu, and Hongsheng Li.
\newblock Comat: Aligning text-to-image diffusion model with image-to-text concept matching.
\newblock \emph{Advances in Neural Information Processing Systems}, 37:\penalty0 76177--76209, 2024.

\bibitem[Jiang et~al.(2025)Jiang, Guo, Zhang, Zong, Li, Zhuo, Yan, Heng, and Li]{jiang2025t2ir1reinforcingimagegeneration}
Dongzhi Jiang, Ziyu Guo, Renrui Zhang, Zhuofan Zong, Hao Li, Le~Zhuo, Shilin Yan, Pheng-Ann Heng, and Hongsheng Li.
\newblock T2i-r1: Reinforcing image generation with collaborative semantic-level and token-level cot, 2025.
\newblock URL \url{https://arxiv.org/abs/2505.00703}.

\bibitem[Labs(2024)]{flux2024}
Black~Forest Labs.
\newblock Flux.
\newblock \url{https://github.com/black-forest-labs/flux}, 2024.

\bibitem[Li et~al.(2023{\natexlab{a}})Li, Wang, Wang, Ge, Ge, and Shan]{li2023seedbenchbenchmarkingmultimodalllms}
Bohao Li, Rui Wang, Guangzhi Wang, Yuying Ge, Yixiao Ge, and Ying Shan.
\newblock Seed-bench: Benchmarking multimodal llms with generative comprehension, 2023{\natexlab{a}}.
\newblock URL \url{https://arxiv.org/abs/2307.16125}.

\bibitem[Li et~al.(2025)Li, Chai, Fu, Xu, and Xie]{li2025sciencet2iaddressingscientificillusions}
Jialuo Li, Wenhao Chai, Xingyu Fu, Haiyang Xu, and Saining Xie.
\newblock Science-t2i: Addressing scientific illusions in image synthesis, 2025.
\newblock URL \url{https://arxiv.org/abs/2504.13129}.

\bibitem[Li et~al.(2022)Li, Li, Xiong, and Hoi]{li2022blipbootstrappinglanguageimagepretraining}
Junnan Li, Dongxu Li, Caiming Xiong, and Steven Hoi.
\newblock Blip: Bootstrapping language-image pre-training for unified vision-language understanding and generation, 2022.
\newblock URL \url{https://arxiv.org/abs/2201.12086}.

\bibitem[Li \& Zhang(2025)Li and Zhang]{li20252dcurridpotwodimensionalcurriculumlearning}
Mengyang Li and Zhong Zhang.
\newblock 2d-curri-dpo: Two-dimensional curriculum learning for direct preference optimization, 2025.
\newblock URL \url{https://arxiv.org/abs/2504.07856}.

\bibitem[Li et~al.(2023{\natexlab{b}})Li, Du, Zhou, Wang, Zhao, and Wen]{li2023evaluatingobjecthallucinationlarge}
Yifan Li, Yifan Du, Kun Zhou, Jinpeng Wang, Wayne~Xin Zhao, and Ji-Rong Wen.
\newblock Evaluating object hallucination in large vision-language models, 2023{\natexlab{b}}.
\newblock URL \url{https://arxiv.org/abs/2305.10355}.

\bibitem[Liang et~al.(2025)Liang, Yu, Luo, Iyer, Dong, Zhou, Ghosh, Lewis, tau Yih, Zettlemoyer, and Lin]{liang2025mixtureoftransformerssparsescalablearchitecture}
Weixin Liang, Lili Yu, Liang Luo, Srinivasan Iyer, Ning Dong, Chunting Zhou, Gargi Ghosh, Mike Lewis, Wen tau Yih, Luke Zettlemoyer, and Xi~Victoria Lin.
\newblock Mixture-of-transformers: A sparse and scalable architecture for multi-modal foundation models, 2025.
\newblock URL \url{https://arxiv.org/abs/2411.04996}.

\bibitem[Liu et~al.(2022)Liu, Li, Du, Torralba, and Tenenbaum]{liu2022compositional}
Nan Liu, Shuang Li, Yilun Du, Antonio Torralba, and Joshua~B Tenenbaum.
\newblock Compositional visual generation with composable diffusion models.
\newblock In \emph{European conference on computer vision}, pp.\  423--439. Springer, 2022.

\bibitem[Liu et~al.(2024)Liu, Duan, Zhang, Li, Zhang, Zhao, Yuan, Wang, He, Liu, Chen, and Lin]{liu2024mmbenchmultimodalmodelallaround}
Yuan Liu, Haodong Duan, Yuanhan Zhang, Bo~Li, Songyang Zhang, Wangbo Zhao, Yike Yuan, Jiaqi Wang, Conghui He, Ziwei Liu, Kai Chen, and Dahua Lin.
\newblock Mmbench: Is your multi-modal model an all-around player?, 2024.
\newblock URL \url{https://arxiv.org/abs/2307.06281}.

\bibitem[Mao et~al.(2025)Mao, Yang, and Shou]{mao2025unirlselfimprovingunifiedmultimodal}
Weijia Mao, Zhenheng Yang, and Mike~Zheng Shou.
\newblock Unirl: Self-improving unified multimodal models via supervised and reinforcement learning, 2025.
\newblock URL \url{https://arxiv.org/abs/2505.23380}.

\bibitem[OpenAI(2024)]{openai2024gpt4technicalreport}
OpenAI.
\newblock Gpt-4 technical report, 2024.
\newblock URL \url{https://arxiv.org/abs/2303.08774}.

\bibitem[Pang et~al.(2024)Pang, Yuan, Cho, He, Sukhbaatar, and Weston]{pang2024iterativereasoningpreferenceoptimization}
Richard~Yuanzhe Pang, Weizhe Yuan, Kyunghyun Cho, He~He, Sainbayar Sukhbaatar, and Jason Weston.
\newblock Iterative reasoning preference optimization, 2024.
\newblock URL \url{https://arxiv.org/abs/2404.19733}.

\bibitem[Peebles \& Xie(2023)Peebles and Xie]{peebles2023scalablediffusionmodelstransformers}
William Peebles and Saining Xie.
\newblock Scalable diffusion models with transformers, 2023.
\newblock URL \url{https://arxiv.org/abs/2212.09748}.

\bibitem[Podell et~al.(2023)Podell, English, Lacey, Blattmann, Dockhorn, M{\"u}ller, Penna, and Rombach]{podell2023sdxl}
Dustin Podell, Zion English, Kyle Lacey, Andreas Blattmann, Tim Dockhorn, Jonas M{\"u}ller, Joe Penna, and Robin Rombach.
\newblock Sdxl: Improving latent diffusion models for high-resolution image synthesis.
\newblock \emph{arXiv preprint arXiv:2307.01952}, 2023.

\bibitem[Qu et~al.(2024)Qu, Zhang, Liu, Wang, Jiang, Gao, Ye, Du, Yuan, and Wu]{qu2024tokenflowunifiedimagetokenizer}
Liao Qu, Huichao Zhang, Yiheng Liu, Xu~Wang, Yi~Jiang, Yiming Gao, Hu~Ye, Daniel~K. Du, Zehuan Yuan, and Xinglong Wu.
\newblock Tokenflow: Unified image tokenizer for multimodal understanding and generation, 2024.
\newblock URL \url{https://arxiv.org/abs/2412.03069}.

\bibitem[Radford et~al.(2021)Radford, Kim, Hallacy, Ramesh, Goh, Agarwal, Sastry, Askell, Mishkin, Clark, Krueger, and Sutskever]{radford2021learningtransferablevisualmodels}
Alec Radford, Jong~Wook Kim, Chris Hallacy, Aditya Ramesh, Gabriel Goh, Sandhini Agarwal, Girish Sastry, Amanda Askell, Pamela Mishkin, Jack Clark, Gretchen Krueger, and Ilya Sutskever.
\newblock Learning transferable visual models from natural language supervision, 2021.
\newblock URL \url{https://arxiv.org/abs/2103.00020}.

\bibitem[Rafailov et~al.(2024)Rafailov, Sharma, Mitchell, Ermon, Manning, and Finn]{rafailov2024directpreferenceoptimizationlanguage}
Rafael Rafailov, Archit Sharma, Eric Mitchell, Stefano Ermon, Christopher~D. Manning, and Chelsea Finn.
\newblock Direct preference optimization: Your language model is secretly a reward model, 2024.
\newblock URL \url{https://arxiv.org/abs/2305.18290}.

\bibitem[Ren \& Sutherland(2025)Ren and Sutherland]{ren2025learningdynamicsllmfinetuning}
Yi~Ren and Danica~J. Sutherland.
\newblock Learning dynamics of llm finetuning, 2025.
\newblock URL \url{https://arxiv.org/abs/2407.10490}.

\bibitem[Rombach et~al.(2022)Rombach, Blattmann, Lorenz, Esser, and Ommer]{rombach2022high}
Robin Rombach, Andreas Blattmann, Dominik Lorenz, Patrick Esser, and Bj{\"o}rn Ommer.
\newblock High-resolution image synthesis with latent diffusion models.
\newblock In \emph{Proceedings of the IEEE/CVF conference on computer vision and pattern recognition}, pp.\  10684--10695, 2022.

\bibitem[Team(2025)]{chameleonteam2025chameleonmixedmodalearlyfusionfoundation}
Chameleon Team.
\newblock Chameleon: Mixed-modal early-fusion foundation models, 2025.
\newblock URL \url{https://arxiv.org/abs/2405.09818}.

\bibitem[Tian et~al.(2024)Tian, Yang, Yang, Gao, Deng, Wang, Yu, Tao, Wan, ZHANG, et~al.]{tian2024videotetris}
Ye~Tian, Ling Yang, Haotian Yang, Yuan Gao, Yufan Deng, Xintao Wang, Zhaochen Yu, Xin Tao, Pengfei Wan, Di~ZHANG, et~al.
\newblock Videotetris: Towards compositional text-to-video generation.
\newblock \emph{Advances in Neural Information Processing Systems}, 37:\penalty0 29489--29513, 2024.

\bibitem[Tong et~al.(2024)Tong, Fan, Zhu, Xiong, Chen, Sinha, Rabbat, LeCun, Xie, and Liu]{tong2024metamorphmultimodalunderstandinggeneration}
Shengbang Tong, David Fan, Jiachen Zhu, Yunyang Xiong, Xinlei Chen, Koustuv Sinha, Michael Rabbat, Yann LeCun, Saining Xie, and Zhuang Liu.
\newblock Metamorph: Multimodal understanding and generation via instruction tuning, 2024.
\newblock URL \url{https://arxiv.org/abs/2412.14164}.

\bibitem[Verma et~al.(2024)Verma, Choi, Sharma, Watson-Daniels, Oh, and Kumar]{verma2024crossmodalprojectionmultimodalllms}
Gaurav Verma, Minje Choi, Kartik Sharma, Jamelle Watson-Daniels, Sejoon Oh, and Srijan Kumar.
\newblock Cross-modal projection in multimodal llms doesn't really project visual attributes to textual space, 2024.
\newblock URL \url{https://arxiv.org/abs/2402.16832}.

\bibitem[Wang et~al.(2024)Wang, Zhang, Luo, Sun, Cui, Wang, Zhang, Wang, Li, Yu, et~al.]{wang2024emu3}
Xinlong Wang, Xiaosong Zhang, Zhengxiong Luo, Quan Sun, Yufeng Cui, Jinsheng Wang, Fan Zhang, Yueze Wang, Zhen Li, Qiying Yu, et~al.
\newblock Emu3: Next-token prediction is all you need.
\newblock \emph{arXiv preprint arXiv:2409.18869}, 2024.

\bibitem[Wang et~al.(2004)Wang, Bovik, Sheikh, and Simoncelli]{wang2004image}
Zhou Wang, Alan~C Bovik, Hamid~R Sheikh, and Eero~P Simoncelli.
\newblock Image quality assessment: from error visibility to structural similarity.
\newblock \emph{IEEE transactions on image processing}, 13\penalty0 (4):\penalty0 600--612, 2004.

\bibitem[Wu et~al.(2024{\natexlab{a}})Wu, Chen, Wu, Ma, Liu, Pan, Liu, Xie, Yu, Ruan, and Luo]{wu2024janusdecouplingvisualencoding}
Chengyue Wu, Xiaokang Chen, Zhiyu Wu, Yiyang Ma, Xingchao Liu, Zizheng Pan, Wen Liu, Zhenda Xie, Xingkai Yu, Chong Ruan, and Ping Luo.
\newblock Janus: Decoupling visual encoding for unified multimodal understanding and generation, 2024{\natexlab{a}}.
\newblock URL \url{https://arxiv.org/abs/2410.13848}.

\bibitem[Wu et~al.(2025{\natexlab{a}})Wu, Jiang, Ma, Liu, Zhao, Yuan, Bai, and Bai]{wu2025liquidlanguagemodelsscalable}
Junfeng Wu, Yi~Jiang, Chuofan Ma, Yuliang Liu, Hengshuang Zhao, Zehuan Yuan, Song Bai, and Xiang Bai.
\newblock Liquid: Language models are scalable and unified multi-modal generators, 2025{\natexlab{a}}.
\newblock URL \url{https://arxiv.org/abs/2412.04332}.

\bibitem[Wu et~al.(2025{\natexlab{b}})Wu, Zhang, Xu, Jin, Wu, Tao, Liu, Li, and Loy]{wu2025harmonizing}
Size Wu, Wenwei Zhang, Lumin Xu, Sheng Jin, Zhonghua Wu, Qingyi Tao, Wentao Liu, Wei Li, and Chen~Change Loy.
\newblock Harmonizing visual representations for unified multimodal understanding and generation.
\newblock \emph{arXiv preprint arXiv:2503.21979}, 2025{\natexlab{b}}.

\bibitem[Wu et~al.(2023)Wu, Hao, Sun, Chen, Zhu, Zhao, and Li]{wu2023humanpreferencescorev2}
Xiaoshi Wu, Yiming Hao, Keqiang Sun, Yixiong Chen, Feng Zhu, Rui Zhao, and Hongsheng Li.
\newblock Human preference score v2: A solid benchmark for evaluating human preferences of text-to-image synthesis, 2023.
\newblock URL \url{https://arxiv.org/abs/2306.09341}.

\bibitem[Wu et~al.(2024{\natexlab{b}})Wu, Zhang, Chen, Tang, Li, Fang, Zhu, Xie, Yin, Yi, et~al.]{wu2024vila}
Yecheng Wu, Zhuoyang Zhang, Junyu Chen, Haotian Tang, Dacheng Li, Yunhao Fang, Ligeng Zhu, Enze Xie, Hongxu Yin, Li~Yi, et~al.
\newblock Vila-u: a unified foundation model integrating visual understanding and generation.
\newblock \emph{arXiv preprint arXiv:2409.04429}, 2024{\natexlab{b}}.

\bibitem[Xie et~al.(2025)Xie, Darrell, Zettlemoyer, and Wang]{xie2025reconstruction}
Ji~Xie, Trevor Darrell, Luke Zettlemoyer, and XuDong Wang.
\newblock Reconstruction alignment improves unified multimodal models.
\newblock \emph{arXiv preprint arXiv:2509.07295}, 2025.

\bibitem[Xie et~al.(2024)Xie, Mao, Bai, Zhang, Wang, Lin, Gu, Chen, Yang, and Shou]{xie2024showosingletransformerunify}
Jinheng Xie, Weijia Mao, Zechen Bai, David~Junhao Zhang, Weihao Wang, Kevin~Qinghong Lin, Yuchao Gu, Zhijie Chen, Zhenheng Yang, and Mike~Zheng Shou.
\newblock Show-o: One single transformer to unify multimodal understanding and generation, 2024.
\newblock URL \url{https://arxiv.org/abs/2408.12528}.

\bibitem[Xu et~al.(2023)Xu, Liu, Wu, Tong, Li, Ding, Tang, and Dong]{xu2023imagerewardlearningevaluatinghuman}
Jiazheng Xu, Xiao Liu, Yuchen Wu, Yuxuan Tong, Qinkai Li, Ming Ding, Jie Tang, and Yuxiao Dong.
\newblock Imagereward: Learning and evaluating human preferences for text-to-image generation, 2023.
\newblock URL \url{https://arxiv.org/abs/2304.05977}.

\bibitem[Yan et~al.(2025)Yan, Lin, Li, Ye, Han, Wang, Liu, Lin, Li, Xu, et~al.]{yan2025can}
Zhiyuan Yan, Kaiqing Lin, Zongjian Li, Junyan Ye, Hui Han, Zhendong Wang, Hao Liu, Bin Lin, Hao Li, Xue Xu, et~al.
\newblock Can understanding and generation truly benefit together--or just coexist?
\newblock \emph{arXiv preprint arXiv:2509.09666}, 2025.

\bibitem[Yang et~al.(2024)Yang, Yu, Meng, Xu, Ermon, and Cui]{yang2024mastering}
Ling Yang, Zhaochen Yu, Chenlin Meng, Minkai Xu, Stefano Ermon, and Bin Cui.
\newblock Mastering text-to-image diffusion: Recaptioning, planning, and generating with multimodal llms.
\newblock In \emph{Forty-first International Conference on Machine Learning}, 2024.

\bibitem[Yang et~al.(2025)Yang, Zhang, Tian, Shang, Xu, Zhang, and Cui]{yang2025hermesflow}
Ling Yang, Xinchen Zhang, Ye~Tian, Chenming Shang, Minghao Xu, Wentao Zhang, and Bin Cui.
\newblock Hermesflow: Seamlessly closing the gap in multimodal understanding and generation.
\newblock \emph{arXiv preprint arXiv:2502.12148}, 2025.

\bibitem[Zhang et~al.(2025)Zhang, Li, Li, Yang, and Cheng]{zhang2025unifiedvisionlanguagemodelsnecessary}
Jihai Zhang, Tianle Li, Linjie Li, Zhengyuan Yang, and Yu~Cheng.
\newblock Are unified vision-language models necessary: Generalization across understanding and generation, 2025.
\newblock URL \url{https://arxiv.org/abs/2505.23043}.

\bibitem[Zheng et~al.(2023)Zheng, Chiang, Sheng, Zhuang, Wu, Zhuang, Lin, Li, Li, Xing, et~al.]{zheng2023judging}
Lianmin Zheng, Wei-Lin Chiang, Ying Sheng, Siyuan Zhuang, Zhanghao Wu, Yonghao Zhuang, Zi~Lin, Zhuohan Li, Dacheng Li, Eric Xing, et~al.
\newblock Judging llm-as-a-judge with mt-bench and chatbot arena.
\newblock \emph{Advances in neural information processing systems}, 36:\penalty0 46595--46623, 2023.

\bibitem[Zhou et~al.(2024)Zhou, Yu, Babu, Tirumala, Yasunaga, Shamis, Kahn, Ma, Zettlemoyer, and Levy]{zhou2024transfusionpredicttokendiffuse}
Chunting Zhou, Lili Yu, Arun Babu, Kushal Tirumala, Michihiro Yasunaga, Leonid Shamis, Jacob Kahn, Xuezhe Ma, Luke Zettlemoyer, and Omer Levy.
\newblock Transfusion: Predict the next token and diffuse images with one multi-modal model, 2024.
\newblock URL \url{https://arxiv.org/abs/2408.11039}.

\end{thebibliography}
\bibliographystyle{iclr2026_conference}

\clearpage
\appendix

\begin{center}
    \Large{\textbf{Appendix}}
\end{center}

\etocdepthtag.toc{mtappendix}
\etocsettagdepth{mtchapter}{none}
\etocsettagdepth{mtappendix}{subsection}
\tableofcontents
\newpage

\section{Additional Details and Full Results on Internal Gap}
\label{app:Additional Experimental Details and Full Results}
\subsection{Additional Details}
\label{app:Details on Internal Gap}
In this section, we provide an overview of MLLMs and tasks evaluated in \Cref{sec:Phenomenon Validation: The Non-unification in MLLMs}. Unified MLLMs aim to integrate generation and understanding, with common approaches including extending understanding MLLMs with external diffusion models for generation \citep{dong2023dreamllm,tong2024metamorphmultimodalunderstandinggeneration,ge2024seed,yang2024mastering,tian2024videotetris,chen2025blip3ofamilyfullyopen,xie2024showosingletransformerunify}, or representing both images and text as discrete tokens and training unified transformers under autoregressive paradigm \citep{chameleonteam2025chameleonmixedmodalearlyfusionfoundation,zhou2024transfusionpredicttokendiffuse,qu2024tokenflowunifiedimagetokenizer,chen2025janusprounifiedmultimodalunderstanding,wang2024emu3}. Despite aiming to unify tasks, most MLLMs emphasize single-task SOTA performance while overlooking models’ internal alignment. Intuitively, truly unified MLLMs should maintain internal consistency between generation and understanding. Therefore, we first quantify at scale the non-unification problem in unified MLLMs.
\label{app:Details on Self-contradiction}
\paragraph{Evaluated MLLMs}
Our evaluation covers the following MLLMs:
\begin{itemize}[leftmargin=0.2in]
    \item \textbf{EMU3} \citep{wang2024emu3} is a unified model for both generation and understanding, which converts multiple modalities such as images, text, and video into discrete tokens, and performs next‑token prediction in mixed multimodal sequences based on an LLM‑style transformer architecture. EMU3 pursues maximal architectural unification between generation and understanding, sharing the same image tokenizer for both tasks and employing a common LLM backbone for generation and understanding.
    \item \textbf{Show-o} \citep{xie2024showosingletransformerunify} also follows an LLM‑style transformer architecture and an autoregressive paradigm. In its default setting, generation and understanding share the same visual understanding/generation encoder and LLM component. A distinctive feature of Show‑o is that it adopts different attention mechanisms for text and image tokens: causal attention for the former and full attention for the latter. Moreover, for image tokens during training, it is modeled using discrete diffusion and incorporates a mask token prediction mechanism similar to that of MaskGIT \citep{chang2022maskgitmaskedgenerativeimage}.
    \item \textbf{VILA-U} \citep{wu2024vila} also adopts a shared LLM and a unified next‑token prediction paradigm to integrate generation and understanding tasks. To better learn the discrete token sequences resulting from concatenated images and text, VILA-U innovatively trains a unified foundation vision tower by applying a CLIP‑like contrastive loss \citep{radford2021learningtransferablevisualmodels} between visual and textual tokens, while simultaneously enforcing accurate reconstruction of images after the decoder. This design promotes the performance of unified MLLMs.
    \item \textbf{Janus-Pro} \citep{chen2025janusprounifiedmultimodalunderstanding} differs slightly from the above models. While continuing to follow the LLM‑style shared transformer and autoregressive paradigm, it emphasizes decoupling generation and understanding tasks at the tokenizer stage. By employing separate image tokenizers for the two tasks, Janus-Pro aims to mitigate conflicts arising from using a single unified tokenizer to serve tasks which require different representations.
    \item \textbf{BAGEL} \citep{deng2025bagel}, in contrast, adopts an architecture that explicitly separates generation and understanding. Inspired by the Mixture‑of‑Transformers (MoT) paradigm \citep{liang2025mixtureoftransformerssparsescalablearchitecture}, BAGEL employs two dedicated transformer experts to handle the two types of information, respectively. The only point of interaction between the tasks is through the self‑attention mechanism within each transformer block, while other components, such as visual tokenizers and FFN, are fully decoupled by task.
    \item \textbf{BLIP3-o} \citep{chen2025blip3ofamilyfullyopen}, compared with the aforementioned models, adopts an even more decoupled design by combining an autoregressive paradigm with diffusion models. Specifically, BLIP3‑o follows an understand‑then‑generate pipeline: it first performs image understanding using an pre-trained understanding MLLM (e.g., Qwen2.5‑VL) to produce visual features that serve as semantic‑level conditions for the subsequent image generation task. Then, leveraging these semantic conditions, DiT \citep{peebles2023scalablediffusionmodelstransformers} learn the distribution of the original image representations in the CLIP \citep{radford2021learningtransferablevisualmodels}  embedding space via flow matching. During inference, a diffusion‑based visual decoder will reconstruct pixel‑level images from the CLIP representations generated by the DiT.
\end{itemize}

\begin{figure}[t]
\centering
    \hfill
    \subfigure[Easy Task]{\label{fig:easy-weak-gen}\includegraphics[width=0.3\linewidth]{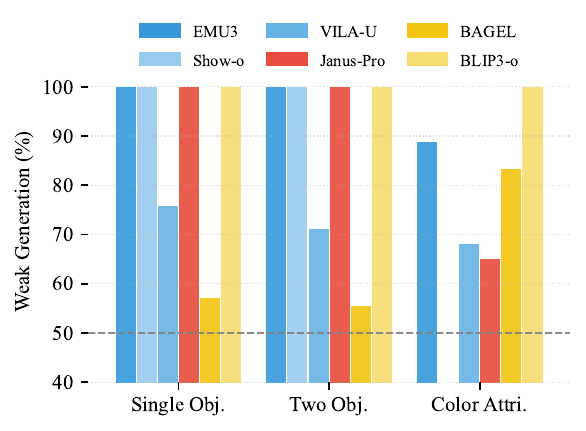}}
    \hfill
    \subfigure[Medium Task]{\label{fig:medium-weak-gen}\includegraphics[width=0.3\linewidth]{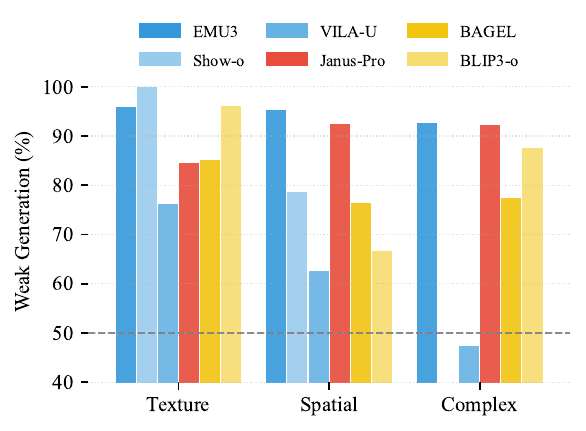}}
    \hfill
    \subfigure[Hard Task]{\label{fig:hard-weak-gen}\includegraphics[width=0.3\linewidth]{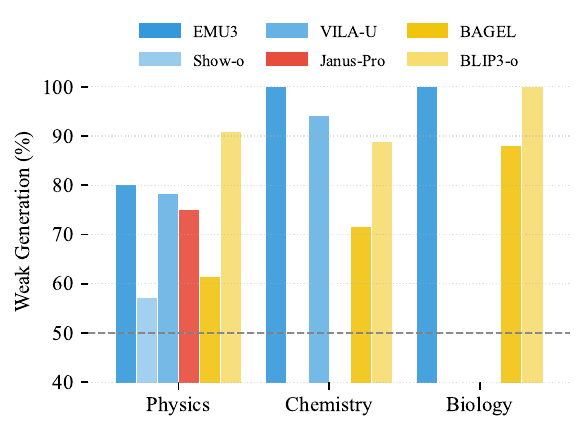}}
    \hfill
\vspace{-0.1in}
\caption{Full Results on Weak Generation. Our evaluation across six MLLMs and nine tasks indicates that the primary cause of non-unification is weak generation, as reflected by weak generation scores exceeding 50\% on the majority of tasks.}
\vspace{-0.15in}
\label{fig:weak-generation-details}
\end{figure}
\paragraph{Evaluated Task} We select nine subtasks from three benchmarks: GenEval \citep{ghosh2023genevalobjectfocusedframeworkevaluating}, T2I-CompBench++ \citep{huang2023t2i}, and Science-T2I \citep{li2025sciencet2iaddressingscientificillusions}. We then categorize subtasks into three difficulty levels (Easy, Medium, Hard) according to the complexity of generation and understanding required. \Cref{tab:task_level} provides a detailed description of each subtask. We observe that Easy subtasks focus on the generation and understanding of \textit{simple single objects}, e.g., \texttt{a cat}. Medium subtasks introduce relatively complex understanding such as spatial relationships (e.g., \texttt{on the top of}) that are typically made \textit{explicit} in prompts, and Hard subtasks involve \textit{implicit} reasoning not stated in the prompt, e.g., \texttt{tree in winter}, requiring MLLMs to leverage strong prior knowledge about physics, chemistry, and biology.
\begin{table}[h]
\centering
\resizebox{0.85\linewidth}{!}{
\begin{tabular}{l l c p{5cm} l}
\toprule
\textbf{Difficulty} & \textbf{Task} & \textbf{Evaluation Size} & \textbf{Prompt Example} & \textbf{Source} \\
\midrule
\multirow{3}{*}{Easy} 
& Single Obj. & 80  & \texttt{a photo of a cat} & \multirow{3}{*}{GenEval} \\
& Two Obj.    & 99 & \texttt{a photo of a stop sign and a dog} & \\
& Color Attri. & 100 & \texttt{a photo of a red cake and a purple chair} & \\
\midrule
\multirow{3}{*}{Medium} 
& Texture & 300 & \texttt{fluffy clouds and a glass table} & \multirow{3}{*}{T2I-CompBench++} \\
& Spatial   & 300 & \texttt{a cat on the top of a sofa} & \\
& Complex & 300 & \texttt{The prickly green cactus contrasted with the smooth white walls.} & \\
\midrule
\multirow{3}{*}{Hard} %
& Physics & 118  & \texttt{A ice block at sixty degrees Celsius, clear, simple and realistic.} & \multirow{3}{*}{Science-T2I-S} \\
& Chemistry    & 49 & \texttt{A iron ball that has been exposed to oxygen for decades, simple, clear and realistic.} & \\
& Biology & 60 & \texttt{A sweetgum tree in winter with high realism.} & \\
\bottomrule
\end{tabular}
}
\caption{Subtasks categorized by difficulty level. As shown in \cref{tab:task_level}, we select nine subtasks from three benchmarks to construct evaluation data with progressively increasing generation and understanding difficulty. Easy tasks involve only object generation, while Medium prompts require both generation and reasoning over spatial relations, colors, and textures. Hard tasks contain implicit reasoning, requiring MLLMs to possess accurate prior knowledge.}
\label{tab:task_level}
\end{table}

\subsection{Full Results}
\label{app:Full Results on Self-contradiction}
\paragraph{Full Results.} Following the non-unification score defined in \Cref{sec:Phenomenon Validation: The Non-unification in MLLMs}, we evaluate six MLLMs on subtasks across three difficulty levels and observe the widespread presence of the internal gap, as shown in \Cref{fig:non-unification}. In addition, we find substantial variation in non‑unification across MLLMs. Show‑o and EMU3 exhibit relatively small internal gaps, whereas recent models such as BAGEL and BLIP3‑o have larger gaps but stronger performance \citep{deng2025bagel,chen2025blip3ofamilyfullyopen}. It should be noted that the absolute performance of an MLLM is independent of its non‑unification score. First, non‑unification measures only the relative discrepancy between generation and understanding, rather than an MLLM’s absolute performance on each task. Moreover, differences in training configurations, such as data scale and pipeline design, can make comparisons between absolute performance and the relative gap across models unreliable.

\paragraph{Stronger Understanding and Human Check.}As described in \Cref{sec:Phenomenon Validation: The Non-unification in MLLMs}, we use stronger external models, such as Qwen2.5-VL-72B-Instruct, to evaluate the scores given by the understanding branch in order to identify the source of the internal capability imbalance in MLLMs, i.e., the internal gap. \Cref{fig:weak-generation-details} presents the weak generation rates across nine subtasks based on Qwen’s judgments, where we observe that most models exhibit more than 50\% weak generation on the majority of tasks. 
\begin{wrapfigure}{r}{0.45\textwidth}
\vspace{-0.25in}
\begin{center}
    \includegraphics[width=0.43\textwidth]{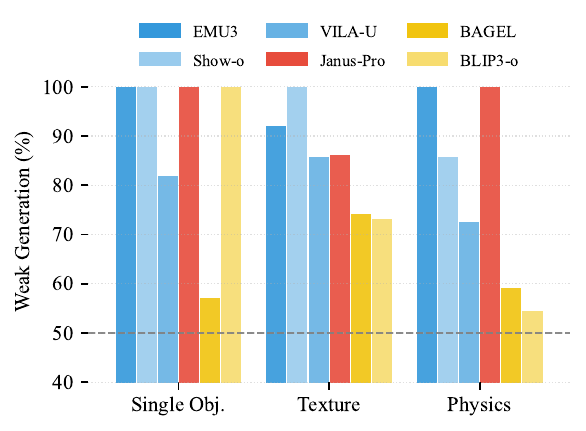}
\end{center}
\vspace{-0.3in}
\caption{Human-evaluated weak generation aligns with Qwen-based results, confirming weak generation as the primary cause of non-unification and supporting the use of Qwen as external judges in win rate.} 
\vspace{-0.25in}
\label{fig:human-check}  
\end{wrapfigure}
It should be noted that a weak generation rate of zero may arise partly from misjudgments of the understanding branch, e.g., Janus-Pro and VILA-U in Biology have nearly zero weak generation, and in other cases, e.g., Show-o, from a non-unification score of zero for that task, which naturally leads to a weak generation rate of zero.

\Cref{fig:human-check} further presents weak generation results based on human evaluation, which yield consistent findings: MLLMs achieve weak generation scores exceeding 50\% on the majority of tasks, further emphasizing that non-unification primarily stems from weak generation rather than misunderstanding. Moreover, the weak generation scores obtained from human evaluation are closely aligned with those derived from Qwen-based evaluation, with an average score difference of 1.01\% for Easy tasks, 8.21\% for Medium tasks, and 19.67\% for Hard tasks. The relatively larger discrepancy for Hard tasks may indicate that Qwen also faces limitations in understanding images involving implicit reasoning. Nevertheless, the overall agreement between human evaluation and Qwen in assessing MLLMs supports the continued use of Qwen as an external judge in subsequent studies, such as evaluating the win rate for understanding in \Cref{sec:Setup}.

\section{Additional Details and Full Results on Self-Improvement}
\label{app:Details on Self-Improvement}
\subsection{Additional Details}
\label{app:Additional Details on Self-Improvement}
\paragraph{Implementation Details.}  During the construction of SFT and DPO datasets, we feed each input image together with its corresponding question:
\begin{keyfindingbox}[Question]
You are a helpful language and vision assistant. You are able to understand the visual content that the user provides, and assist the user with a variety of tasks using natural language. Does this original image describe \{prompt\}? If it describes the scene, score 1; if it does not fully describe, score 0. Please answer in the following format: The score is \{your score\}.
\end{keyfindingbox}

We record the prediction probability from the understanding branch and select the image with the highest predicted probability of \textit{\{your score\}} = 1 as the chosen sample, and the image with the highest predicted probability of \textit{\{your score\}} = 0 as the rejected sample. The chosen images are used both as positive samples for DPO and as SFT samples, whereas the rejected images are used as negative samples for DPO. It is worth noting that, for DPO, we adopt the common practice of applying the negative log-likelihood (NLL) loss \citep{pang2024iterativereasoningpreferenceoptimization,grattafiori2024llama3herdmodels} over the preferred response in each pair, in order to enhance DPO. We conduct self-improvement on Janus-Pro-7B and Show-o (option (a) and $512 \times 512$) using four 80 GB NVIDIA A800 GPUs, with self-improvement epochs set to 20 for SFT and 30 for DPO, respectively. Self-improvement requires approximately 7–8 hours. The detailed hyperparameter configurations are presented in \Cref{tab:param}.
\begin{table}[h]
\centering
\resizebox{0.7\linewidth}{!}{
\begin{tabular}{l c c}
\toprule
\textbf{Hyperparameter} & \textbf{Janus-Pro-7B \tablefootnote{Our implementation is based on https://github.com/PKU-Alignment/align-anything.}} & \textbf{Show-o \tablefootnote{Our implementation is based on https://github.com/ZiyuGuo99/Image-Generation-CoT.}} \\
\midrule 
\multicolumn{3}{l}{\textit{Optimization}} \\[0.02in]
Optimizer & Adam & AdamW \\
Learning rate & $1\times 10^{-7}$ & $1\times 10^{-5}$ \\
Adam(W) $\beta$ & $[0.9, 0.95]$ & $[0.9, 0.999]$ \\
Weight decay & 0.05 & 0.01 \\
Warmup steps (Ratio) & 0.03 & 0.1  \\
Epoch & 20 (SFT) / 30 (DPO) & 20 (SFT) / 30 (DPO)\\
Grad. accumulation & 1 & 1 \\
Per-GPU batch size & 1 & 1\\
\midrule
\multicolumn{3}{l}{\textit{Trainable modules}} \\[0.02in]
Trainable parts & LLM & LLM   \\
Full Fine-tuning & \cmark & \cmark   \\
\midrule
\multicolumn{3}{l}{\textit{Loss weights}} \\[0.02in]
DPO $\beta$ & 0.01 & 0.01   \\
Weight NLL & 0.1 & 0.1    \\
CFG Weight & 5 & 5   \\
\midrule
\multicolumn{3}{l}{\textit{Data Construction}} \\[0.02in]
Image Size & $384 \times 384$ & $512 \times 512$ \\
Images per Prompt & 10 & 10   \\
Data Size & 1326   & 226    \\
\bottomrule
\end{tabular}
}
\caption{Hyperparameter configurations in self-improvement. For trainable parts, we only consider the LLM components shared by generation and understanding, which are sufficient to promote MLLMs. Additional trainable modules are discussed in \Cref{app:Fine-tuned architecture}.}
\label{tab:param}
\end{table}

\paragraph{Evaluation.}In addition to evaluating the self-improved MLLMs on the validation set of T2I-CompBench++, we also conduct evaluations on GenEval and Science-T2I. As introduced in \Cref{app:Details on Internal Gap}, GenEval is a relatively simple benchmark focusing on object and its basic attributes, whereas Science-T2I involves more complex prompts that require implicit reasoning. For image generation metrics, we follow the evaluation protocols and metric definitions specified by each benchmark. In addition, we adopt the definition of unification from \Cref{sec:Phenomenon Validation: The Non-unification in MLLMs}, namely 
\begin{align*}
\text{unification} \coloneqq 1-\text{non-unification score}.
\end{align*}
For evaluating understanding capability, we introduce the win rate metric. Specifically, the win rate (excluding ties) is defined as the proportion of samples where the understanding prediction changes after self-improvement and agrees with the score of stronger judge—Qwen2.5-VL-72B-Instruct.
We let $\pi_{\mathrm{pre}}$ and $\pi_{\mathrm{self}}$ denote the pre-trained and self-improved MLLMs, respectively. We define generations by pre-trained MLLMs as $\mathbf{x}_{\mathrm{pre}} = \pi^{\mathrm{gen}}_{\mathrm{pre}}(\mathbf{y})$ for the prompt $\mathbf{y}$. Win rate is:
\begin{align*}
\text{Win rate}
&:= \frac{
    \sum\limits_{\mathbf{y}} 
    \mathbb{I}\left[
      \pi^{\mathrm{und}}_{\mathrm{pre}}(\mathbf{x}_{\mathrm{pre}}, q(\mathbf{y})) \ne \pi^{\mathrm{und}}_{\mathrm{self}}(\mathbf{x}_{\mathrm{pre}}, q(\mathbf{y}))
      \ \land\ 
      \pi^{\mathrm{und}}_{\mathrm{self}}(\mathbf{x}_{\mathrm{pre}}, q(\mathbf{y})) = s_{\mathrm{Qwen}}
    \right]
}{
    \sum\limits_{\mathbf{y}} 
    \mathbb{I}\left[
      \pi^{\mathrm{und}}_{\mathrm{pre}}(\mathbf{x}_{\mathrm{pre}}, q(\mathbf{y})) \ne \pi^{\mathrm{und}}_{\mathrm{self}}(\mathbf{x}_{\mathrm{pre}}, q(\mathbf{y}))
    \right]
}
\end{align*}
 where $s_{\mathrm{Qwen}}(\mathbf{x}_{\mathrm{pre}}, q(\mathbf{y})) \in \{0, 1\}$ is oracle label provided by Qwen.  We introduce the win rate metric, which enables the simultaneous quantification of generation, understanding, and unification within the same task, thereby providing a better depiction of the synchronous changes between generation and understanding. In addition, we evaluate the understanding performance of MLLMs on dedicated benchmarks and provide illustrative examples for both before and after self-improvement.
\begin{table*}[h] 
\centering
\vspace{-0.1in}                    
\setlength{\tabcolsep}{2.8pt}     
\renewcommand{\arraystretch}{1.05}
\begin{adjustbox}{max width=\textwidth, max height=\textheight, keepaspectratio}
  \begin{tabular}{l ccc |ccc |ccc |ccc |ccc |ccc |ccc}
    \toprule
    \multirow{2}{*}{Model} & \multicolumn{3}{c}{Texture} & \multicolumn{3}{c}{Shape} & \multicolumn{3}{c}{Color} & \multicolumn{3}{c}{Spatial} & \multicolumn{3}{c}{Non-Spatial} & \multicolumn{3}{c}{Complex} & \multicolumn{3}{c}{Overall} \\
    \cmidrule(lr){2-4}\cmidrule(lr){5-7}\cmidrule(lr){8-10}\cmidrule(lr){11-13}\cmidrule(lr){14-16}\cmidrule(lr){17-19}\cmidrule(lr){20-22}
    & Gen.$\uparrow$ & Und.$\uparrow$ & Non.$\downarrow$
    & Gen.$\uparrow$ & Und.$\uparrow$ & Non.$\downarrow$
    & Gen.$\uparrow$ & Und.$\uparrow$ & Non.$\downarrow$
    & Gen.$\uparrow$ & Und.$\uparrow$ & Non.$\downarrow$
    & Gen.$\uparrow$ & Und.$\uparrow$ & Non.$\downarrow$
    & Gen.$\uparrow$ & Und.$\uparrow$ & Non.$\downarrow$
    & Gen.$\uparrow$ & Und.$\uparrow$ & Non.$\downarrow$ \\
    \midrule
\multicolumn{10}{l}{\textit{Gen. and Und.}}{\vspace{0.02in}}  \\
\pz\pz Janus-Pro-7B$_{\textit{\scriptsize (Baseline)}}$ & 38.63 & 50.00 & 43.33 & 33.49 & 50.00 & 43.00 & 53.22 & 50.00 & 27.33 & 16.81 & 50.00 & 31.00 & 31.40 & 50.00 & 2.33 & 37.73 & 50.00 & 10.33 & 35.21 & 50.00 & 26.22\\
\pz\pz\pz + \textit{SFT} & 53.93 & 65.22 & 29.67 &38.63 & 53.85 & 34.00 & 73.41 & 54.62 & 10.85 & 23.73 & 26.67& 22.00 & 31.45 & 75.00 & 1.00 & 38.57 & 75.00 & 4.33 & 43.29 & 58.39 & 16.98 \\ 
\pz\pz\pz + \textit{C-SFT} & 56.38 & 66.67&28.33&39.86&64.52&33.67&73.77&52.14&12.20&24.87&38.46&21.67&31.44&75.00&2.33&38.78&70.00&3.33&44.18&61.13&16.92
 \\ 
\pz\pz\pz + \textit{DPO} &40.98&53.85&43.00&33.49
&57.89&47.00&51.72&63.64&27.12&16.49&41.67&30.00&31.32&66.67&2.00&38.61&50.00&6.67&35.44&55.62&25.97
\\
\pz\pz\pz + \textit{C-DPO} &42.13&53.33&45.33&33.46&55.56&40.00&53.17&55.71&28.81&15.74&42.86&32.33&31.38&50.00&2.00&37.98&78.57&6.33
&35.64&56.00&25.80\\
\pz\pz  T2I-R1$_{\textit{\scriptsize (External)}}$ & 50.91 & 52.50 & 34.67 & 37.80 & 53.49 & 36.00& 70.47 & 35.29 & 11.33 & 24.22 & 45.00 & 23.67 & 31.38 & 75.00 & 1.00 & 38.53 & 72.73 & 3.33 & 42.22 & 55.67 & 18.33\\
\midrule
\multicolumn{10}{l}{\textit{Gen. and Und.}}{\vspace{0.02in}}  \\
\pz\pz Show-o$_{\textit{\scriptsize (Baseline)}}$ & 66.80 & 50.00 & 0.33 & 52.72 & 50.00 & 0.67 & 72.50 & 50.00 & 0.00 & 39.31 & 50.00 & 4.67 & 31.43 & 50.00 & 0.00 & 35.17 & 50.00 & 0.00 & 49.66 & 50.00 & 0.95 \\
\pz\pz\pz + \textit{SFT}&73.26&50.00&0.00&59.53&100.00&0.00&72.93&50.00&0.00&42.66&100.00&0.67&31.32&50.00&0.00&36.33&50.00&0.00&52.67&66.67&0.11\\
\pz\pz\pz + \textit{C-SFT}&74.11&50.00&0.00&59.75&100.00&0.00&72.38&50.00&0.00&42.70&100.00&0.33&31.53&50.00&0.00&36.42&50.00&0.00&52.82&66.67&0.06 \\
\pz\pz\pz + \textit{DPO} & 69.97&50.00&0.33&55.45&50.00&0.00&73.67&50.00&0.34&42.59&66.67&2.00&31.61&50.00&0.00&35.71&50.00&0.00&51.50&52.78&0.45
\\
\pz\pz\pz + \textit{C-DPO}&70.32&50.00&0.00&57.32&50.00&1.00&75.39&50.00&0.00&44.55&100.00&1.33&31.52&50.00&0.00&35.47&50.00&0.00&52.43&58.33&0.39 \\
\pz\pz  Hermsflow$_{\textit{\scriptsize (External)}}$ & 67.96 & 50.00 & 0.33 & 51.81 & 50.00 & 0.33 & 72.96 & 50.00 & 0.34 & 38.45 & 0.00 & 4.00 & 31.42 & 50.00 & 0.00 & 35.28 & 50.00 & 0.00 & 49.65 &  41.67 & 0.83 \\

    \bottomrule
  \end{tabular}
\end{adjustbox}
\vspace{-0.1in}
\caption{Evaluation Results on T2I-CompBench++. Self-improvement enhances MLLMs in generation, understanding, and unification, achieving results comparable to or even surpassing those of baselines that leverage external rewards.}
\label{tab:t2i}
\end{table*}

\begin{table*}[h] 
\centering
\vspace{-0.1in} 
\setlength{\tabcolsep}{2.8pt}     
\renewcommand{\arraystretch}{1.05}

\begin{adjustbox}{max width=\textwidth, max height=\textheight, keepaspectratio}
  \begin{tabular}{l ccc |ccc |ccc |ccc |ccc |ccc |ccc}
    \toprule
    \multirow{2}{*}{Model} & \multicolumn{3}{c}{Single Obj.} & \multicolumn{3}{c}{Two Obj.} & \multicolumn{3}{c}{Counting} & \multicolumn{3}{c}{Colors} & \multicolumn{3}{c}{Position} & \multicolumn{3}{c}{Color Attri.} & \multicolumn{3}{c}{Overall} \\
    \cmidrule(lr){2-4}\cmidrule(lr){5-7}\cmidrule(lr){8-10}\cmidrule(lr){11-13}\cmidrule(lr){14-16}\cmidrule(lr){17-19}\cmidrule(lr){20-22}
    & Gen.$\uparrow$ & Und.$\uparrow$ & Non.$\downarrow$
    & Gen.$\uparrow$ & Und.$\uparrow$ & Non.$\downarrow$
    & Gen.$\uparrow$ & Und.$\uparrow$ & Non.$\downarrow$
    & Gen.$\uparrow$ & Und.$\uparrow$ & Non.$\downarrow$
    & Gen.$\uparrow$ & Und.$\uparrow$ & Non.$\downarrow$
    & Gen.$\uparrow$ & Und.$\uparrow$ & Non.$\downarrow$
    & Gen.$\uparrow$ & Und.$\uparrow$ & Non.$\downarrow$ \\
    \midrule
\multicolumn{10}{l}{\textit{Gen. and Und.}}{\vspace{0.02in}}  \\
\pz\pz Janus-Pro-7B$_{\textit{\scriptsize (Baseline)}}$ & 98.75 & 50.00 & 3.75 &85.86 & 50.00 & 4.04 & 61.50 & 50.00 & 2.50 &84.04 & 50.00 & 2.13 & 75.00 & 50.00 & 5.00 & 71.00 & 50.00 & 20.00 & 79.36 & 50.00 & 6.24 \\
\pz\pz\pz + \textit{SFT}& 96.25 & 100.00&2.50&87.88&0.00&7.07&65.00&50.00&5.00&87.23&66.67&1.06&78.00&40.00&5.00&65.00&50.00&13.00&79.89&51.11&5.61\\ 
\pz\pz\pz + \textit{C-SFT} & 98.75 & 100.00 & 6.25 & 88.89 & 0.00 & 5.05 & 66.25 & 0.00 & 6.25 & 88.30 & 100.00 & 8.51 & 79.00 & 40.00 & 6.00 & 64.00 & 66.67 & 15.00 & 80.87 & 51.11 & 7.84 \\ 
\pz\pz\pz + \textit{DPO} & 98.75 & 50.00 & 2.50 & 89.90 & 0.00 & 6.06 & 56.25 & 50.00 & 6.25 & 88.30 & 50.00 & 3.19 & 73.00 & 50.00 & 6.00 & 69.00 & 100.00 & 13.00 & 79.20 & 50.00 &  6.17\\
\pz\pz\pz + \textit{C-DPO}&97.50&100.00&4.25
&85.86&0.00&5.10&60.00&50.00&5.25&88.30&100.00&1.06
&82.00&50.00&1.00&69.00&43.33&20.00&80.44&57.22&6.11
 \\
\pz\pz  T2I-R1$_{\textit{\scriptsize (External)}}$ & 98.75 & 50.00 & 7.50 & 86.87 & 0.00 & 8.08 & 58.75 & 0.00 & 7.50 & 87.23 & 100.00 & 1.06 & 83.00 & 60.00 & 5.00 & 70.00 & 50.00 & 22.00 & 80.77 & 43.30 & 8.52\\
\midrule
\multicolumn{10}{l}{\textit{Gen. and Und.}}{\vspace{0.02in}}  \\
\pz\pz Show-o$_{\textit{\scriptsize (Baseline)}}$ & 97.50 & 50.00 & 1.25 & 80.81& 50.00 & 2.02 & 76.25 & 50.00 &2.50 & 85.11 & 50.00 & 0.00 & 28.00 & 50.00 & 2.00 & 53.00 & 50.00 & 0.00 & 70.11 & 50.00 & 1.30 \\
\pz\pz\pz + \textit{SFT} & 97.50&50.00&1.25&91.92&50.00&0.00&61.25&50.00&0.00&78.72&50.00&0.00&37.00&50.00&2.00&
62.00&50.00&0.00&71.40&50.00&0.54
 \\
\pz\pz\pz + \textit{C-SFT}&96.25&50.00&1.25&86.87&50.00&0.00&67.50&50.00&0.00&78.72&50.00&1.06&44.00&50.00&1.00&66.00&50.00&1.00&73.22&50.00&0.72 \\
\pz\pz\pz + \textit{DPO}&97.25&50.00&1.25&84.85&50.00&0.00&71.25&50.00&0.00
&84.04&50.00&0.00&38.00&50.00&1.00&52.00&50.00&0.00&71.23&50.00&0.38\\ 
\pz\pz\pz + \textit{C-DPO}&97.50&50.00&1.25&84.85&50.00&0.00&70.00&50.00&0.00&86.17&50.00&0.00&37.00&50.00&1.00&59.00&50.00&0.00&72.42&50.00&0.38 \\ 
\pz\pz  Hermsflow$_{\textit{\scriptsize (External)}}$ & 96.25 & 50.00 & 1.25 & 83.84 & 50.00 & 1.01 & 66.25 & 50.00 & 1.25 & 80.85& 50.00 & 1.06 &
35.00 & 50.00 & 2.00 & 46.00 & 50.00 & 0.00 & 68.03 & 50.00 & 1.10 \\
    \bottomrule
  \end{tabular}
\end{adjustbox}
\vspace{-0.1in}
\caption{Evaluation Results on Geneval. Self-improvement enhances MLLMs in generation, understanding, and unification, achieving results comparable to or even surpassing those of baselines that leverage external rewards.}
\label{tab:geneval}
\end{table*}

\begin{table}[h]
\centering
\vspace{-0.1in}
\renewcommand{\arraystretch}{1.2}
\setlength{\tabcolsep}{4.5pt}
\resizebox{1\linewidth}{!}{
\begin{tabular}{lccc |ccc |ccc |ccc}
\toprule
\multirow{2}{*}{\pz\pz Model} & \multicolumn{3}{c}{Physics} & \multicolumn{3}{c}{Chemistry} & \multicolumn{3}{c}{Biology} & \multicolumn{3}{c}{Overall}\\
\cmidrule(lr){2-4} \cmidrule(lr){5-7} \cmidrule(lr){8-10} \cmidrule(lr){11-13}
 &  Gen.$\uparrow$ & Und.$\uparrow$ & Non.$\downarrow$ &  Gen.$\uparrow$ & Und.$\uparrow$ & Non.$\downarrow$ &  Gen.$\uparrow$ & Und.$\uparrow$ & Non.$\downarrow$ &  Gen.$\uparrow$ & Und.$\uparrow$ & Non.$\downarrow$  \\
\midrule
\multicolumn{10}{l}{\textit{Gen. and Und.}}{\vspace{0.02in}}  \\
\pz\pz Janus-Pro-7B$_{\textit{\scriptsize (Baseline)}}$    &  25.37 & 50.00 & 3.39 & 25.57 & 50.00 & 2.04 & 22.54 & 50.00 &  5.00 & 24.49 & 50.00 & 3.48\\
\pz\pz\pz + \textit{SFT} & 25.47 & 33.33 & 3.39 & 26.85 & 100.00 & 0.00 & 22.90 & 75.00 & 3.33 & 25.07 & 69.44 & 2.24 \\ 
\pz\pz\pz + \textit{C-SFT}&25.48&25.00&1.69
&26.66&100.00&2.04&23.41&80.00&6.67&25.18&68.33&3.47
 \\ 
\pz\pz\pz + \textit{DPO} & 25.72 & 50.00 & 1.69
& 25.37 & 100.00 & 0.00 & 23.49 & 0.00 & 3.33 & 24.86 & 50.00 & 1.67\\
\pz\pz\pz + \textit{C-DPO}&25.72&50.00&1.39&25.44&50.00&1.16&22.76&66.67&5.00&24.64&55.56&2.52
 \\
\pz\pz  T2I-R1$_{\textit{\scriptsize (External)}}$  & 25.52  &0.00  & 2.54 & 25.28 & 100.00 & 2.04 & 22.64 & 66.67 & 5.00 & 24.48 & 55.56 & 3.19\\
\midrule
\multicolumn{10}{l}{\textit{Gen. and Und.}}{\vspace{0.02in}}  \\
\pz\pz Show-o$_{\textit{\scriptsize (Baseline)}}$   &  25.56 &50.00  & 5.93  & 26.13 & 50.00 & 0.00 & 22.48 & 50.00 &   0.00 &24.72 & 50.00 & 1.98\\
\pz\pz\pz + \textit{SFT} & 26.57&60.00&1.69&26.62&50.00&0.00&22.48&50.00&0.00&25.22&53.33&0.56 \\ 
\pz\pz\pz + \textit{C-SFT}&27.12&60.00&0.85&27.63&50.00&0.00&23.38&50.00&0.00&26.04&53.33&0.28 \\ 
\pz\pz\pz + \textit{DPO}&26.05&0.00&5.08&25.76&50.00&0.00&21.53&50.00&0.00&24.44&33.33&1.69\\ 
\pz\pz\pz + \textit{C-DPO}&25.93&50.00&5.93&25.71&50.00&0.00&22.51&50.00&0.00&24.72&50.00&1.98 \\ 
\pz\pz HermesFlow$_{\textit{\scriptsize (External)}}$  & 25.61& 54.00 & 5.46 & 26.47 & 50.00 & 0.00 & 21.91 & 50.00 & 0.00 & 24.66 & 51.33 & 1.82 \\
\bottomrule
\end{tabular}
}
\caption{Evaluation Results on Science-T2I-S. Self-improvement enhances MLLMs in generation, understanding, and unification, achieving results comparable to or even surpassing those of baselines that leverage external rewards.}
\label{tab:science}
\end{table}
\subsection{Full Results}
\label{app:Full Results on Self-Improvement}

\paragraph{Full Results on Self-Improvement.} \Cref{tab:t2i}, \Cref{tab:geneval}, and \Cref{tab:science} report the improvements in generation, understanding, and unification of self-improvemed MLLMs across three benchmarks. Results of self-improvemed MLLMs are comparable to, and even surpass, two baselines, T2I-R1 and Hermesflow, which rely on external rewards. Taking Janus-Pro under SFT as an example, self-improvement boosts its generation and unification performance on T2I-CompBench++ by an average of 8\% and 10\%, respectively. Moreover, compared to pre-trained Janus-Pro, its understanding capability is enhanced with win rate greater than 50\%. Improvement also observed on GenEval and Science-T2I. These experiemnts verify the effectiveness of our proposed approach.

\begin{wrapfigure}{r}{.5\textwidth}
\vspace{-.3in}
\centering
  \subfigure[SFT, Janus-Pro-7B]{
    \includegraphics[width=0.45\linewidth]{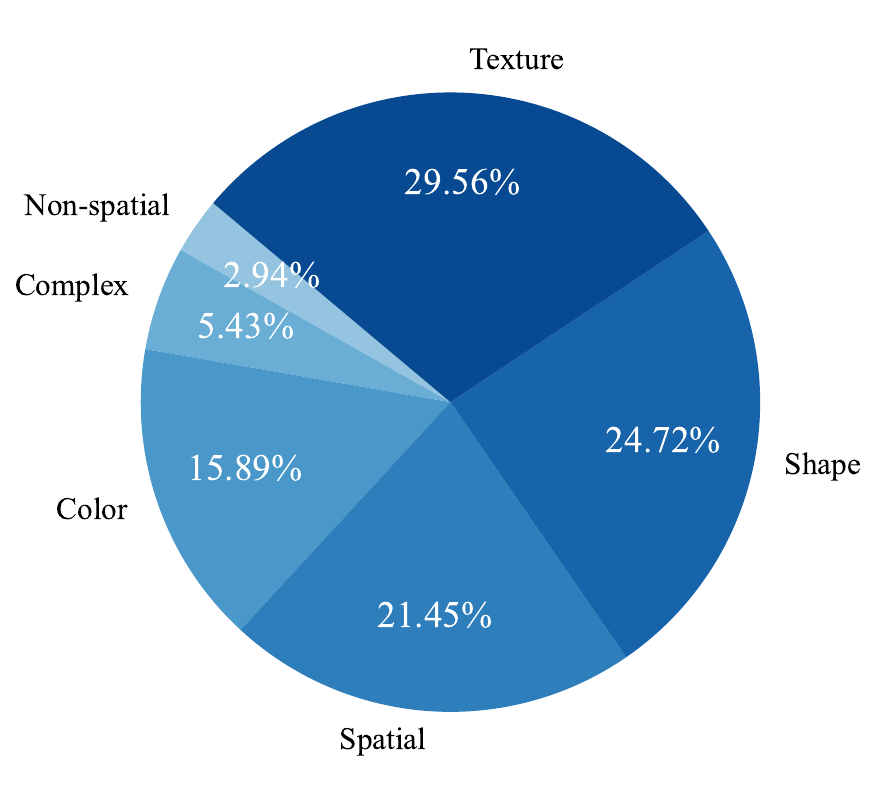}
    \label{fig:sft_janus_train}
  }
  \subfigure[SFT, Show-o]{
    \includegraphics[width=0.45\linewidth]{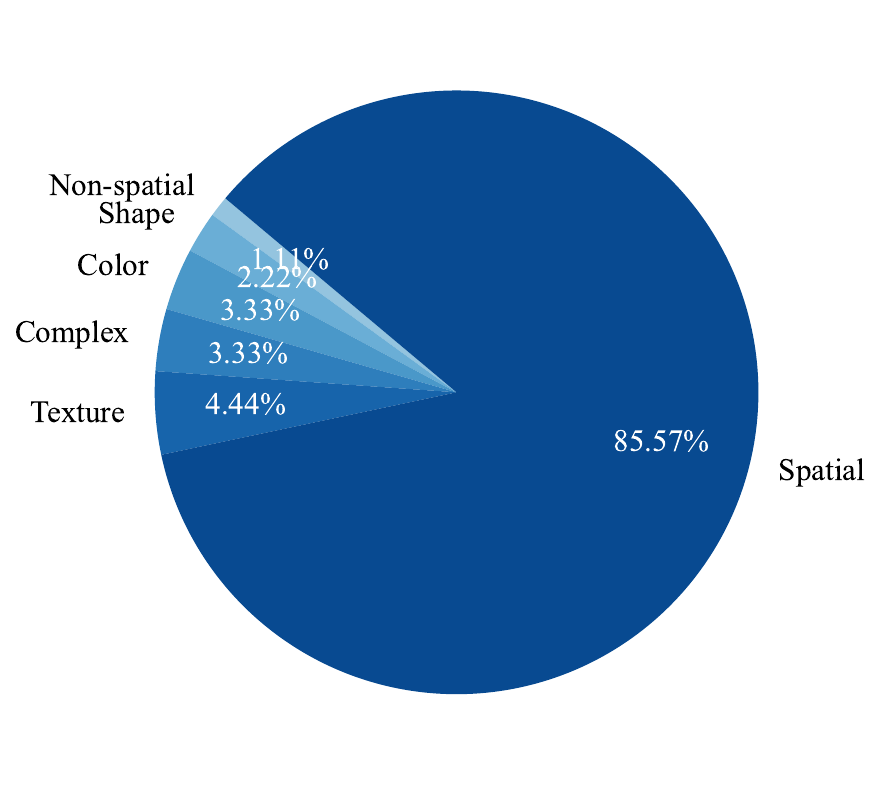}
     \label{fig:sft_showo_train}
  }
\vspace{-0.2in}
\caption{Building self-improvement data based on internal gaps yields more samples from large gap tasks, thus guiding more gains on such tasks.}
\vspace{-0.3in}
\label{fig:training_distribution}
\end{wrapfigure}
Additionally, MLLMs with larger internal gaps (e.g., Janus-Pro-7B) and larger gap subtasks (e.g., \texttt{Texture}) exhibit greater gains after self-improvement. We claim that this may be because tasks with larger internal gaps encourage more samples from those subtasks in the post-training data, thereby benefiting the learning of those specific subtasks. \Cref{fig:training_distribution} demonstrates that subtasks with larger internal gaps constitute a higher proportion of the post-training data, which contributes to their greater performance gains, supporting our hypothesis.

\paragraph{Improved Understanding: Additional Results on Understanding Benchmarks and Examples} For image understanding evaluation, we consider the benchmarks POPE \citep{li2023evaluatingobjecthallucinationlarge}, MMB \citep{liu2024mmbenchmultimodalmodelallaround}, SEED \citep{li2023seedbenchbenchmarkingmultimodalllms}, GQA \citep{hudson2019gqanewdatasetrealworld}, and MMMU, and conduct the evaluation using \href{https://github.com/open-compass/VLMEvalKit}{\texttt{VLMEvalKit}}. Since all these benchmarks are in a multiple-choice format, we compute accuracy using exact matching. \Cref{tab:janus_und} presents the results of the pre-trained Janus-Pro and the self-improved Janus-Pro on various understanding benchmarks, showing that the MLLM’s understanding ability is further enhanced after self-improvement, with gains up to 3\%. 

\begin{figure}[h]
  \centering
  \includegraphics[width=1\textwidth, height=0.28\textheight]{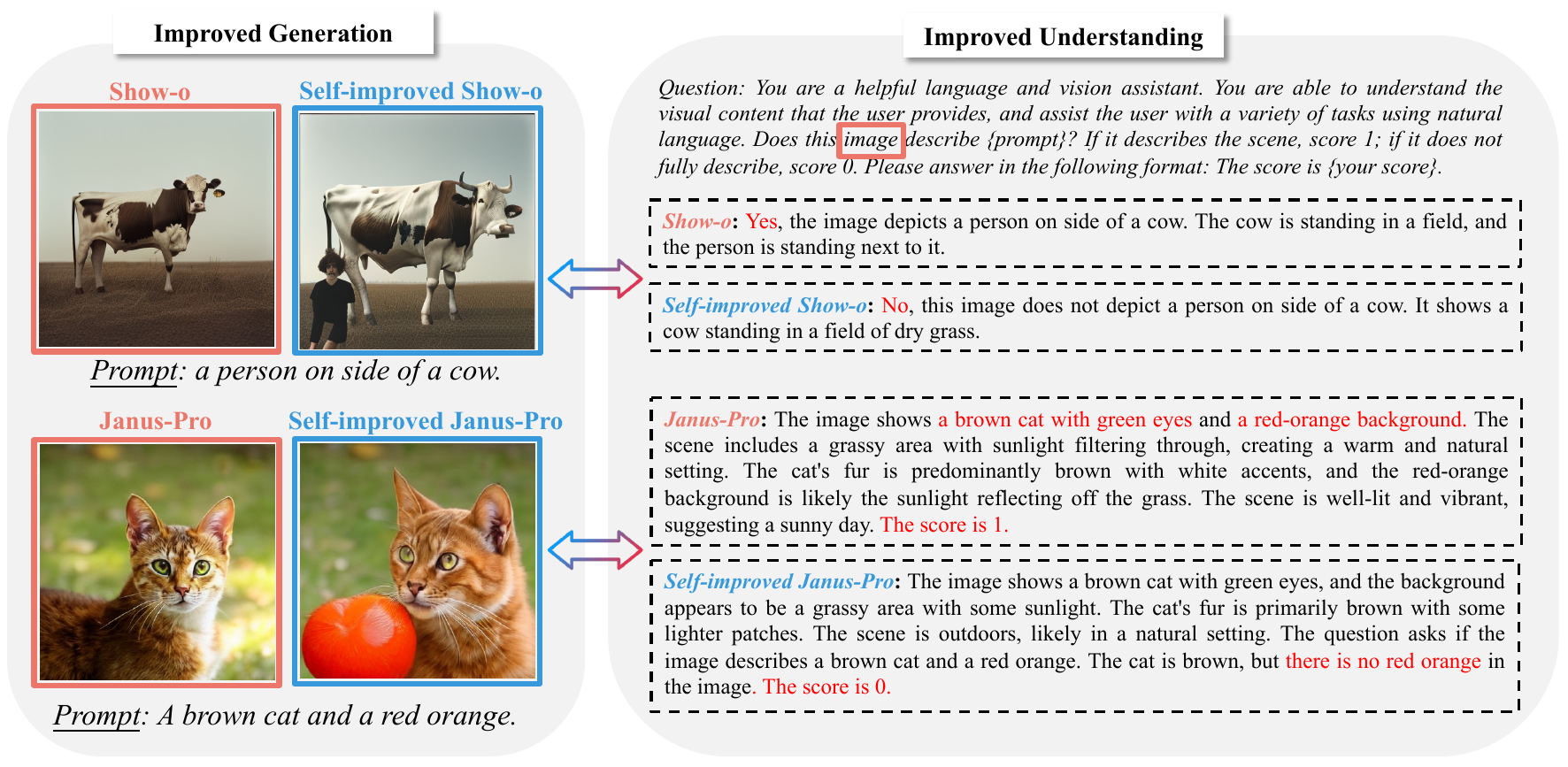}
\vspace{-0.3in}
  \caption{Examples of co-improvements in generation and understanding of self‑improved Janus-Pro and Show-o under SFT. We observe that, after self‑improvement, Show-o and Janus-Pro generate images that align prompts and accurately identify when images produced by the pre‑trained MLLM are misaligned with the prompts.}
\vspace{-0.2in}
  \label{fig:example_Generation-Understanding_SFT}
\end{figure}

\begin{figure}[h]
  \centering
  \includegraphics[width=1\textwidth, height=0.28\textheight]{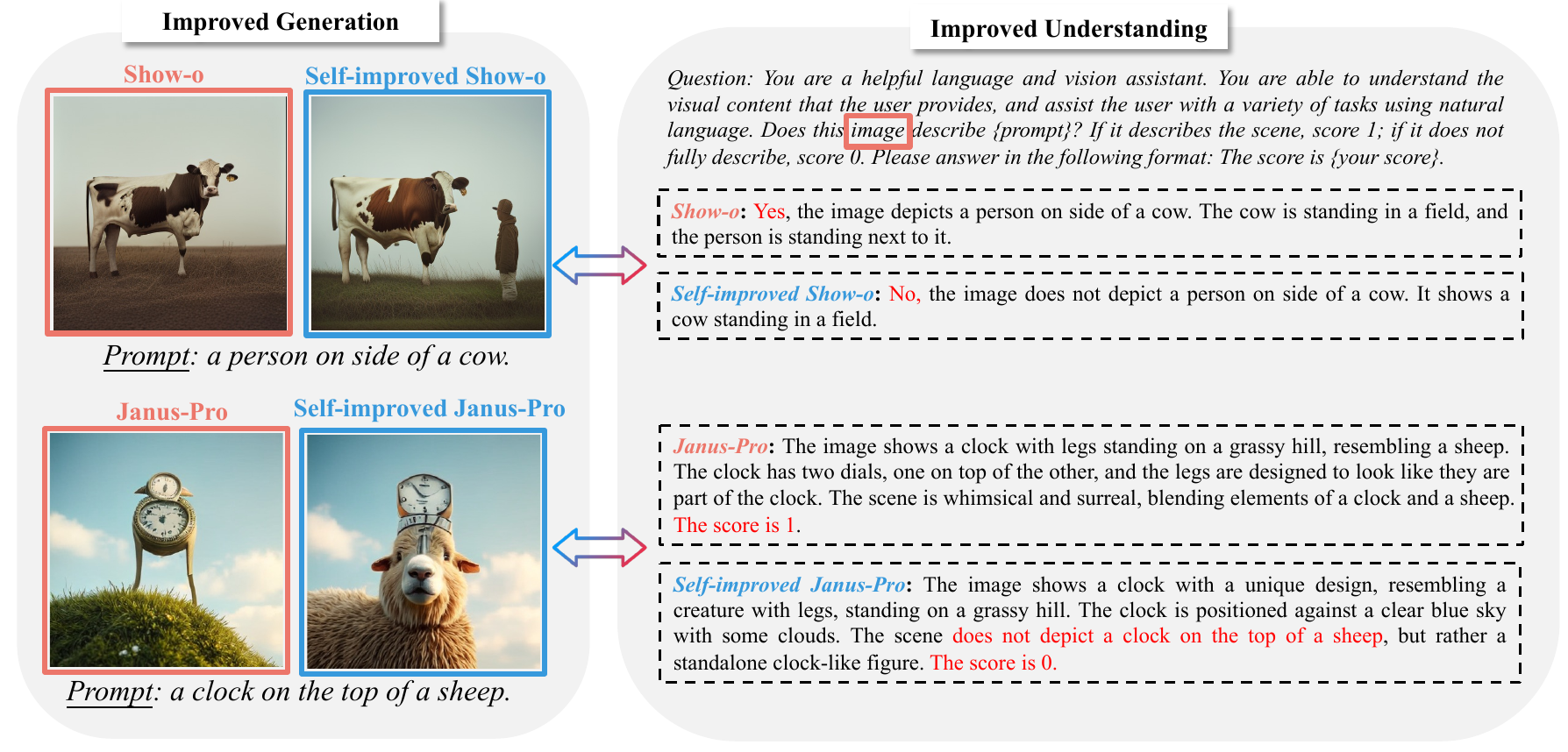}
\vspace{-0.3in}
  \caption{Examples of co-improvements in generation and understanding of self‑improved Janus-Pro and Show-o with DPO. We observe that, after self‑improvement, Show-o and Janus-Pro generate images that align prompts and accurately identify when images produced by the pre‑trained MLLM are misaligned with the prompts.}
\vspace{-0.2in}
  \label{fig:example_Generation-Understanding_DPO}
\end{figure}

We further present examples of self-improvement for Janus-Pro and Show-o under SFT (\Cref{fig:example_Generation-Understanding_SFT}) and DPO (\Cref{fig:example_Generation-Understanding_DPO}), which clearly demonstrate that after self-improvement, the models not only generate images that better satisfy the prompts, but also more accurately identify misalignments between the original image and the prompt, thereby providing correct evaluation scores (from score 1 to score 0). The improvements observed on understanding benchmarks, together with these concrete examples, further support the co-improvement conclusion in \Cref{sec:Results}: generation‑targeted self‑improvement can also enhance understanding.
\begin{table*}[h]
\centering
\vspace{-0.1in}
\begin{tabular}{@{}lcccccccccc@{}}
\toprule
{\pz\pz\pz Model} & POPE$\uparrow$ & MMB $\uparrow$ & SEED $\uparrow$& GQA$\uparrow$& MMMU $\uparrow$\\
\midrule
Janus-Pro-7B & 89.04 &76.23 & 70.09 & 56.02 & 32.86   \\ 
\pz\pz + \textit{SFT} &88.45 &  \underline{76.97} & {70.44} &56.12& \textbf{35.24}\\ 
\pz\pz + \textit{DPO} & \underline{89.06}  &  76.41 &  70.10 & \textbf{56.26}&  33.71   \\ 
\pz\pz + \textit{C-SFT} & 89.03  & \textbf{77.18} &  \underline{70.48}  &56.02 & \textbf{35.24}\ \\ 
\pz\pz + \textit{C-DPO} & \textbf{89.10} & 76.47 & \textbf{70.86} &\underline{56.17} &\underline{34.33}  \\
\bottomrule
\end{tabular}
\caption{The self-improved MLLMs demonstrated improvements on understanding benchmarks.}
\label{tab:janus_und}
\vspace{-0.2in}
\end{table*}


\section{Additional Details and Full Results on Curriculum-Learning-Based Self-Improvement}
\label{app:Details on Curriculum-Learning-Based Self-Improvement}
In this section, we present the training details and full experimental results of the curriculum learning–based self‑improvement method.
\subsection{Additional Details}
\label{app:Additional Details on Curriculum-Learning-Based Self-Improvement}

\paragraph{Implementation Details.}Following \Cref{sec:Curriculum Learning for Stronger Self-Improvement}, we leverage the improved generation–understanding model to revisit prompts that were not utilized by the pre‑trained MLLM due to weak generation or weak understanding capabilities (see details in Alg~\ref{alg:curriculum}). This process can be regarded as a form of curriculum learning based on prompt complexity \citep{li20252dcurridpotwodimensionalcurriculumlearning}. We follow the training configurations in \Cref{tab:param} and perform curriculum replay for both SFT‑based and DPO‑based self‑improvement at epoch 10. In \Cref{app:Curriculum Learning Parameters}, we conduct an ablation study to discuss the choice of epoch for curriculum learning. \Cref{tab:cl_data} shows the data expansion for Janus‑Pro and Show‑o with curriculum learning, which increases sample size by up to 50\%.
\begin{table}[h]
\centering
\resizebox{0.85\linewidth}{!}{
\begin{tabular}{c c c c c}
\toprule
\textbf{MLLM} & \textbf{Self-improvement Strategy} & \textbf{Curriculum Epoch} & \textbf{Original Data} & \textbf{Expansion Data} \\
\midrule
\multirow{2}{*}{Janus-Pro-7B} 
& SFT & 10 & \multirow{2}{*}{2265} & +1091 \\
& DPO & 10 &  &  +359 \\
\midrule
\multirow{2}{*}{Show-o} 
& SFT & 10 & \multirow{2}{*}{226} & +64 \\
& DPO & 10 &  & +59\\
\bottomrule
\end{tabular}
}
\caption{Expansion of post-training data with introducing curriculum learning.}
\label{tab:cl_data}
\end{table}

\paragraph{Evaluation.}Consistent with the evaluation in \Cref{app:Additional Details on Self-Improvement}, we employ the same metrics to measure MLLMs in generation, understanding and unification.

\subsection{Full Results}
\label{app:Full Results on Curriculum-Learning-Based Self-Improvement}
As shown in \Cref{tab:t2i}, \Cref{tab:geneval}, and \Cref{tab:science}, the self-improvement with curriculum learning (denoted as \textit{C-SFT} and \textit{C-DPO}) demonstrates that the increased post-training data benefiting from curriculum learning further enhances self-improvement MLLMs' performance and unification, particularly in understanding and generation.





\section{Understanding Co-improvement in Self-Improvement}
\label{app:Learning Dynamics of Generation and Understanding}
\Cref{sec:Empirical Verification} explains why co-improvement occurs when self-improvement is performed with SFT and provides empirical evidence based on Janus-Pro. In this section, we first detail the computation of the empirical evidence in \Cref{fig:evidence}, then additionally present empirical evidence on Show-o with SFT to further support the theoretical analysis in \Cref{sec:Learning Dynamics of Generation and Understanding}.
\subsection{Full Theoretical Analysis under SFT}
\label{app:sft}

\paragraph{Details on Empirical Evidence.}\Cref{fig:evidence}(a) explains that samples from the false positive correction group $(\mathbf{y}_0 ,\mathbf{x}_0 )$, i.e., the primary source of improvement in comprehension capability, exhibit higher similarity to their corresponding post-training samples $(\mathbf{y}_u ,\mathbf{x}_u )$. Specifically, we separately compute text similarity and image similarity as proxies for eNTK term $\mathcal{K}$: for each $\mathbf{y}_0$, we first identify its nearest neighbor $\mathbf{y}_u$ in the post-training data, then compute the similarity between the corresponding images $\mathbf{x}_0$ and $\mathbf{x}_u$ . For text similarity, we use pre-trained model all-MiniLM-L6-v2 \footnote{https://huggingface.co/sentence-transformers/all-MiniLM-L6-v2} to encode each prompt into a 384-dimensional vector and compute the cosine similarity between vector pairs. For image similarity, we use an equal-weighted combination of MSE and SSIM \citep{wang2004image} to measure both pixel-level and structural similarity. To evaluate whether false positive correction group indeed exhibits higher similarity, we randomly sample \textit{random group} $(\mathbf{y} ,\mathbf{x})$ (with the same size as false positive correction group) and calculate same. \Cref{fig:evidence}(a) shows false positive correction group demonstrates significantly higher similarity in $(\mathbf{y}_u ,\mathbf{x}_u )$, particularly in text.

\Cref{fig:evidence}(b) shows the Frobenius norm $\bigl\|\textcolor{lightblue}{\mathcal{K}^t(\mathcal{Y}_0,\mathcal{Y}_u)}\bigr\|_F$ exceeds $\bigl\|\textcolor{lightred}{\mathcal{K}^t(\mathcal{Y}_i,\mathcal{Y}_u)}\bigr\|_F$. This indicates that at iteration $t$, the training dynamics of the understanding branch are primarily driven by \textcolor{lightblue}{$\mathcal{K}^t(\mathcal{Y}_0,\mathcal{Y}_u)$}, which tends to align $\Delta U_t$ in \Cref{eq:und-main} and $\Delta G_t$ in \Cref{eq:gen-main}. To substantiate this, we use data similarity as a proxy for the eNTK. Specifically, for each sample $(\mathbf{y}_0,\mathbf{x}_0)$ in the false positive correction group, we first identify its closest $(\mathbf{y}_u,\mathbf{x}_u)$ based on the most similar prompt and compute text and image similarities using the same metrics as in \Cref{fig:evidence}(a); this serves as the proxy for $\bigl\|\textcolor{lightblue}{\mathcal{K}^t(\mathcal{Y}_0,\mathcal{Y}_u)}\bigr\|_F$. For $\bigl\|\textcolor{lightred}{\mathcal{K}^t(\mathcal{Y}_i,\mathcal{Y}_u)}\bigr\|_F$, we compute the text and image similarity between each non-$(\mathbf{y}_0,\mathbf{x}_0)$ sample $(\mathbf{y}_i,\mathbf{x}_i)$ and $(\mathbf{y}_u,\mathbf{x}_u)$, and average these similarities over all $(\mathbf{y}_i,\mathbf{x}_i)$ as the proxy.

\Cref{fig:evidence}(c) shows, for samples in the false positive correction group, the probability of prompt-misaligned generation, i.e., the prompt-misaligned probability $\pi_{\theta}(\mathbf{x}_0 \mid \mathbf{y}_0)$ decreases. To quantify this change, for each validation prompt $\mathbf{y}_0$, we first use the pre-trained MLLM to generate $\mathbf{x}_0$ and record its image token sequence and the log-probability of that sequence as $\log \pi_{\theta_0}(\mathbf{x}_0 \mid \mathbf{y}_0)$. We then use the self-improved MLLMs to re-evaluate the conditional log-probability of the same token sequence, obtaining $\log \pi_{\theta_t}(\mathbf{x}_0 \mid \mathbf{y}_0)$. Following the definition of the generation-branch learning dynamics in \Cref{sec:Learning Dynamics of Generation and Understanding}, we compute
$\Delta G_t = \log \pi_{\theta_t}(\mathbf{x}_0 \mid \mathbf{y}_0) - \log \pi_{\theta_0}(\mathbf{x}_0 \mid \mathbf{y}_0)$.

\begin{figure}[h]
\vspace{-0.1in}
\centering
    \hfill
    \subfigure[\scriptsize Dominant False Positive Correction]{\label{fig:showo-und-score}\includegraphics[width=0.235\linewidth]{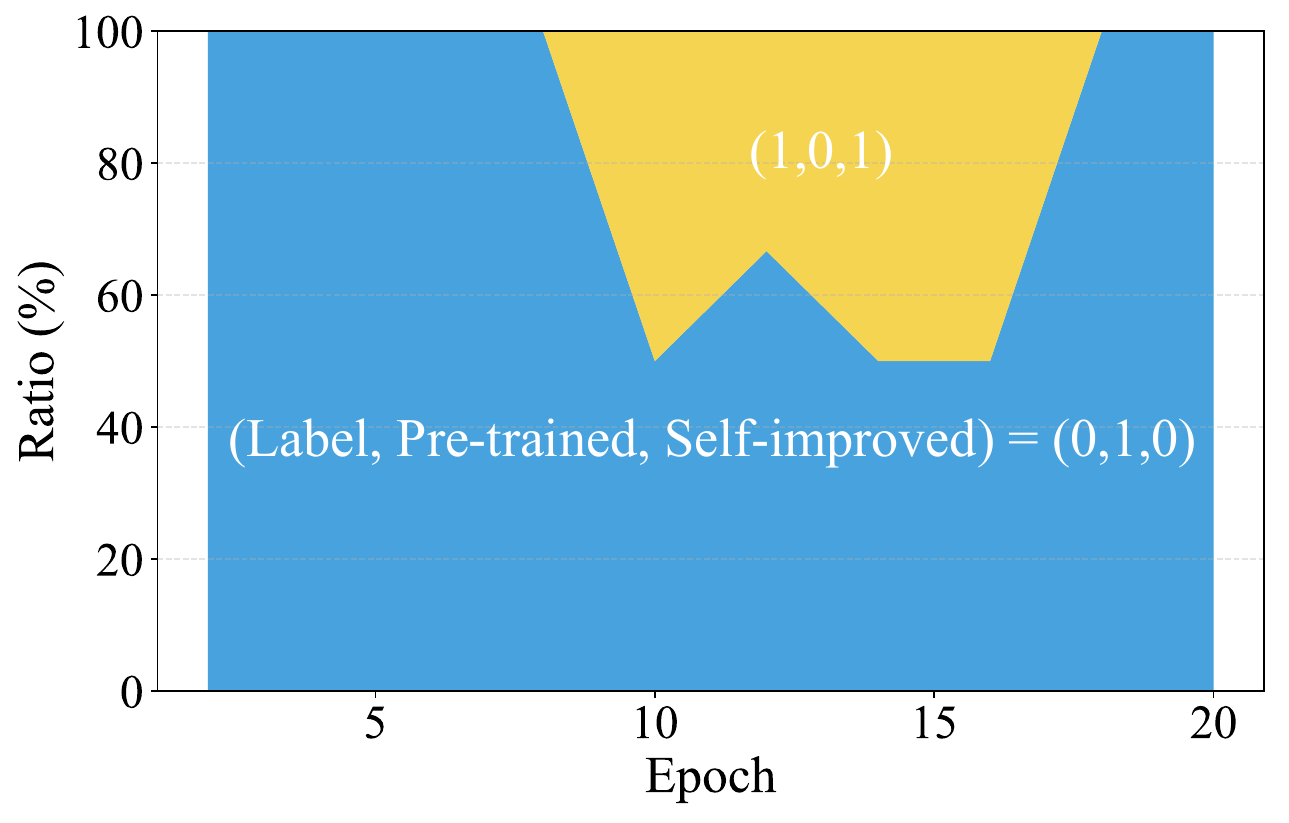}}
    \hfill
    \subfigure[\scriptsize Similar $\mathcal{Y}_0$ and $\mathcal{Y}_u$]{\label{fig:showo-rank-}\includegraphics[width=0.245\linewidth]{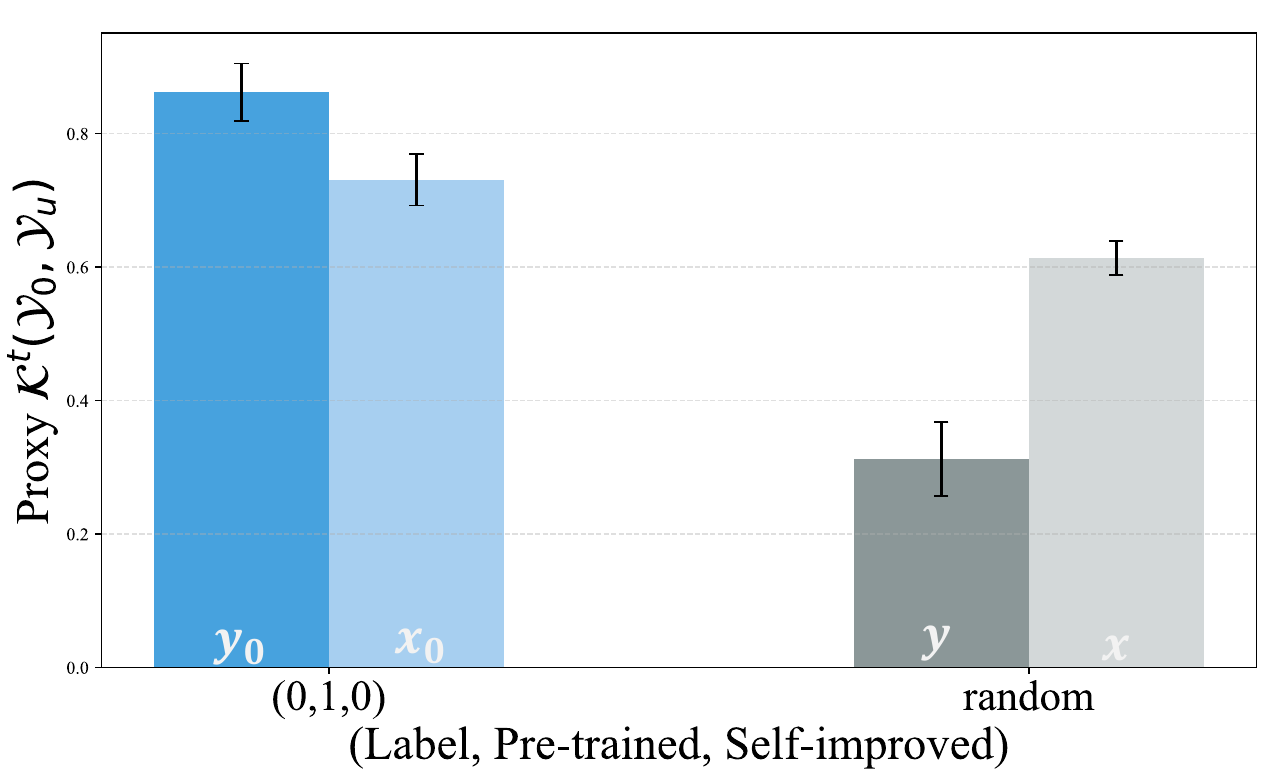}}
    \hfill
    \subfigure[\scriptsize Proxy $\|\textcolor{lightblue}{\mathcal{K}^t(\mathcal{Y}_0,\mathcal{Y}_u)}\|_F > \|\textcolor{lightred}{\mathcal{K}^t(\mathcal{Y}_i,\mathcal{Y}_u)}\|_F$]{\label{fig:showo-chosen-}\includegraphics[width=0.245\linewidth]{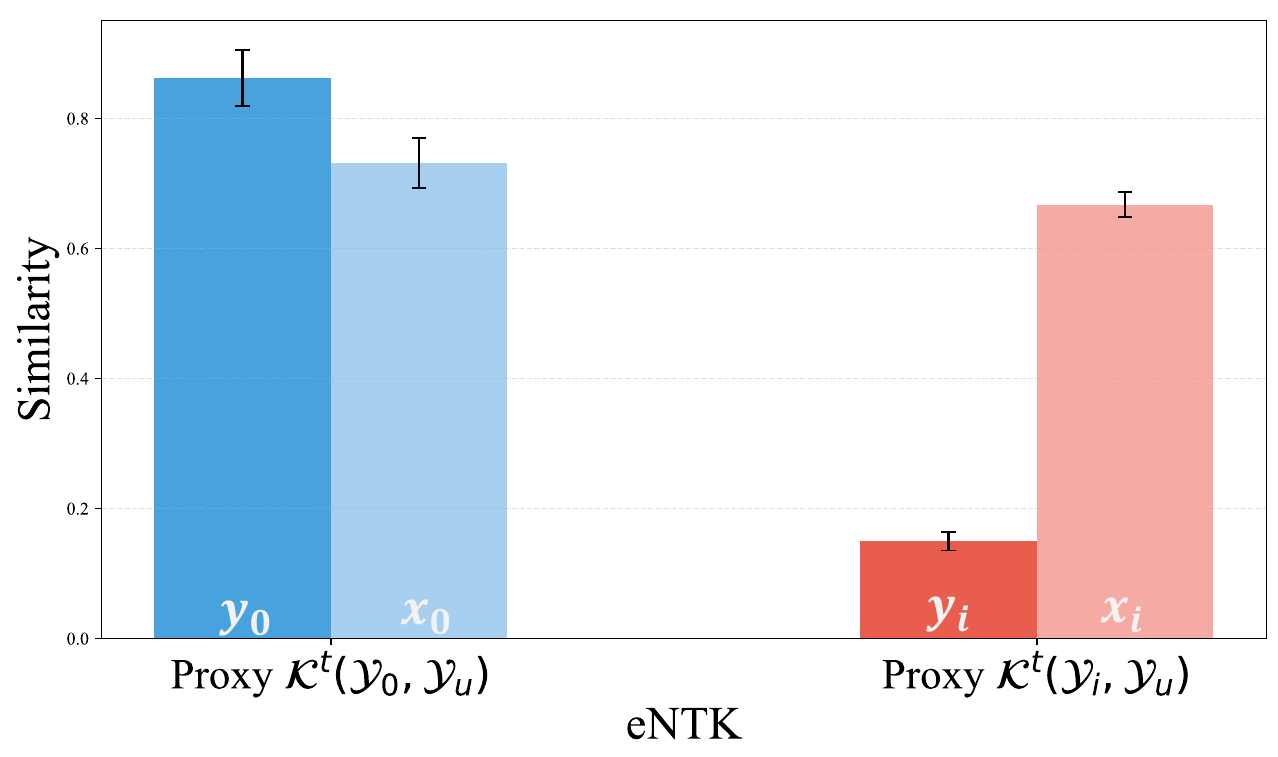}}
    \hfill
    \subfigure[\scriptsize False Positive Correction Group: $\Delta G_t < 0$]{\label{fig:showo-reject}\includegraphics[width=0.245\linewidth]{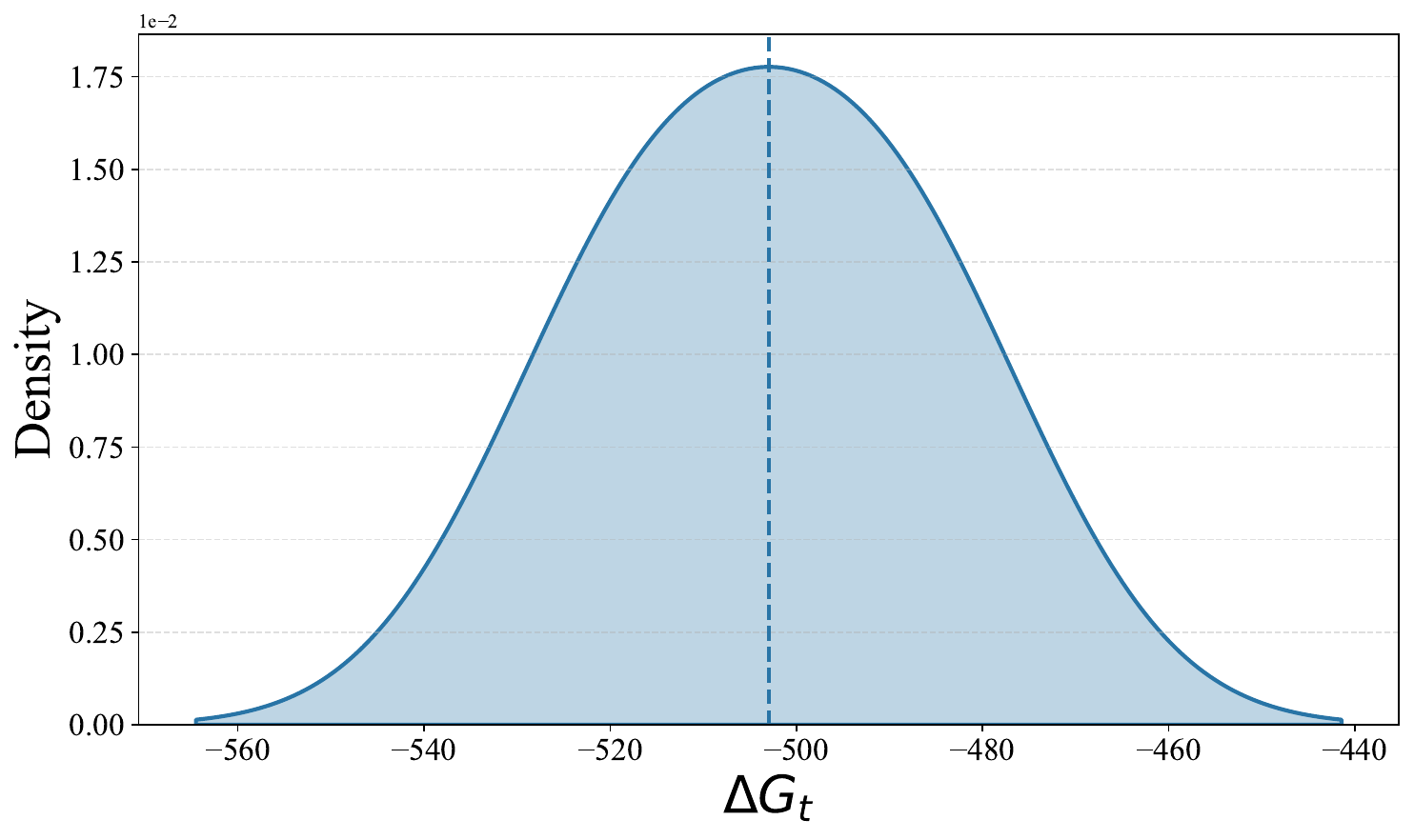}}
    \hfill
\vspace{-0.1in}
\caption{Empirical Evidence from Self‑Improvement with Show-o and SFT. (a) On T2I-CompBench++, understanding gains primarily arise from the false positive correction group. (b) Compared to random samples, those in the false positive correction group are more likely to be matched with highly similar post-training pairs $(\mathbf{y}_u,\mathbf{x}_u)$ (average cosine similarity 0.8). (c) Such high similarity makes $\textcolor{lightblue}{\mathcal{K}^{\,t}_{k,r}(\mathcal{Y}_0,\mathcal{Y}_u)}$ be the dominant term in \Cref{eq:und-main}, thereby promoting aligned learning dynamics between understanding in \Cref{eq:und-main} and generation in \Cref{eq:gen-main}. (d) With aligned dynamics, $\Delta G_t < 0$ implies $\Delta U_t < 0$: both the probability of misaligned generation $\pi_{\theta}(\mathbf{x}_0 \mid \mathbf{y}_0)$ and misjudging  $\pi_{\theta}(\mathbf{y}_0 \mid \mathbf{x}_0)$, are reduced. This manifests as false positive correction and jointly as co-improvement.}
\vspace{-0.15in}
\label{fig:sft-showo}
\end{figure}

\paragraph{More Empirical Evidence on Show-o.} \Cref{sec:Empirical Verification} explains why co-improvement occurs when post-self-improvement is performed with SFT and provides empirical evidence based on Janus-Pro. In this section, we additionally present empirical evidence on Show-o with SFT to further support the theorical analysis in \Cref{sec:Learning Dynamics of Generation and Understanding}. \Cref{fig:sft-showo}(a) shows that, for Show-o under supervised fine-tuning (SFT), the primary gains in understanding still come from false positive correction, i.e., (Label, Pre-trained, Self-improved) = (0, 1, 0). Moreover, there exists post-training data similar to the false positive correction group, with an average cosine similarity of 0.8 (see \Cref{fig:sft-showo}(b)). \Cref{fig:sft-showo}(c) indicates that the high sample similarity makes $\bigl\|\textcolor{lightblue}{\mathcal{K}^t(\mathcal{Y}_0,\mathcal{Y}_u)}\bigr\|_F$ the dominant term, encouraging alignment between the training dynamics of generation and understanding. Together with \Cref{fig:sft-showo}(d), which shows $\Delta G_t < 0$, this suggests $\Delta U_t < 0$, meaning the model identifies false positives and achieves joint improvement. Empirical evidence for Show-o under SFT further corroborates the theoretical explanation in \Cref{sec:Empirical Verification}.

\subsection{Full Theoretical Analysis under DPO}
\label{app:dpo}
\paragraph{Proposition.}In this section, we discuss why DPO-based self-improvement also leads to co-improvement (see \Cref{tab:t2i}, \Cref{tab:geneval}, \Cref{tab:science}, and \Cref{tab:janus_und}). For DPO, we define a post-training preference pair $(\mathbf{y}_u, \mathbf{x}^+_u, \mathbf{x}^-_u)$
where the chosen image $\mathbf{x}^+_u$ and the rejected image $\mathbf{x}^-_u$ share the same prompt $\mathbf{y}_u$. The DPO loss is
\begin{equation}
\mathcal{L}_{\mathrm{DPO}}(\mathbf{y}_u, \mathbf{x}_u^+, \mathbf{x}_u^-) 
= - \mathbb{E}_{(\mathbf{y}_u, \mathbf{x}_u^+, \mathbf{x}_u^-) }
\left[
    \log \sigma \left(
        \beta \log \frac{\pi_\theta(\mathbf{x}_u^+ \mid \mathcal{Y}_u^+)}{\pi_{\mathrm{ref}}(\mathbf{x}_u^+ \mid \mathcal{Y}_u^+)}
        - \beta \log \frac{\pi_\theta(\mathbf{x}_u^- \mid \mathcal{Y}_u^-)}{\pi_{\mathrm{ref}}(\mathbf{x}_u^- \mid \mathcal{Y}_u^-)}
    \right)
\right],
\end{equation}
where $\mathcal{Y}_u^+$ denotes the concatenation obtained by appending the embedding of $\mathbf{y}_u$ to the embedding of $\mathbf{x}_u^+$, and $\mathcal{Y}_u^-$ denotes the concatenation obtained by appending the embedding of $\mathbf{y}_u$ to the embedding of $\mathbf{x}_u^-$. Then, we have the following proposition:

\begin{proposition}
[Learning Dynamics of Generation and Understanding under DPO]
\label{theorem:interpaly-dpo}
    Consider self-improvement proposed in \Cref{sec:Mitigating Non-Unification: A Self-Improvement Framework} with DPO. 
    
At epoch $t$, the one-step learning dynamics of \textbf{generation} is
\begin{align}
  \label{eq:gen-main-dpo}
 &\Delta G_t(\mathbf{x}_0\mid\mathcal{Y}_0) \notag\\
    &= -\eta \beta \sigma(-\alpha) \sum_{k=1}^{M}\sum_{r=1}^{M} (\mathbf{e}_{x_{0,k}}-{\pi}^{0}_{k})^{\top} \Big[ \mathcal{K}^t_{k,r}(\mathcal{Y}_0,\mathcal{Y}^+_u)({\pi}^{u,+}_{r}-\mathbf{e}_{x^+_{u,r}})-\mathcal{K}^t_{k,r}(\mathcal{Y}_0,\mathcal{Y}^-_u)({\pi}^{u,-}_{r}-\mathbf{e}_{x^-_{u,r}})\Big]\notag\\
    &\quad + \mathcal{O}(\eta^2),
\end{align}
where the margin $\alpha \coloneqq  \beta \log \frac{\pi_\theta(\mathbf{x}_u^+ \mid \mathcal{Y}_u^+)}{\pi_{\mathrm{ref}}(\mathbf{x}_u^+ \mid \mathcal{Y}_u^+)}
        - \beta \log \frac{\pi_\theta(\mathbf{x}_u^- \mid \mathcal{Y}_u^-)}{\pi_{\mathrm{ref}}(\mathbf{x}_u^- \mid \mathcal{Y}_u^-)}$ and ${\pi}^{u,+}_{r}=\mathrm{softmax}(\mathbf{z}^{u,+}_{r})$ and $\mathbf{z}^{u,+}_{r}=[h_{\theta}(\mathcal{Y}^+_u)]_r$ are the logits at image position $r$ obtained by running $h_{\theta}$ on $\mathcal{Y}^+_u$. The neural tangent kernel $\mathcal{K}^t_{k,r}(\mathcal{Y}_0,\mathcal{Y}^+_u) \coloneqq \nabla_{\theta}\mathbf{z}^{0}_{k}(\nabla_{\theta}\mathbf{z}^{u,+}_{r})^{\!\top}$
and $\mathcal{K}^t_{k,r}(\mathcal{Y}_0,\mathcal{Y}^-_u) \coloneqq \nabla_{\theta}\mathbf{z}^{0}_{k}(\nabla_{\theta}\mathbf{z}^{u,-}_{r})^{\!\top}$.

 The one-step learning dynamics of \textbf{understanding} is
\begin{equation}
\label{eq:und-main-dpo}
\resizebox{\linewidth}{!}{$
\begin{aligned}
&\Delta U_t(\mathbf{y}_0\mid\mathcal{X}_0)\\
&= -\eta \beta \sigma(-\alpha) \sum^M_{k=1}\sum_{r=1}^{M}\sum_{\mathbf y_i\neq \mathbf y_0} w_{\theta_t}(\mathbf y_i\mid \mathbf x_0)\,
\Bigg((\mathbf{e}_{x_{0,k}}-{\pi}^{0}_{k})^{\!\top} \underbrace{\Big(
\mathcal{K}^{\,t}_{k,r}(\mathcal{Y}_0,\mathcal{Y}^+_u)({\pi}^{u,+}_{r}-\mathbf{e}_{x^+_{u,r}})
-
\mathcal{K}^{\,t}_{k,r}(\mathcal{Y}_0,\mathcal{Y}^-_u)({\pi}^{u,-}_{r}-\mathbf{e}_{x^-_{u,r}})
\Big)}_{\textcolor{lightblue}{\textbf{Term I}}}
\\
&\quad -
(\mathbf{e}_{x_{0,k}}-{\pi}^{i}_{k})^{\!\top} \underbrace{\Big(
\mathcal{K}^{\,t}_{k,r}(\mathcal{Y}_i,\mathcal{Y}^+_u)({\pi}^{u,+}_{r}-\mathbf{e}_{x^+_{u,r}})
-
\mathcal{K}^{\,t}_{k,r}(\mathcal{Y}_i,\mathcal{Y}^-_u)({\pi}^{u,-}_{r}-\mathbf{e}_{x^-_{u,r}})
\Big)}_{\textcolor{lightred}{\textbf{Term II}}}\Bigg) + \mathcal{O}(\eta^2)
\end{aligned}
$}
\end{equation}
where $\mathcal{Y}_i$ denote the concatenation obtained by appending the embedding of $\mathbf y_i$ to $\mathbf U_0$.
\end{proposition}

\begin{figure}[h]
\centering
    \hfill
    \subfigure[\scriptsize Dominant False Positive Correction, Janus-Pro]{\label{fig:janus-und-score}\includegraphics[width=0.235\linewidth]{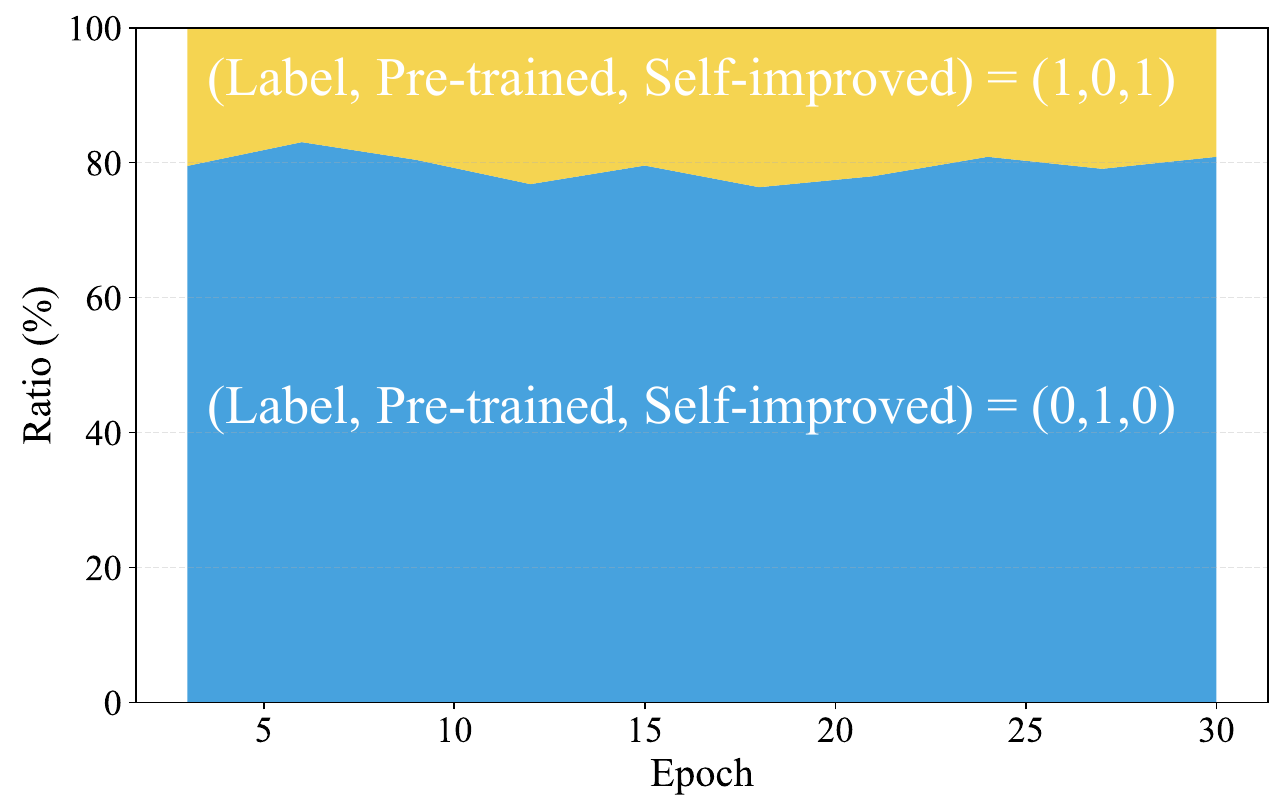}}
    \hfill
    \subfigure[\scriptsize Dominant False Positive Correction, Show-o]{\label{fig:showo-und-score-}\includegraphics[width=0.235\linewidth]{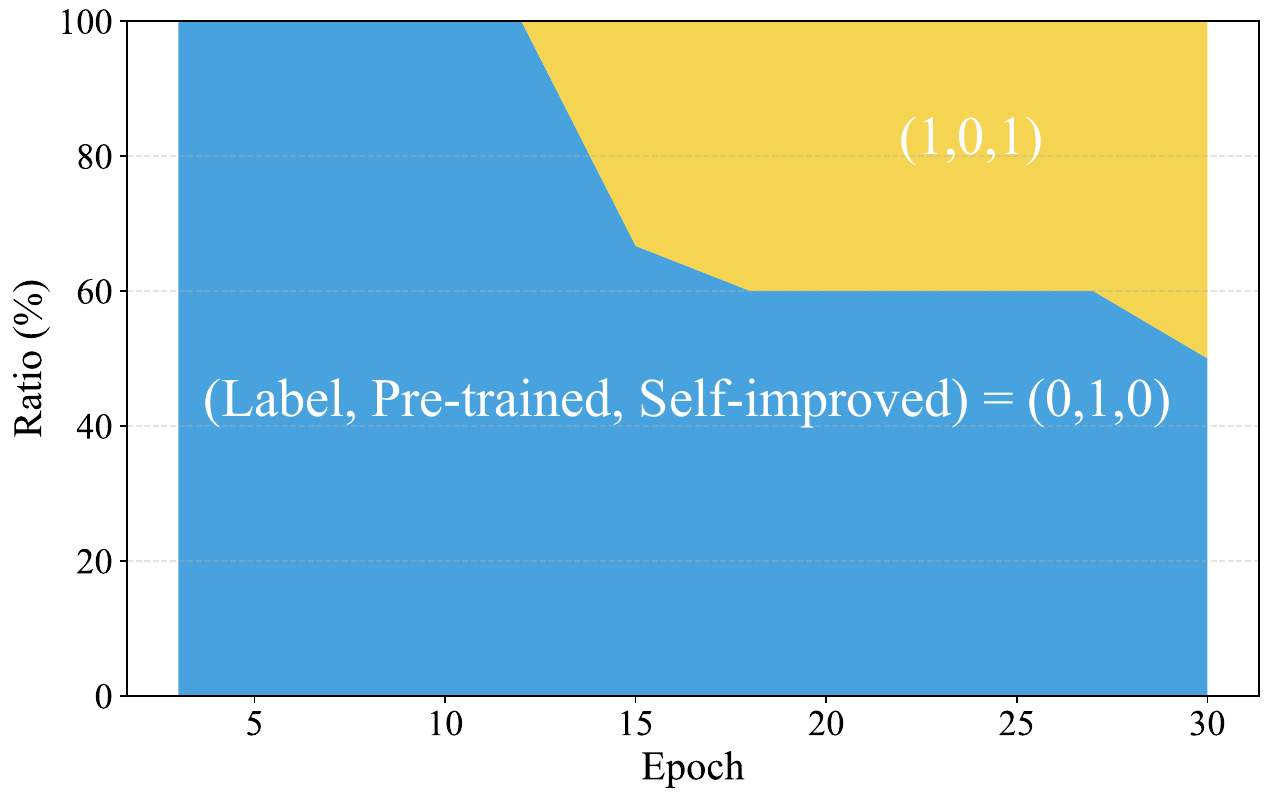}}
    \hfill
    \subfigure[\scriptsize False Positive Correction Group for Janus-Pro: $\Delta G_t < 0$]{\label{fig:janus-reject}\includegraphics[width=0.245\linewidth]{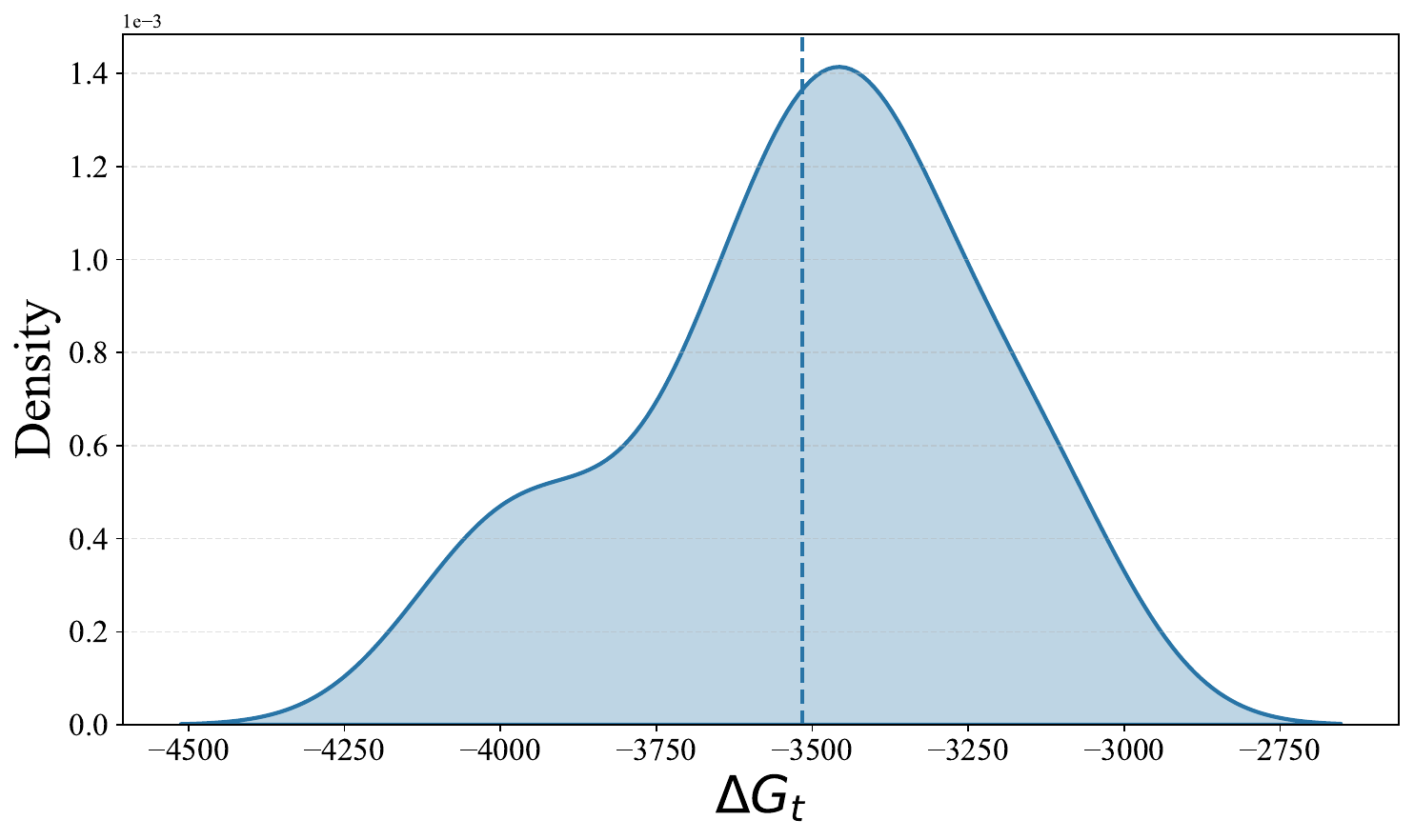}}
    \hfill
        \subfigure[\scriptsize False Positive Correction Group for Show-o: $\Delta G_t < 0$]{\label{fig:showo-reject-}\includegraphics[width=0.245\linewidth]{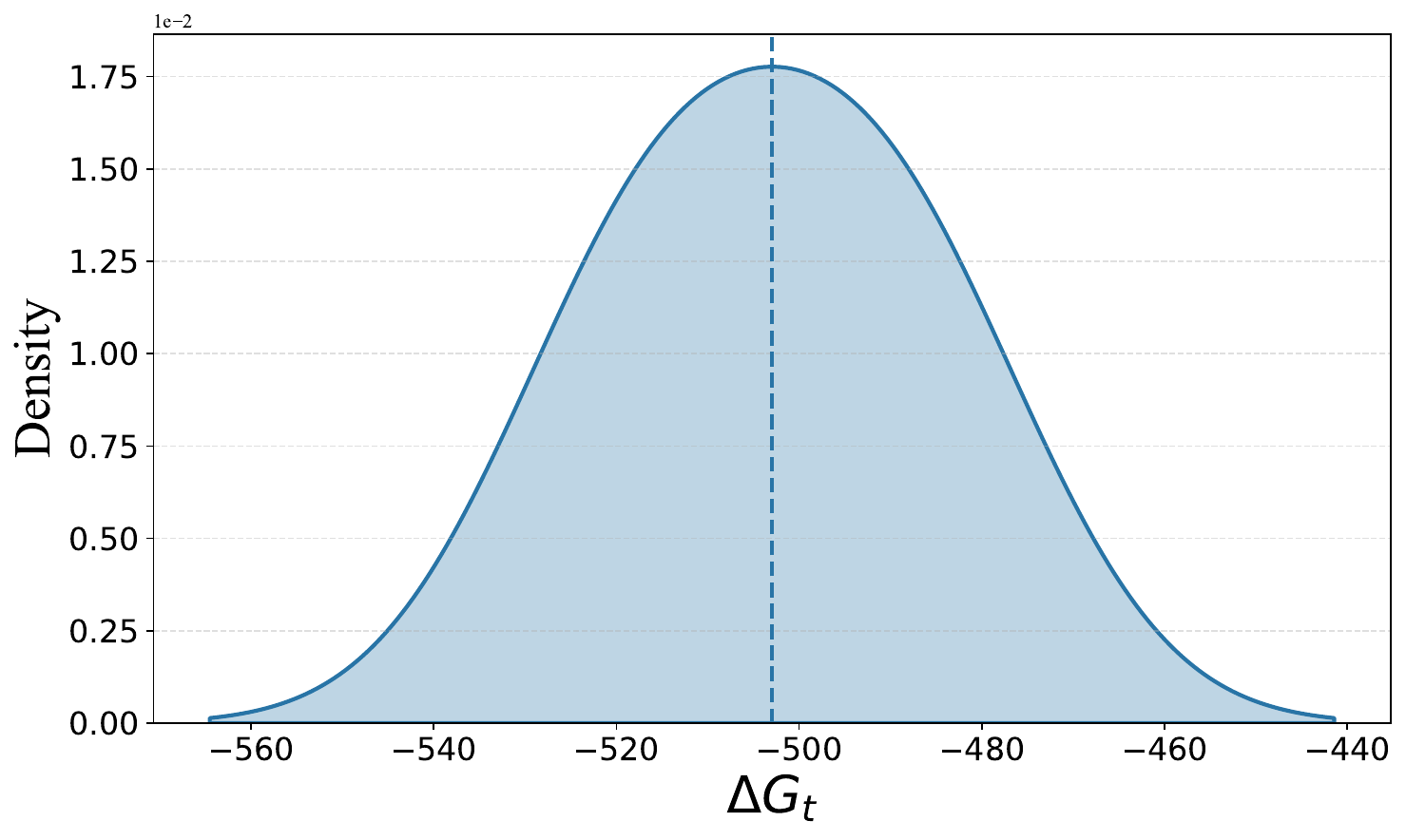}}
     \hfill
\vspace{-0.1in}
\caption{Empirical Evidence from DPO-based Self‑Improvement with Janus-Pro and Show-o. (a)(b) On T2I-CompBench++, understanding gains primarily arise from the false positive correction group. (c)(d) For prompts $\mathbf{y}_0$ in the false positive correction group, the self-improved MLLM also reduces the probability of generating the prompt‑misaligned image $\mathbf{x}_0$, i.e., $\Delta G_t < 0$.}
\vspace{-0.15in}
\label{fig:dpo-evi-1}
\end{figure}

We can interpret \Cref{theorem:interpaly-dpo} by analogy with \Cref{theorem:interpaly}. Specifically, when $\mathcal{Y}_0$ is more similar to the post-training data $\mathcal{Y}_u$ than any other $\mathcal{Y}_i$, that is, the Frobenius norm of \textcolor{lightblue}{Term I} exceeds that of \textcolor{lightred}{Term II}, both the generation and understanding branches are dominated by the same alignment \textcolor{lightblue}{Term I}, yielding consistent update signs.
\paragraph{Theoretical Analysis with Empirical Evidence.} First, \Cref{fig:dpo-evi-1}(a)(b) show that under DPO, gains in understanding still primarily come from correcting false positives: across training steps, this accounts for roughly 60\%–100\% of the gains. Hence, we focus on $\mathbf{y}_0$ and its misaligned image $\mathbf{x}_0$ generated by pre-trained MLLMs.

\begin{figure}[h]
\centering
    \hfill
    \subfigure[\scriptsize Similar $\mathcal{Y}_0$ and $\mathcal{Y}^+_u$]{\label{fig:janus-rank}\includegraphics[width=0.24\linewidth]{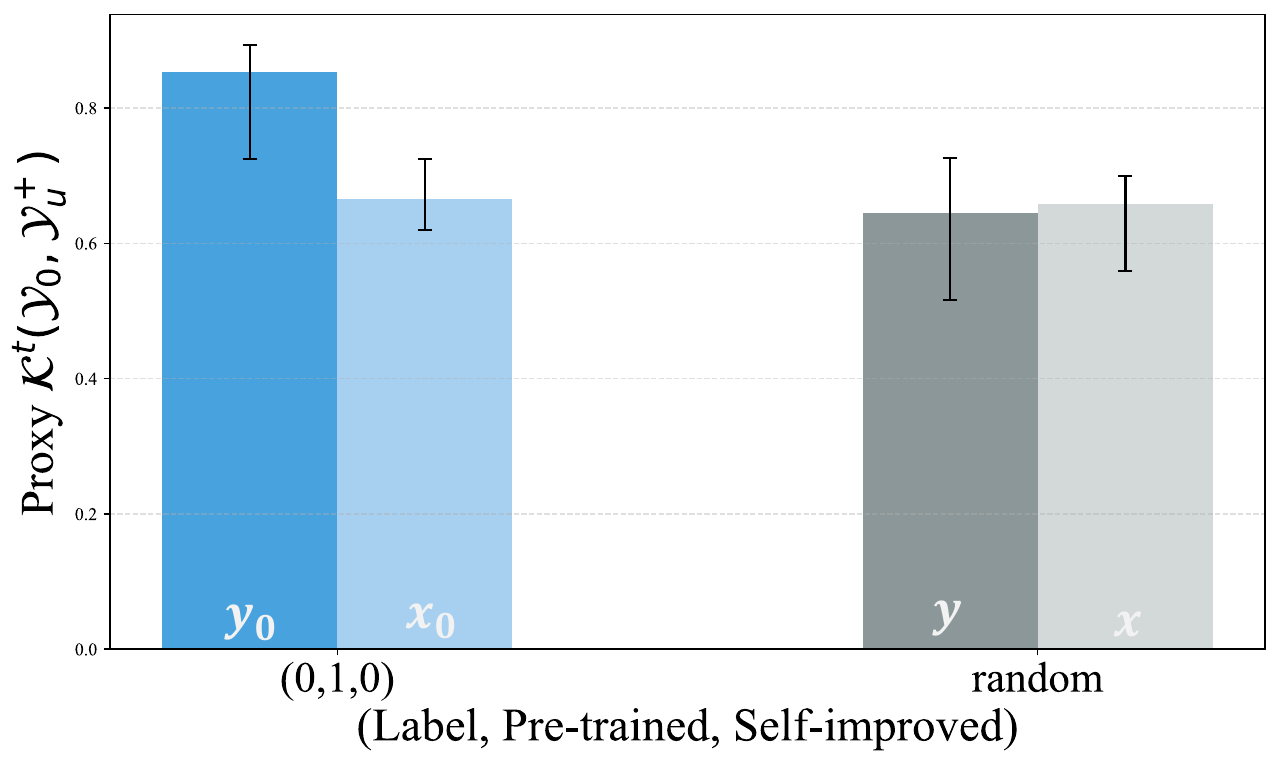}}
    \hfill
    \subfigure[\scriptsize Proxy $\|\textcolor{lightblue}{\mathcal{K}^t(\mathcal{Y}_0,\mathcal{Y}^+_u)}\|_F > \|\textcolor{lightred}{\mathcal{K}^t(\mathcal{Y}_i,\mathcal{Y}^+_u)}\|_F$]{\label{fig:janus-chosen}\includegraphics[width=0.245\linewidth]{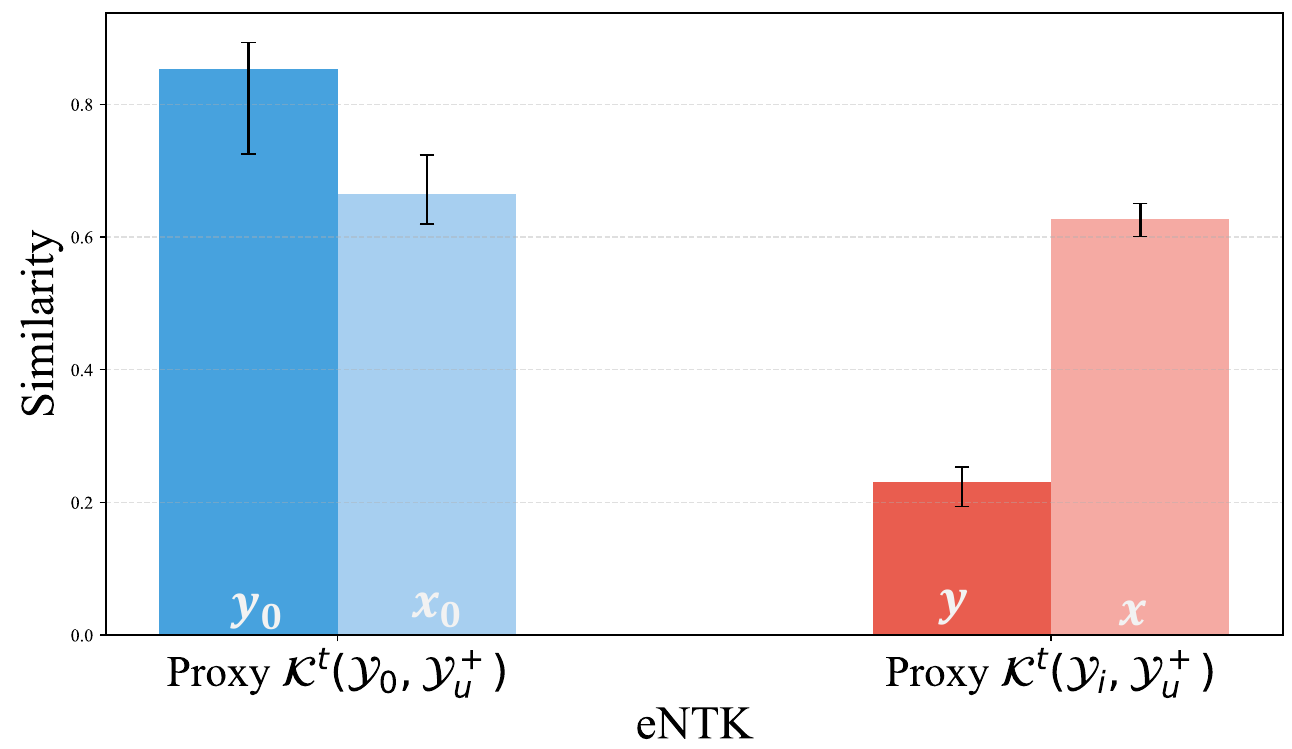}}
    \hfill
    \subfigure[\scriptsize Similar $\mathcal{Y}_0$ and $\mathcal{Y}^-_u$]{\label{fig:janus-rank-}\includegraphics[width=0.24\linewidth]{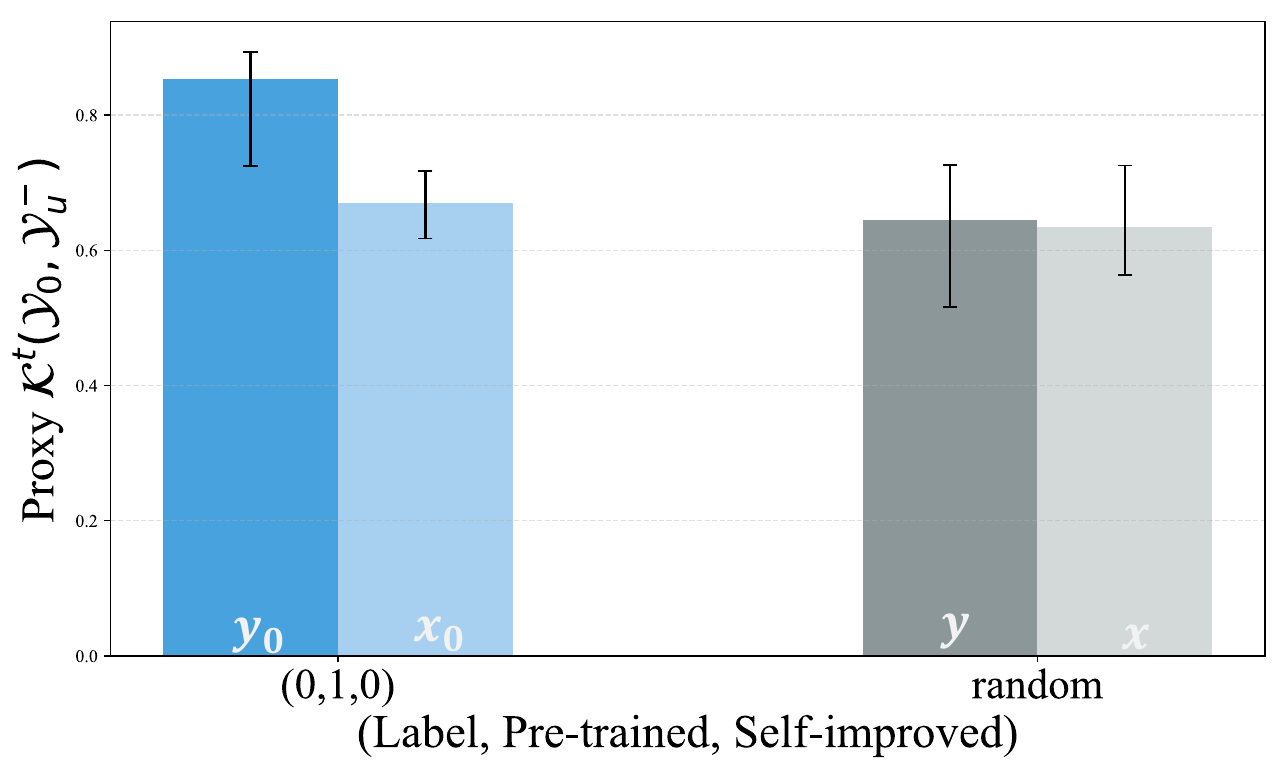}}
    \hfill
    \subfigure[\scriptsize Proxy $\|\textcolor{lightblue}{\mathcal{K}^t(\mathcal{Y}_0,\mathcal{Y}^-_u)}\|_F > \|\textcolor{lightred}{\mathcal{K}^t(\mathcal{Y}_i,\mathcal{Y}^-_u)}\|_F$]{\label{fig:janus-chosen-}\includegraphics[width=0.245\linewidth]{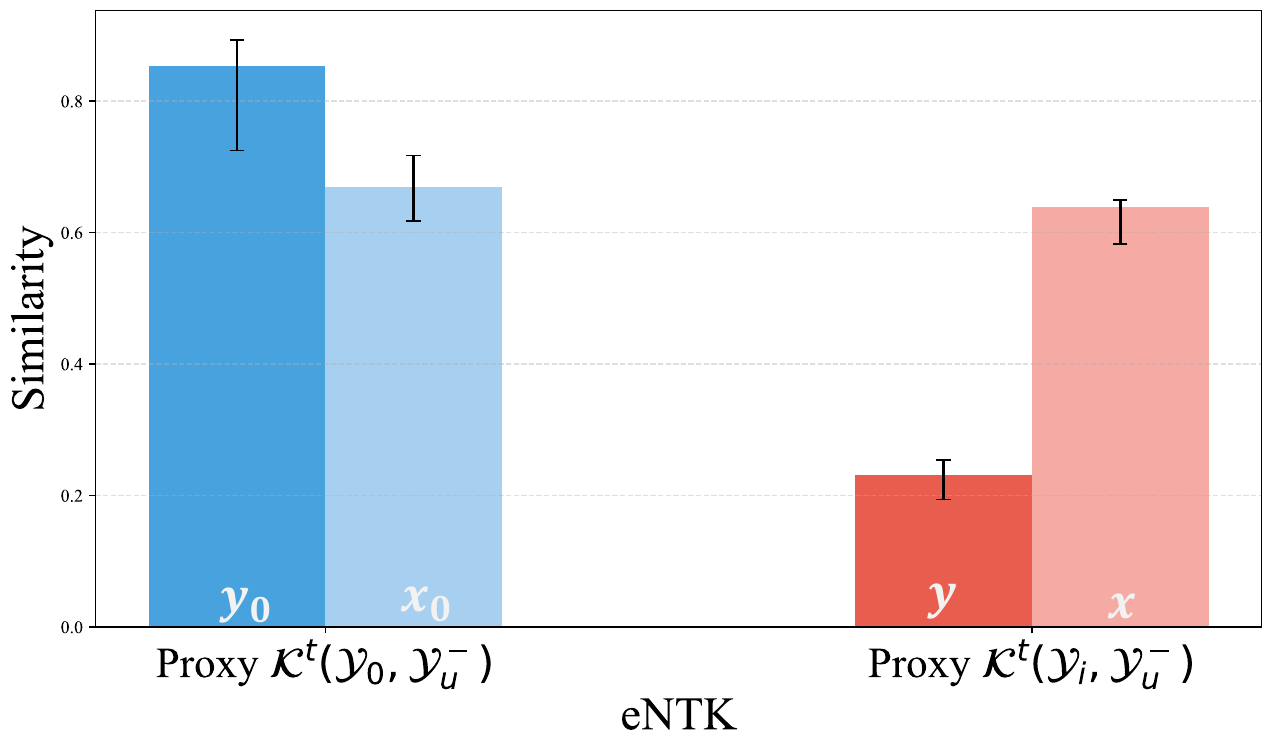}}
     \hfill
\vspace{-0.1in}
\caption{Empirical Evidence from Self‑Improvement with Janus-Pro and DPO. (a)(c) Compared to random samples, those in the false positive correction group are more likely to be matched with highly similar post-training pairs $(\mathbf{y}_u,\mathbf{x}_u)$ (average cosine similarity 0.8). (b)(d) Such high similarity makes $\textcolor{lightblue}{\text{Term I}}$ be the dominant term in \Cref{eq:und-main-dpo}, thereby promoting aligned learning dynamics between understanding in \Cref{eq:und-main-dpo} and generation in \Cref{eq:gen-main-dpo}.}
\vspace{-0.15in}
\label{fig:dpo-evi-2}
\end{figure}

\begin{figure}[h]
\centering
    \hfill
    \subfigure[\scriptsize Similar $\mathcal{Y}_0$ and $\mathcal{Y}^+_u$]{\label{fig:showo-rank}\includegraphics[width=0.24\linewidth]{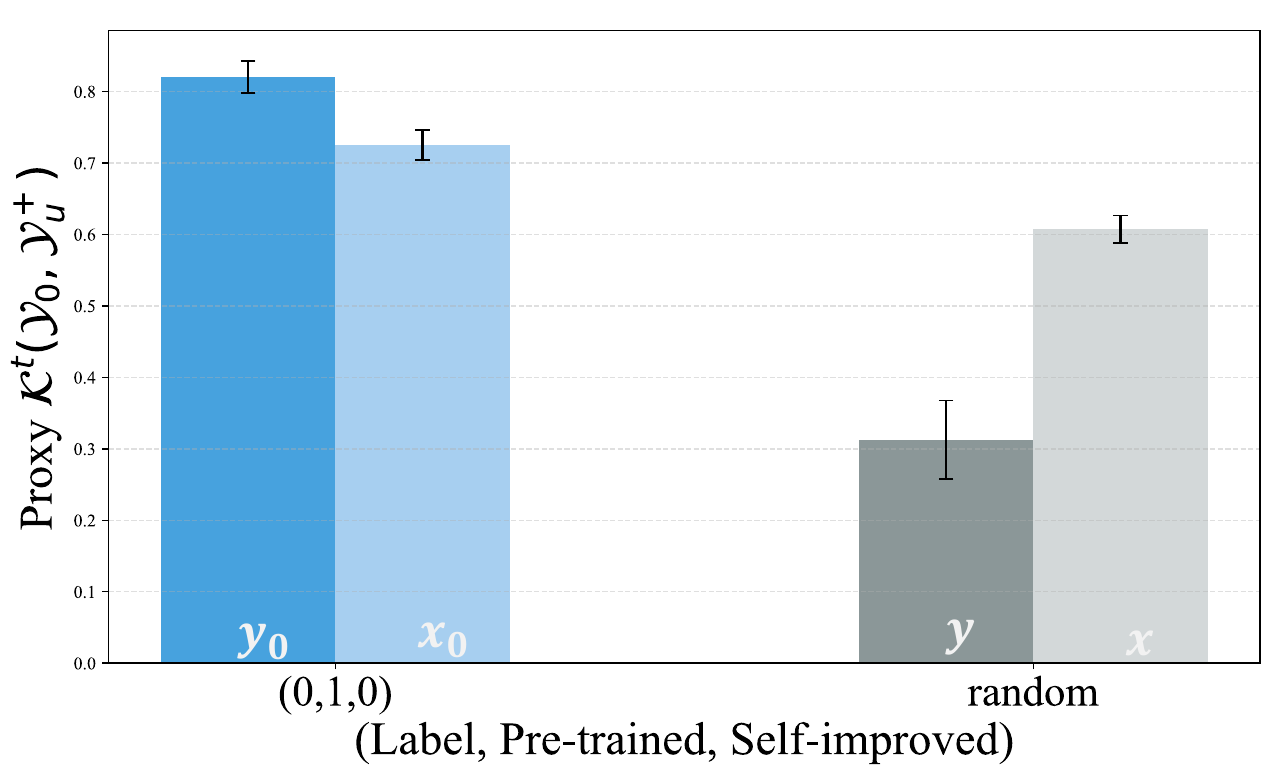}}
    \hfill
    \subfigure[\scriptsize Proxy $\|\textcolor{lightblue}{\mathcal{K}^t(\mathcal{Y}_0,\mathcal{Y}^+_u)}\|_F > \|\textcolor{lightred}{\mathcal{K}^t(\mathcal{Y}_i,\mathcal{Y}^+_u)}\|_F$]{\label{fig:showo-chosen--}\includegraphics[width=0.245\linewidth]{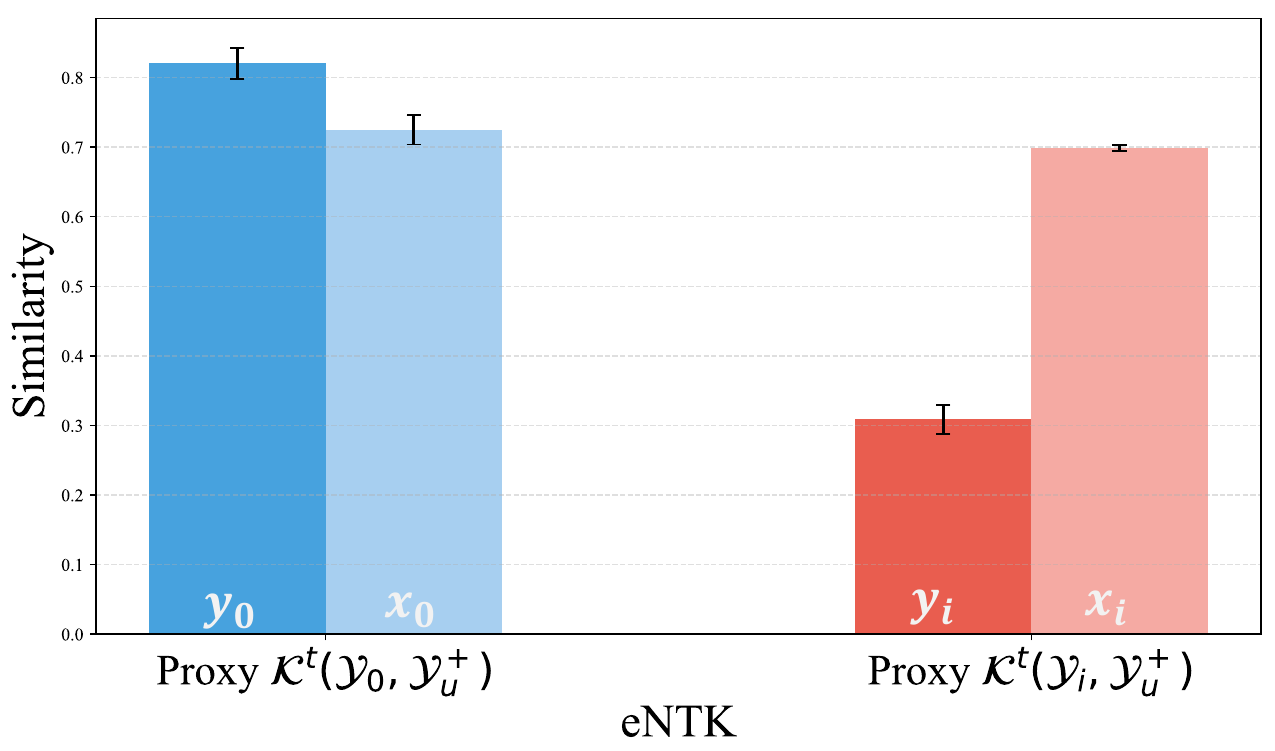}}
    \hfill
    \subfigure[\scriptsize Similar $\mathcal{Y}_0$ and $\mathcal{Y}^-_u$]{\label{fig:showo-rank--}\includegraphics[width=0.24\linewidth]{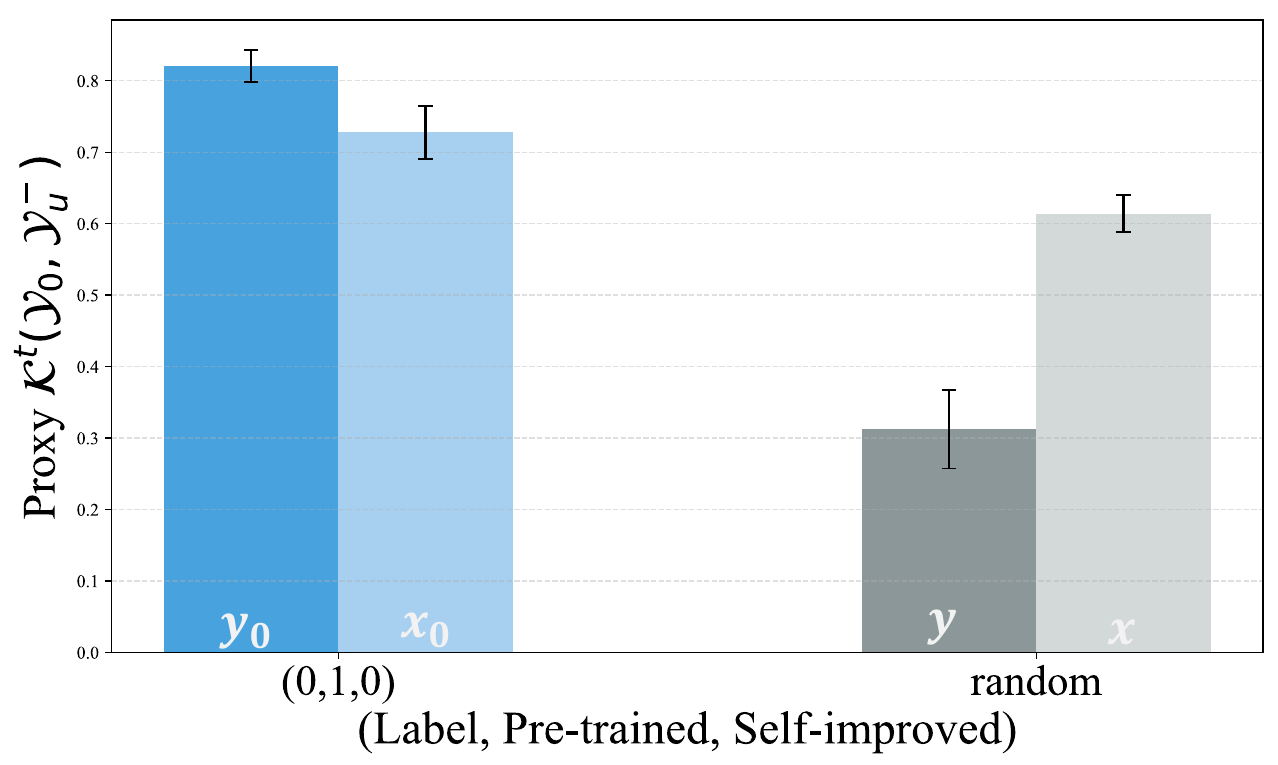}}
    \hfill
    \subfigure[\scriptsize Proxy $\|\textcolor{lightblue}{\mathcal{K}^t(\mathcal{Y}_0,\mathcal{Y}^-_u)}\|_F > \|\textcolor{lightred}{\mathcal{K}^t(\mathcal{Y}_i,\mathcal{Y}^-_u)}\|_F$]{\label{fig:showo-chosen}\includegraphics[width=0.245\linewidth]{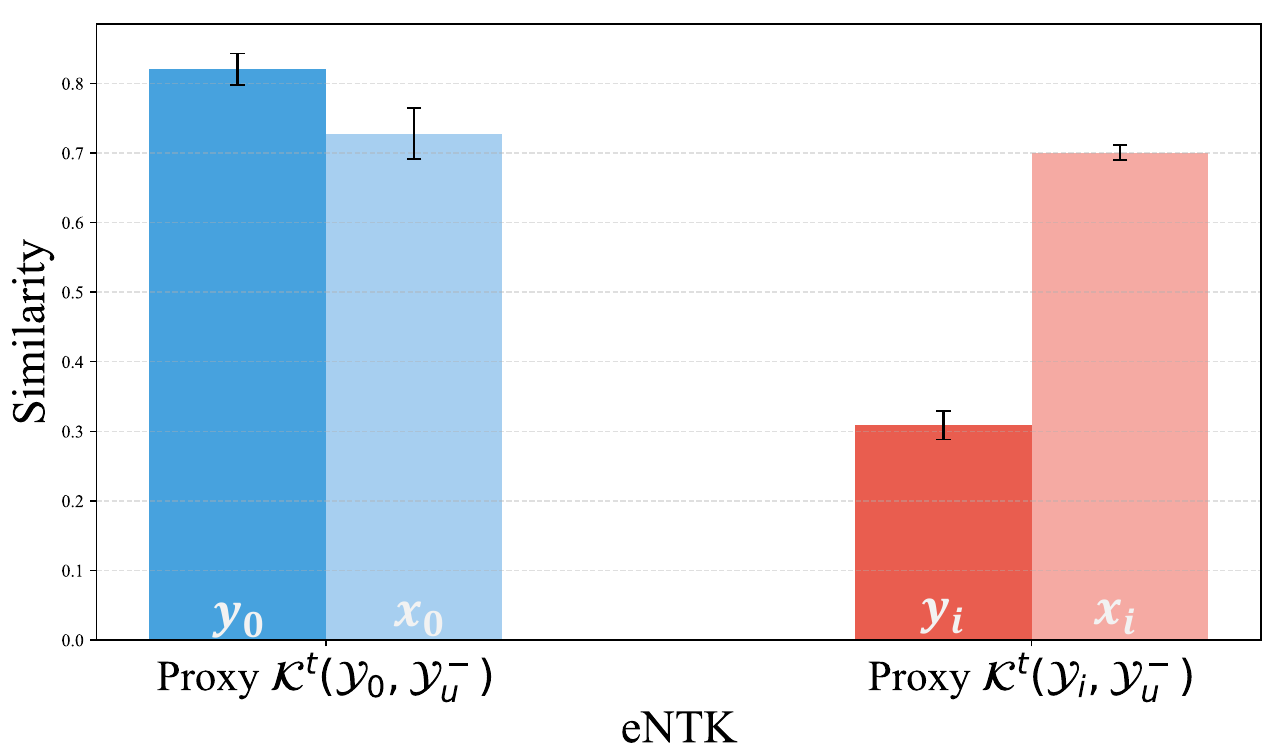}}
    \hfill
\vspace{-0.1in}
\caption{Empirical Evidence from Self‑Improvement with Show-o and DPO. (a)(c) Compared to random samples, those in the false positive correction group are more likely to be matched with highly similar post-training pairs $(\mathbf{y}_u,\mathbf{x}_u)$ (average cosine similarity 0.8). (b)(d) Such high similarity makes $\textcolor{lightblue}{\text{Term I}}$ be the dominant term in \Cref{eq:und-main-dpo}, thereby promoting aligned learning dynamics between understanding in \Cref{eq:und-main-dpo} and generation in \Cref{eq:gen-main-dpo}.}
\vspace{-0.15in}
\label{fig:dpo-evi-3}
\end{figure}

For self-improved Janus-Pro with DPO, by \Cref{theorem:interpaly-dpo}, co-improvement can arise when the post-training data include pairs $(\mathbf{y}_u,\mathbf{x}_u)$ whose prompt $\mathbf{y}_u$ is more similar to $\mathbf{y}_0$ than any other prompt $\mathbf{y}_i$ (empirical evidence in \Cref{fig:dpo-evi-2}(a)(c) and \Cref{fig:dpo-evi-3}(a)(c)). In this case, the understanding update $\Delta U_t$ in \Cref{eq:und-main-dpo} is dominated by \textcolor{lightblue}{Term I} rather than \textcolor{lightred}{Term II} (empirical evidence in \Cref{fig:dpo-evi-2}(b)(d) and \Cref{fig:dpo-evi-3}(b)(d)). Note that, because $\mathcal{K}^{\,t}_{k,r}(\mathcal{Y}_0,\mathcal{Y}^+_u)$ and $\mathcal{K}^{\,t}_{k,r}(\mathcal{Y}_0,\mathcal{Y}^-_u)$ share the same prompt $\mathbf{y}_u$, their Frobenius norms are both large, reflecting the high similarity between $\mathbf{y}_0$ and $\mathbf{y}_u$ (empirical evidence in \Cref{fig:dpo-evi-2}(a)(c) and \Cref{fig:dpo-evi-3}(a)(c)). By contrast, $\mathcal{K}^{\,t}_{k,r}(\mathcal{Y}_i,\mathcal{Y}^+_u)$ and $\mathcal{K}^{\,t}_{k,r}(\mathcal{Y}_i,\mathcal{Y}^-_u)$ are significantly smaller due to the lower similarity between $\mathbf{y}_i$ and $\mathbf{y}_u$ (also in \Cref{fig:dpo-evi-2}(b)(d) and \Cref{fig:dpo-evi-3}(b)(d)). The same \textcolor{lightblue}{Term I} therefore aligns the learning dynamics of generation (\Cref{eq:gen-main-dpo}) and understanding (\Cref{eq:und-main-dpo}), yielding consistent update signs $\Delta G_t$ and $\Delta U_t$.

Moreover, such similar post-training pairs $(\mathbf{y}_u,\mathbf{x}_u)$ improve generation by lowering the probability of misaligned outputs, $\pi_{\theta}(\mathbf{x}_0 \mid \mathbf{y}_0)$, leading to $\Delta G_t < 0$ (empirical evidence in \Cref{fig:dpo-evi-1}(c)(d)). Due to the aligned dynamics, $\Delta U_t < 0$ as well, meaning the probability of misjudging, $\pi_{\theta}(\mathbf{y}_0 \mid \mathbf{x}_0)$, is reduced. Consequently, false positive correction emerges, manifesting as co-improvement.

\section{Derivations and Proof Details}
\label{app:proof details}
\paragraph{Preliminaries.}We define the unified vocabulary $\mathcal{V}$ of discrete text and image tokens, with size $V=|\mathcal{V}|$.
Since fine-tuning only updates the LLM part $\pi_{\theta}$ of the MLLM, we work directly in the LLM input space. Let $d$ denote the input embedding dimension. 

We consider the setting where both image generation and image understanding \textit{share the same tokenizer} as the default Show-o and EMU3. This contrasts with decoupled designs such as Janus-Pro, where generation and understanding use separate tokenizers. Nevertheless, our analysis shows that results derived under the shared-tokenizer assumption continue to hold for decoupled architectures like Janus-Pro. Specifically, at inference time, for each sequence of image token IDs $\mathbf{x}_{0}=(x_{0,1},\ldots,x_{0,M})$ and text token IDs $\mathbf{y}_{0}=(y_{0,1},\ldots,y_{0,L})$, we encode them as sequences of embeddings as the inputs of LLM. The image sequence is represented by embeddings
\[
\mathbf{U}_0 \;=\; [\,\mathbf{u}_{0,1}\ \cdots\ \mathbf{u}_{0,M}\,]\in\mathbb{R}^{d\times M},
\]
and the evaluation prompt is represented by
\[
\mathbf{V}_0 \;=\; [\,\mathbf{v}_{0,1}\ \cdots\ \mathbf{v}_{0,L}\,]\in\mathbb{R}^{d\times L}.
\]
where usually $|\mathcal{V}| \gg \max(L,M)$. Similarly, the fine-tuning data pair $(\mathbf{u}_u,\mathbf{v}_u)$ yields
$\mathbf{U}_u\in\mathbb{R}^{d\times M}$ and $\mathbf{V}_u\in\mathbb{R}^{d\times L}$\footnote{Across datapoints, the image length $M$ is fixed while the text length $L$ may vary; we use a common symbol $L$ for simplicity.}.

We consider the typical causal-masking mechanism applied in MLLMs \citep{wu2024janusdecouplingvisualencoding,wang2024emu3,wu2025harmonizing}.  Under this mechanism, $\pi_{\theta}$ takes the full concatenation of image and text embeddings as input and predicts the next token(s) \citep{ren2025learningdynamicsllmfinetuning}. We denote the concatenated inputs by
\[
\mathcal{X}_0=[\,\mathbf{U}_0\mid\mathbf{V}_0\,]\in\mathbb{R}^{d\times(M+L)}\quad\text{(Understanding)},
\]
\[
\mathcal{Y}_0=[\,\mathbf{V}_0\mid\mathbf{U}_0\,]\in\mathbb{R}^{d\times(L+M)}\quad\text{(Generation)}.
\]
where we omit potential special tokens (e.g., \texttt{[SOI]}) for simplicity.

Let $h_{\theta}$ denote the logits network with causal mask implemented. For understanding,
\[
\mathbf{z}^0_{\mathrm{und}}\coloneqq h_{\theta}(\mathcal{X}_0)_{[:,\,M+1:M+L]}\in\mathbb{R}^{V\times L},\qquad
{\Pi}_{\mathrm{und}}\coloneqq\mathrm{softmax}_{\text{col}}(\mathbf{z}^0_{\mathrm{und}})\in\mathbb{R}^{V\times L},
\]
and for generation\footnote{Typically, generation branch includes a projector as generation head. For example, Janus-Pro uses a 2-layer MLP to map LLM outputs to generation tokenizer’s codebook. In our setting, the generation head is frozen.},
\[
\mathbf{z}^0_{\mathrm{gen}}\coloneqq h_{\theta}(\mathcal{Y}_0)_{[:,\,L+1:L+M]}\in\mathbb{R}^{V\times M},\qquad
{\Pi}_{\mathrm{gen}}\coloneqq\mathrm{softmax}_{\text{col}}(\mathbf{z}^0_{\mathrm{gen}})\in\mathbb{R}^{V\times M}.
\]
Let $y_{0,l}\in\mathcal{V}$ and $x_{0,k}\in\mathcal{V}$ denote the scalar ground-truth token ids at text position $l$ and image position $k$, respectively.
Then the modeling of understanding and generation can be factorized as
\begin{align*}
&\log \pi_{\theta}(\mathbf{y}_0\mid \mathcal{X}_0)
=\sum_{l} \log \pi_\theta(y_{0,l}\mid \mathbf{x}_0,\mathbf{y}_{0,<l})=\sum_{l=1}^{L} \log \big[{\Pi}_{\mathrm{und}}\big]_{y_{0,l},l},\\
&\log \pi_{\theta}(\mathbf{x}_0\mid \mathcal{Y}_0)
=\sum_{k} \log \pi_\theta(x_{0,k}\mid \mathbf{y}_0,\mathbf{x}_{0,<k})= \sum_{k=1}^{M} \log \big[{\Pi}_{\mathrm{gen}}\big]_{x_{0,k},k}.
\end{align*}
At epoch $t$, we define the \textit{one-step learning dynamics} of evaluation data pair $(\mathbf{x}_0,\mathbf{y}_0)$ after training one-step on fine-tuning data $(\mathbf{x}_u,\mathbf{y}_u)$ as
\begin{align}
\Delta G_t\!\paren*{\mathbf{x}_0\mid\mathcal{Y}_0}
&\coloneqq
\log \pi_{\theta_{t+1}}\!\paren*{\mathbf{x}_0\mid\mathcal{Y}_0}
-
\log \pi_{\theta_{t}}\!\paren*{\mathbf{x}_0\mid\mathcal{Y}_0}
\tag{Generation}
\\
\Delta U_t\!\paren*{\mathbf{y}_0\mid\mathcal{X}_0}
&\coloneqq
\log \pi_{\theta_{t+1}}\!\paren*{\mathbf{y}_0\mid\mathcal{X}_0}
-
\log \pi_{\theta_{t}}\!\paren*{\mathbf{y}_0\mid\mathcal{X}_0}
\label{eq:ld-und-2}\tag{Understanding}
\end{align}

It is worth noting that \Cref{sec:Results} evaluates understanding improvement in terms of binary classification $f_\theta (\cdot)$ 
whereas the theory focuses on log-likelihood
$\log \pi_{\theta}(\mathbf{y}_0 \mid \mathcal{X}_0)$. 
We introduce a decision rule to bridge the continuous log-likelihood 
with the discrete binary score:
\[
f_\theta(\mathbf{y}_0\mid \mathcal{X}_0) = \mathbf{1}\{\pi_{\theta}(\mathbf{y}_0 \mid \mathcal{X}_0) > \tau\},
\]
where $\tau$ is a threshold. When $\Delta U_t(\mathbf{y}_0 \mid \mathcal{X}_0)$ increases, the understanding branch is encouraged to raise the log-likelihood, making it more likely to yield a score of~1 under the decision rule.

We first show the connection between the learning dynamics of generation and understanding. First, we obtain
\begin{align*}
&\pi_{\theta}(\mathbf{x}_0\mid \mathcal{Y}_0)
= \prod_{k=1}^{M} \pi_{\theta}\!\big(x_{0,k}\mid \mathbf{y}_0,\mathbf{x}_{0,<k}\big)
= \pi_{\theta}(\mathbf{x}_0\mid \mathbf{y}_0),\\
&\pi_{\theta}(\mathbf{y}_0\mid \mathcal{X}_0)
= \prod_{l=1}^{L} \pi_{\theta}\!\big(y_{0,l}\mid \mathbf{x}_0,\mathbf{y}_{0,<l}\big)
= \pi_{\theta}(\mathbf{y}_0\mid \mathbf{x}_0).
\end{align*}
Given the prompts follows a Uniform distribution, Bayes’ rule yields
\begin{align*}
\log \pi_{\theta}(\mathbf{y}_0\mid \mathbf{x}_0)
= \log \pi_{\theta}(\mathbf{x}_0\mid \mathbf{y}_0)
- \log \pi_{\theta}(\mathbf{x}_0) + C.
\end{align*}
where $C \coloneqq \log P(\mathbf{y}_0)$ is a constant under the uniform prompt prior.
Therefore,
\begin{align}
\label{eq:bayes-1}
    \Delta U_t\!\paren*{\mathbf{y}_0\mid\mathcal{X}_0} &= \log \pi_{\theta_{t+1}}\!\paren*{\mathbf{y}_0\mid\mathcal{X}_0}-\log \pi_{\theta_{t}}\!\paren*{\mathbf{y}_0\mid\mathcal{X}_0} \notag\\
    &=(\log \pi_{\theta_{t+1}}(\mathbf{x}_0\mid \mathbf{y}_0)-\log \pi_{\theta_{t}}(\mathbf{x}_0\mid \mathbf{y}_0)) - (\log \pi_{\theta_{t+1}}(\mathbf{x}_0) - \log \pi_{\theta_{t}}(\mathbf{x}_0))\notag\\
    &=\Delta G_t\!\paren*{\mathbf{x}_0\mid\mathcal{Y}_0} - \Delta \log \pi_t(\mathbf{x}_0).
\end{align}

\Cref{eq:bayes-1} implies that the learning dynamics of understanding 
$\Delta U_t(\mathbf{y}_0 \mid \mathcal{X}_0)$ and those of generation 
$\Delta G_t(\mathbf{x}_0 \mid \mathcal{Y}_0)$ differ only in the change of the 
marginal distribution $\pi_t(\mathbf{x}_0)$ between consecutive steps. We next consider the training dynamics of the generation and understanding branches under different post-training strategies, SFT and DPO.

\subsection{Learning Dynamics under SFT}
Following \eqref{eq:bayes-1}, we first discuss the training dynamics of the generation branch, i.e.,$\Delta G_t\!\paren*{\mathbf{x}_0\mid\mathcal{Y}_0}$, and then provide an indirect estimation for the understanding branch $\Delta U_t\!\paren*{\mathbf{y}_0\mid\mathcal{X}_0}$.
\begin{lemma}
[Learning Dynamics of Generation under SFT]
\label{lemma:gen-ld-sft}
Consider self-improvement proposed in \Cref{sec:Mitigating Non-Unification: A Self-Improvement Framework} with SFT. At epoch $t$,
    the one-step learning dynamics of \textbf{generation} is
\begin{align}
  \label{eq:app-gen-main}
  \Delta G_t(\mathbf{x}_0\mid\mathcal{Y}_0) =
  -\eta\sum_{k=1}^{M}\sum_{r=1}^{M}
  (\mathbf{e}_{x_{0,k}}-{\pi}^{0}_{k})^{\top}
  {{\mathcal{K}^t_{k,r}(\mathcal{Y}_0,\mathcal{Y}_u)}}
  ({\pi}^{u}_{r}-\mathbf{e}_{x_{u,r}})
  + \mathcal{O}(\eta^2),
\end{align}
   where ${\pi}^{u}_{r}=\mathrm{softmax}(\mathbf{z}^{u}_{r})$ and $\mathbf{z}^{u}_{r}=[h_{\theta}(\mathcal{Y}_u)]_r$ are the logits at image position $r$ obtained by running $h_{\theta}$ on $\mathcal{Y}_u$ and $\mathcal{K}^t_{k,r}(\mathcal{Y}_0,\mathcal{Y}_u)\coloneqq(\nabla_{\theta_t}\mathbf{z}^{0}_{k})(\nabla_{\theta_t}\mathbf{z}^{u}_{r})^{\!\top} \in \mathbb{R}^{V \times V}$ is empirical neural tangent kernel (eNTK). 
\end{lemma}
\begin{proof}
    We first show the learning dynamic of generation, i.e., $\Delta G_t\!\paren*{\mathbf{x}_0\mid\mathcal{Y}_0}$ under the SFT setting. Consider the $k$-th image token
\begin{align}
\label{eq:taylor-1}
\big(\Delta G_t(\mathbf{x}_0\mid\mathcal{Y}_0)\big)_k
&\coloneqq
\big[\log \pi_{\theta_{t+1}}(\mathbf{x}_0\mid\mathcal{Y}_0)\big]_k
-\big[\log \pi_{\theta_t}(\mathbf{x}_0\mid\mathcal{Y}_0)\big]_k \notag\\
&= \nabla_{\theta}\big[\log \pi_{\theta_t}(\mathbf{x}_0\mid\mathcal{Y}_0)\big]_k^{\!\top}
   (\theta_{t+1}-\theta_t)
   +\mathcal{O}\!\big(\|\theta_{t+1}-\theta_t\|^2\big).
\end{align}
where $\big[\log \pi_{\theta}(\mathbf{x}_0\mid \mathcal{Y}_0)\big]_k
\coloneqq \log \pi_{\theta}\!\big(x_{0,k}\mid \mathbf{y}_0,\mathbf{x}_{0,<k}\big).$

Given the post-training data $(\mathbf{x}_u,\mathbf{y}_u)$, for generation, the negative log-likelihood loss of SFT is 
    \begin{align*}
        \mathcal{L}_{\text{SFT}}(\mathcal{Y}_u)=-\sum_{r=1}^{M}\log \pi_{\theta}(x_r=x_{u,r} \mid \mathcal{Y}_u)= -\sum_{r=1}^{M}\log \big[{\pi}^{u}_{r}\big]_{x_{u,r}}
    \end{align*}
where ${\pi}^{u}_{r}=\mathrm{softmax}(\mathbf{z}^{u}_{r})$ and $\mathbf{z}^{u}_{r}=[h_{\theta}(\mathcal{Y}_u)]_r$ are the logits at image position $r$ obtained by running $h_{\theta}$ on $\mathcal{Y}_u=[\mathbf{V}_u\mid\mathbf{U}_u]$. One-step SGD yields
\begin{align*}
    \theta_{t+1}-\theta_t
= - \eta \nabla_{\theta}\mathcal{L}_{\mathrm{SFT}}(\mathcal{Y}_u)
= -\eta\sum_{r=1}^{M}(\nabla_{\theta}\mathbf{z}^{u}_{r})^{\!\top}\mathcal{G}_{r},
\end{align*}
where $\mathcal{G}_{r}\coloneqq {\pi}^{u}_{r}-\mathbf{e}_{x_{u,r}}\in\mathbb{R}^{V}$.

Then, we obtain
\begin{align*}
    \nabla_{\theta}\big[\log \pi_{\theta_t}(\mathbf{x}_0\mid\mathcal{Y}_0)\big]_k = (\nabla_{\theta_t}\mathbf{z}^{0}_{k})^{\!\top}(\mathbf{e}_{x_{0,k}}-{\pi}^{0}_{k}).
\end{align*}

Therefore, \Cref{eq:taylor-1} can be rewritten as
\begin{align*}
    \big(\Delta G_t(\mathbf{x}_0\mid\mathcal{Y}_0)\big)_k &= -\eta\sum_{r=1}^{M}(\mathbf{e}_{x_{0,k}}-{\pi}^{0}_{k})^{\top}(\nabla_{\theta}\mathbf{z}^{0}_{k})(\nabla_{\theta}\mathbf{z}^{u}_{r})^{\!\top}\mathcal{G}_{r} + \mathcal{O}(\eta^2)\\
    &= -\eta\sum_{r=1}^{M} (\mathbf{e}_{x_{0,k}}-{\pi}^{0}_{k})^{\top} \mathcal{K}^t_{k,r}(\mathcal{Y}_0,\mathcal{Y}_u)({\pi}^{u}_{r}-\mathbf{e}_{x_{u,r}}) + \mathcal{O}(\eta^2)
\end{align*}
where $\mathcal{K}^t_{k,r}(\mathcal{Y}_0,\mathcal{Y}_u)\coloneqq(\nabla_{\theta_t}\mathbf{z}^{0}_{k})(\nabla_{\theta_t}\mathbf{z}^{u}_{r})^{\!\top} \in \mathbb{R}^{V \times V}$.

Finally, we have the sequence-level one-step change as:
\begin{align*}
  \Delta G_t(\mathbf{x}_0\mid\mathcal{Y}_0) 
&= \sum_k \big[\log \pi_{\theta_{t+1}}(\mathbf{x}_0\mid \mathcal{Y}_0)\big]_k-\sum_k \big[\log \pi_{\theta_{t}}(\mathbf{x}_0\mid \mathcal{Y}_0)\big]_k\\
  &=\sum_{k=1}^{M}\big(\Delta G_t(\mathbf{x}_0\mid\mathcal{Y}_0)\big)_k \\
&=  -\eta\sum_{k=1}^{M}\sum_{r=1}^{M} (\mathbf{e}_{x_{0,k}}-{\pi}^{0}_{k})^{\top} \mathcal{K}^t_{k,r}(\mathcal{Y}_0,\mathcal{Y}_u)({\pi}^{u}_{r}-\mathbf{e}_{x_{u,r}}) + \mathcal{O}(\eta^2).
\end{align*}

The proof is complete.
\end{proof}

\begin{lemma}
[Learning Dynamics of Understanding under SFT]
\label{lemma:und-ld-sft}
Consider self-improvement proposed in \Cref{sec:Mitigating Non-Unification: A Self-Improvement Framework} with SFT. At epoch $t$, the one-step learning dynamics of \textbf{understanding} is
\begin{equation}
\label{eq:app-und-main}
\resizebox{\linewidth}{!}{$
\begin{aligned}
\Delta U_t(\mathbf{y}_0\mid\mathcal{X}_0)&= -\eta \sum^M_{k=1}\sum_{r=1}^{M}\sum_{\mathbf y_i\neq \mathbf y_0} w_{\theta_t}(\mathbf y_i\mid \mathbf x_0)\,
\Big(
(\mathbf{e}_{x_{0,k}}-{\pi}^{0}_{k})^{\!\top}
{{\mathcal{K}^{\,t}_{k,r}(\mathcal{Y}_0,\mathcal{Y}_u)}}
-
(\mathbf{e}_{x_{0,k}}-{\pi}^{i}_{k})^{\!\top}
{{\mathcal{K}^{\,t}_{k,r}(\mathcal{Y}_i,\mathcal{Y}_u)}}
\Big)
({\pi}^{u}_{r}-\mathbf{e}_{x_{u,r}})\\
&\quad + \mathcal{O}(\eta^2)
\end{aligned}
$}
\end{equation}
   where   $w_{\theta_t}(\mathbf y \mid \mathbf x_0)\;\coloneqq\;
\frac{\pi_{\theta_t}(\mathbf x_0\mid \mathbf y)}{\sum_{\mathbf y'} \pi_{\theta_t}(\mathbf x_0\mid \mathbf y')}$ and $\mathcal{Y}_i$ denotes the concatenation of prompt $\mathbf y_i \neq \mathbf y_0$ and $ \mathbf{x}_0$.
\end{lemma}
\begin{proof}
We then analyze the learning dynamics of the understanding branch. 
By \Cref{eq:bayes-1} and a first–order log-sum-exp expansion, we obtain
\begin{align*}
\Delta \log \pi_t(\mathbf{x}_0) 
&\coloneqq \log \pi_{\theta_{t+1}}(\mathbf{x}_0) - \log \pi_{\theta_t}(\mathbf{x}_0)\\
&= \log \sum_{\mathbf y}\pi_{\theta_{t+1}}(\mathbf{x}_0 \mid \mathbf y) - \log \sum_{\mathbf y} \pi_{\theta_t}(\mathbf{x}_0 \mid \mathbf y)\\
&= \Big\langle \sum_{\mathbf y} w_{\theta_t}(\mathbf y \mid \mathbf x_0)\,
\nabla_\theta \log \pi_{\theta_t}(\mathbf{x}_0\mid \mathbf y)\,,\ \theta_{t+1}-\theta_t \Big\rangle
+ \mathcal{O}\!\big(\|\theta_{t+1}-\theta_t\|^2\big)
\end{align*}
where the posterior weight is
\[
w_{\theta_t}(\mathbf y \mid \mathbf x_0)\;\coloneqq\;
\frac{\pi_{\theta_t}(\mathbf x_0\mid \mathbf y)}{\sum_{\mathbf y'} \pi_{\theta_t}(\mathbf x_0\mid \mathbf y')}.
\]

Following \Cref{lemma:gen-ld-sft} and \Cref{eq:bayes-1}, we obtain
\begin{equation}
\label{eq:taylor-2}
\resizebox{\linewidth}{!}{$
\begin{aligned}
&\Delta U_t(\mathbf{y}_0\mid\mathcal{X}_0)\\
&=\Delta G_t\!\paren*{\mathbf{x}_0\mid\mathcal{Y}_0} - \Delta \log \pi_t(\mathbf{x}_0) \\
&= \nabla_{\theta}\log \pi_{\theta_t}(\mathbf{x}_0\mid\mathcal{Y}_0)^{\!\top}(\theta_{t+1}-\theta_t)
- \sum_{\mathbf y_i} w_{\theta_t}(\mathbf y_i \mid \mathbf x_0)\,
   \nabla_{\theta}\log \pi_{\theta_t}(\mathbf{x}_0\mid\mathcal{Y}_i)^{\!\top}(\theta_{t+1}-\theta_t)
+ \mathcal{O}\!\big(\|\theta_{t+1}-\theta_t\|^2\big)
\\
&= -\eta \sum^M_{k=1}\sum_{r=1}^{M}\sum_{\mathbf y_i\neq \mathbf y_0} w_{\theta_t}(\mathbf y_i\mid \mathbf x_0)\,
\Big(
(\mathbf{e}_{x_{0,k}}-{\pi}^{0}_{k})^{\!\top}
\mathcal{K}^{\,t}_{k,r}(\mathcal{Y}_0,\mathcal{Y}_u)
-
(\mathbf{e}_{x_{0,k}}-{\pi}^{i}_{k})^{\!\top}
\mathcal{K}^{\,t}_{k,r}(\mathcal{Y}_i,\mathcal{Y}_u)
\Big)
({\pi}^{u}_{r}-\mathbf{e}_{x_{u,r}})
\\
&\quad + \mathcal{O}(\eta^2)
\end{aligned}
$}
\end{equation}
where $\mathcal{Y}_i$ denote the concatenation obtained by appending the embedding of $\mathbf y_i$ to $\mathbf U_0$.
The proof is complete.
\end{proof}

\subsection{Learning Dynamics under DPO}

\begin{lemma}
[Learning Dynamics of Generation under DPO]
\label{lemma:gen-ld-dpo}
Consider self-improvement proposed in \Cref{sec:Mitigating Non-Unification: A Self-Improvement Framework} with DPO. At epoch $t$, the one-step learning dynamics of \textbf{generation} is
\begin{align}
\label{eq:app-gen-dpo}
    &\Delta G_t(\mathbf{x}_0\mid\mathcal{Y}_0) \notag\\
    &= -\eta \beta \sigma(-\alpha) \sum_{k=1}^{M}\sum_{r=1}^{M} (\mathbf{e}_{x_{0,k}}-{\pi}^{0}_{k})^{\top} \Big[ \mathcal{K}^t_{k,r}(\mathcal{Y}_0,\mathcal{Y}^+_u)({\pi}^{u,+}_{r}-\mathbf{e}_{x^+_{u,r}})-\mathcal{K}^t_{k,r}(\mathcal{Y}_0,\mathcal{Y}^-_u)({\pi}^{u,-}_{r}-\mathbf{e}_{x^-_{u,r}})\Big]\notag\\
    &\quad + \mathcal{O}(\eta^2)
\end{align}
where the margin $\alpha \coloneqq  \beta \log \frac{\pi_\theta(\mathbf{x}_u^+ \mid \mathcal{Y}_u^+)}{\pi_{\mathrm{ref}}(\mathbf{x}_u^+ \mid \mathcal{Y}_u^+)}
        - \beta \log \frac{\pi_\theta(\mathbf{x}_u^- \mid \mathcal{Y}_u^-)}{\pi_{\mathrm{ref}}(\mathbf{x}_u^- \mid \mathcal{Y}_u^-)}$ and ${\pi}^{u,+}_{r}=\mathrm{softmax}(\mathbf{z}^{u,+}_{r})$ and $\mathbf{z}^{u,+}_{r}=[h_{\theta}(\mathcal{Y}^+_u)]_r$ are the logits at image position $r$ obtained by running $h_{\theta}$ on $\mathcal{Y}^+_u$. The neural tangent kernel $\mathcal{K}^t_{k,r}(\mathcal{Y}_0,\mathcal{Y}^+_u) \coloneqq \nabla_{\theta}\mathbf{z}^{0}_{k}(\nabla_{\theta}\mathbf{z}^{u,+}_{r})^{\!\top}$
and $\mathcal{K}^t_{k,r}(\mathcal{Y}_0,\mathcal{Y}^-_u) \coloneqq \nabla_{\theta}\mathbf{z}^{0}_{k}(\nabla_{\theta}\mathbf{z}^{u,-}_{r})^{\!\top}$.
\end{lemma}
\begin{proof}
    Following \eqref{eq:taylor-1}, one-step SGD yields
\begin{align*}
    \theta_{t+1}-\theta_t
&= - \eta \nabla_{\theta}\mathcal{L}_{\mathrm{DPO}}(\mathcal{Y}_u)\\
&= -\eta \sum^M_{r=1}  \Big[(\nabla_{\theta}\mathbf{z}^{u,+}_{r})^{\!\top}\nabla_{\mathbf{z}^{u,+}_{r}}\mathcal{L}_{\text{DPO}}+(\nabla_{\theta}\mathbf{z}^{u,-}_{r})^{\!\top}\nabla_{\mathbf{z}^{u,-}_{r}}\mathcal{L}_{\text{DPO}}\Big]\\
&= -\eta \beta \sigma(-\alpha) \sum^M_{r=1}  \Big[(\nabla_{\theta}\mathbf{z}^{u,+}_{r})^{\!\top}(\pi^{u,+}_r-\mathbf{e}_{x^+_{u,r}})-(\nabla_{\theta}\mathbf{z}^{u,-}_{r})^{\!\top}(\pi^{u,-}_r-\mathbf{e}_{x^-_{u,r}})\Big],
\end{align*}
where the margin $\alpha \coloneqq  \beta \log \frac{\pi_\theta(\mathbf{x}_u^+ \mid \mathcal{Y}_u^+)}{\pi_{\mathrm{ref}}(\mathbf{x}_u^+ \mid \mathcal{Y}_u^+)}
        - \beta \log \frac{\pi_\theta(\mathbf{x}_u^- \mid \mathcal{Y}_u^-)}{\pi_{\mathrm{ref}}(\mathbf{x}_u^- \mid \mathcal{Y}_u^-)}$. And ${\pi}^{u,+}_{r}=\mathrm{softmax}(\mathbf{z}^{u,+}_{r})$ and $\mathbf{z}^{u,+}_{r}=[h_{\theta}(\mathcal{Y}^+_u)]_r$ are the logits at image position $r$ obtained by running $h_{\theta}$ on $\mathcal{Y}^+_u=[\mathbf{V}_u\mid\mathbf{U}^+_u]$.

Then, we have
\begin{align*}
    &\big(\Delta G_t(\mathbf{x}_0\mid\mathcal{Y}_0)\big)_k \\
    &= -\eta\beta \sigma(-\alpha) \sum_{r=1}^{M}(\mathbf{e}_{x_{0,k}}-{\pi}^{0}_{k})^{\top}(\nabla_{\theta}\mathbf{z}^{0}_{k})\Big[(\nabla_{\theta}\mathbf{z}^{u,+}_{r})^{\!\top}(\pi^{u,+}_r-\mathbf{e}_{x^+_{u,r}})-(\nabla_{\theta}\mathbf{z}^{u,-}_{r})^{\!\top}(\pi^{u,-}_r-\mathbf{e}_{x^-_{u,r}})\Big] + \mathcal{O}(\eta^2)\\
    &= -\eta \beta \sigma(-\alpha) \sum_{r=1}^{M} (\mathbf{e}_{x_{0,k}}-{\pi}^{0}_{k})^{\top} \Big[ \mathcal{K}^t_{k,r}(\mathcal{Y}_0,\mathcal{Y}^+_u)({\pi}^{u,+}_{r}-\mathbf{e}_{x^+_{u,r}})-\mathcal{K}^t_{k,r}(\mathcal{Y}_0,\mathcal{Y}^-_u)({\pi}^{u,-}_{r}-\mathbf{e}_{x^-_{u,r}})\Big] + \mathcal{O}(\eta^2)
\end{align*}
where the neural tangent kernel $\mathcal{K}^t_{k,r}(\mathcal{Y}_0,\mathcal{Y}^+_u) \coloneqq \nabla_{\theta}\mathbf{z}^{0}_{k}(\nabla_{\theta}\mathbf{z}^{u,+}_{r})^{\!\top}$
and $\mathcal{K}^t_{k,r}(\mathcal{Y}_0,\mathcal{Y}^-_u) \coloneqq \nabla_{\theta}\mathbf{z}^{0}_{k}(\nabla_{\theta}\mathbf{z}^{u,-}_{r})^{\!\top}$.

Finally, we have the sequence-level one-step change as:
\begin{align*}
    &\Delta G_t(\mathbf{x}_0\mid\mathcal{Y}_0)\\
    &= -\eta \beta \sigma(-\alpha) \sum_{k=1}^{M}\sum_{r=1}^{M} (\mathbf{e}_{x_{0,k}}-{\pi}^{0}_{k})^{\top} \Big[ \mathcal{K}^t_{k,r}(\mathcal{Y}_0,\mathcal{Y}^+_u)({\pi}^{u,+}_{r}-\mathbf{e}_{x^+_{u,r}})-\mathcal{K}^t_{k,r}(\mathcal{Y}_0,\mathcal{Y}^-_u)({\pi}^{u,-}_{r}-\mathbf{e}_{x^-_{u,r}})\Big]\\
    &\quad + \mathcal{O}(\eta^2)
\end{align*}
The proof is complete.
\end{proof}

\begin{lemma}
[Learning Dynamics of Understanding under DPO]
\label{lemma:und-ld-dpo}
Consider self-improvement proposed in \Cref{sec:Mitigating Non-Unification: A Self-Improvement Framework} with DPO. At epoch $t$, the one-step learning dynamics of \textbf{understanding} is
\begin{equation}
\label{eq:app-und-dpo}
\resizebox{\linewidth}{!}{$
\begin{aligned}
&\Delta U_t(\mathbf{y}_0\mid\mathcal{X}_0)\\
&= -\eta \beta \sigma(-\alpha) \sum^M_{k=1}\sum_{r=1}^{M}\sum_{\mathbf y_i\neq \mathbf y_0} w_{\theta_t}(\mathbf y_i\mid \mathbf x_0)\,
\Bigg((\mathbf{e}_{x_{0,k}}-{\pi}^{0}_{k})^{\!\top} \Big(
\mathcal{K}^{\,t}_{k,r}(\mathcal{Y}_0,\mathcal{Y}^+_u)({\pi}^{u,+}_{r}-\mathbf{e}_{x^+_{u,r}})
-
\mathcal{K}^{\,t}_{k,r}(\mathcal{Y}_0,\mathcal{Y}^-_u)({\pi}^{u,-}_{r}-\mathbf{e}_{x^-_{u,r}})
\Big)
\\
&\quad -
(\mathbf{e}_{x_{0,k}}-{\pi}^{i}_{k})^{\!\top} \Big(
\mathcal{K}^{\,t}_{k,r}(\mathcal{Y}_i,\mathcal{Y}^+_u)({\pi}^{u,+}_{r}-\mathbf{e}_{x^+_{u,r}})
-
\mathcal{K}^{\,t}_{k,r}(\mathcal{Y}_i,\mathcal{Y}^-_u)({\pi}^{u,-}_{r}-\mathbf{e}_{x^-_{u,r}})
\Big)\Bigg) + \mathcal{O}(\eta^2)
\end{aligned}
$}
\end{equation}
where $\mathcal{Y}_i$ denote the concatenation obtained by appending the embedding of $\mathbf y_i$ to $\mathbf U_0$.
\end{lemma}
\begin{proof}
Following \ref{lemma:und-ld-sft}, for the learning dynamics of understanding under DPO, we have   
\begin{equation}
\label{eq:taylor-3}
\resizebox{\linewidth}{!}{$
\begin{aligned}
&\Delta U_t(\mathbf{y}_0\mid\mathcal{X}_0)\\
&=\Delta G_t\!\paren*{\mathbf{x}_0\mid\mathcal{Y}_0} - \Delta \log \pi_t(\mathbf{x}_0) \\
&= \nabla_{\theta}\log \pi_{\theta_t}(\mathbf{x}_0\mid\mathcal{Y}_0)^{\!\top}(\theta_{t+1}-\theta_t)
- \sum_{\mathbf y_i} w_{\theta_t}(\mathbf y_i \mid \mathbf x_0)\,
   \nabla_{\theta}\log \pi_{\theta_t}(\mathbf{x}_0\mid\mathcal{Y}_i))^{\!\top}(\theta_{t+1}-\theta_t)
+ \mathcal{O}\!\big(\|\theta_{t+1}-\theta_t\|^2\big)
\\
&= \sum_{\mathbf y_i\neq \mathbf y_0} w_{\theta_t}(\mathbf y_i\mid \mathbf x_0) \Big( \nabla_{\theta}\log \pi_{\theta_t}(\mathbf{x}_0\mid\mathcal{Y}_0)^{\!\top}- \nabla_{\theta}\log \pi_{\theta_t}(\mathbf{x}_0\mid\mathcal{Y}_i)^{\!\top}\Big)(\theta_{t+1}-\theta_t) + \mathcal{O}\!\big(\|\theta_{t+1}-\theta_t\|^2\big) \\
&= -\eta \beta \sigma(-\alpha) \sum^M_{k=1}\sum_{r=1}^{M}\sum_{\mathbf y_i\neq \mathbf y_0} w_{\theta_t}(\mathbf y_i\mid \mathbf x_0)\,
\Bigg(\Big(
(\mathbf{e}_{x_{0,k}}-{\pi}^{0}_{k})^{\!\top}
\mathcal{K}^{\,t}_{k,r}(\mathcal{Y}_0,\mathcal{Y}^+_u)
-
(\mathbf{e}_{x_{0,k}}-{\pi}^{i}_{k})^{\!\top}
\mathcal{K}^{\,t}_{k,r}(\mathcal{Y}_i,\mathcal{Y}^+_u)
\Big)
({\pi}^{u,+}_{r}-\mathbf{e}_{x^+_{u,r}})
\\
&\quad -
\Big(
(\mathbf{e}_{x_{0,k}}-{\pi}^{0}_{k})^{\!\top}
\mathcal{K}^{\,t}_{k,r}(\mathcal{Y}_0,\mathcal{Y}^-_u)
-
(\mathbf{e}_{x_{0,k}}-{\pi}^{i}_{k})^{\!\top}
\mathcal{K}^{\,t}_{k,r}(\mathcal{Y}_i,\mathcal{Y}^-_u)
\Big)
({\pi}^{u,-}_{r}-\mathbf{e}_{x^-_{u,r}})\Bigg) + \mathcal{O}(\eta^2)\\
&= -\eta \beta \sigma(-\alpha) \sum^M_{k=1}\sum_{r=1}^{M}\sum_{\mathbf y_i\neq \mathbf y_0} w_{\theta_t}(\mathbf y_i\mid \mathbf x_0)\,
\Bigg((\mathbf{e}_{x_{0,k}}-{\pi}^{0}_{k})^{\!\top} \Big(
\mathcal{K}^{\,t}_{k,r}(\mathcal{Y}_0,\mathcal{Y}^+_u)({\pi}^{u,+}_{r}-\mathbf{e}_{x^+_{u,r}})
-
\mathcal{K}^{\,t}_{k,r}(\mathcal{Y}_0,\mathcal{Y}^-_u)({\pi}^{u,-}_{r}-\mathbf{e}_{x^-_{u,r}})
\Big)
\\
&\quad -
(\mathbf{e}_{x_{0,k}}-{\pi}^{i}_{k})^{\!\top} \Big(
\mathcal{K}^{\,t}_{k,r}(\mathcal{Y}_i,\mathcal{Y}^+_u)({\pi}^{u,+}_{r}-\mathbf{e}_{x^+_{u,r}})
-
\mathcal{K}^{\,t}_{k,r}(\mathcal{Y}_i,\mathcal{Y}^-_u)({\pi}^{u,-}_{r}-\mathbf{e}_{x^-_{u,r}})
\Big)\Bigg) + \mathcal{O}(\eta^2)
\end{aligned}
$}
\end{equation}
where $\mathcal{Y}_i$ denote the concatenation obtained by appending the embedding of $\mathbf y_i$ to $\mathbf U_0$.
\end{proof}




\section{Ablation Study}
\label{app:Ablation Study}

\subsection{Ablation on Updated Model Components}
\label{app:Fine-tuned architecture}
In \Cref{sec:Empirical Validation of Self-Improvement on MLLMs}, we update only the parameters of the LLM component during self-improvement, while keeping all other components frozen. This design aligns with prior work on MLLMs (focused solely on image understanding), which suggests that optimizing the LLM alone is sufficient to improve MLLM performance, while updating other components yields limited gains \citep{verma2024crossmodalprojectionmultimodalllms}. \Cref{tab:archi} supports our setting: fine-tuning only the LLM already enables the self-improved Janus-Pro-7B to achieve improvements in generation, understanding and unification. However, expanding the parameter updates to include the image aligner (a two-layer MLP projector that maps image tokens to the LLM input space), the generation head (a two-layer MLP that projects LLM output into tokenizer’s codebook space), or even the vision tower, did not lead to significant performance gains in generation and slight declines were observed in both understanding and unification.

\begin{table*}[h] 
\centering
\setlength{\tabcolsep}{2.8pt}     
\renewcommand{\arraystretch}{1.05}

\begin{adjustbox}{max width=\textwidth, max height=\textheight, keepaspectratio}
  \begin{tabular}{l ccc |ccc |ccc |ccc |ccc |ccc |ccc}
    \toprule
    \multirow{2}{*}{Model} & \multicolumn{3}{c}{Texture} & \multicolumn{3}{c}{Shape} & \multicolumn{3}{c}{Color} & \multicolumn{3}{c}{Spatial} & \multicolumn{3}{c}{Non-Spatial} & \multicolumn{3}{c}{Complex} & \multicolumn{3}{c}{Overall} \\
    \cmidrule(lr){2-4}\cmidrule(lr){5-7}\cmidrule(lr){8-10}\cmidrule(lr){11-13}\cmidrule(lr){14-16}\cmidrule(lr){17-19}\cmidrule(lr){20-22}
    & Gen.$\uparrow$ & Und.$\uparrow$ & Non.$\downarrow$
    & Gen.$\uparrow$ & Und.$\uparrow$ & Non.$\downarrow$
    & Gen.$\uparrow$ & Und.$\uparrow$ & Non.$\downarrow$
    & Gen.$\uparrow$ & Und.$\uparrow$ & Non.$\downarrow$
    & Gen.$\uparrow$ & Und.$\uparrow$ & Non.$\downarrow$
    & Gen.$\uparrow$ & Und.$\uparrow$ & Non.$\downarrow$
    & Gen.$\uparrow$ & Und.$\uparrow$ & Non.$\downarrow$ \\
    \midrule
\multicolumn{10}{l}{\textit{Gen. and Und.}}{\vspace{0.02in}}  \\
\pz\pz Janus-Pro-7B$_{\textit{\scriptsize (Baseline)}}$ & 38.63 &50.00  &43.33 &33.49& 50.00& 43.00&53.22 & 50.00 & 27.33& 16.81 & 50.00&  31.00&  31.40&   50.00 &  2.33& 37.73 &50.00 &10.33&35.21 &50.00 &26.22\\
\pz\pz\pz + \textit{LLM} & 53.93& 65.22 &29.67  &38.63  &53.85  &34.00 &73.41  &54.62 & 10.85 &23.73  &26.67  &22.00&31.45 &75.00 &1.00&38.57 &75.00 &4.33 &43.29 &58.39 &16.98 \\
\pz\pz\pz + \textit{LLM and Projector} &52.98&51.72&31.33
&40.88&56.67&37.67
&73.61&22.73&13.90
&21.04&35.71&23.33
&31.41&66.67&2.00
&38.70&75.00&4.67
&42.10&51.42&18.82\\
\pz\pz\pz + \textit{LLM and Projector and Vision Tower} & 53.62&55.17&28.00&39.39&56.67&36.00&73.56&25.00&10.17&22.45&33.33&21.00&31.41&100.00&0.67&38.64&63.64&6.33&43.18&55.64&17.02 \\
    \bottomrule
  \end{tabular}
\end{adjustbox}
\vspace{-0.1in}
\caption{Based on Janus-Pro-7B, we conducted self-improvement via SFT and observed that only fine-tuning the LLM was sufficient to achieve improvements in both performance and unification. Updating other components, such as the vision tower and projectors, yielded no significant gains.}
\label{tab:archi}
\end{table*}


\subsection{Ablation on Curriculum Learning}
\label{app:Curriculum Learning Parameters}
We introduced curriculum learning at different training epochs (4 and 10). Curriculum replay at both epochs improved self-improvement performance, though performance was better when replay was applied at epoch 10. This is likely because the model’s generative and understanding capabilities had improved by that stage, enabling a more effective use of earlier samples for expanding post-training data. Accordingly, we use epoch 10 for curriculum replay in all experiments.
\begin{table*}[h] 
\centering
\setlength{\tabcolsep}{2.8pt}     
\renewcommand{\arraystretch}{1.05}

\begin{adjustbox}{max width=\textwidth, max height=\textheight, keepaspectratio}
  \begin{tabular}{l ccc |ccc |ccc |ccc |ccc |ccc |ccc}
    \toprule
    \multirow{2}{*}{Model} & \multicolumn{3}{c}{Texture} & \multicolumn{3}{c}{Shape} & \multicolumn{3}{c}{Color} & \multicolumn{3}{c}{Spatial} & \multicolumn{3}{c}{Non-Spatial} & \multicolumn{3}{c}{Complex} & \multicolumn{3}{c}{Overall} \\
    \cmidrule(lr){2-4}\cmidrule(lr){5-7}\cmidrule(lr){8-10}\cmidrule(lr){11-13}\cmidrule(lr){14-16}\cmidrule(lr){17-19}\cmidrule(lr){20-22}
    & Gen.$\uparrow$ & Und.$\uparrow$ & Non.$\downarrow$
    & Gen.$\uparrow$ & Und.$\uparrow$ & Non.$\downarrow$
    & Gen.$\uparrow$ & Und.$\uparrow$ & Non.$\downarrow$
    & Gen.$\uparrow$ & Und.$\uparrow$ & Non.$\downarrow$
    & Gen.$\uparrow$ & Und.$\uparrow$ & Non.$\downarrow$
    & Gen.$\uparrow$ & Und.$\uparrow$ & Non.$\downarrow$
    & Gen.$\uparrow$ & Und.$\uparrow$ & Non.$\downarrow$ \\
    \midrule
\multicolumn{10}{l}{\textit{Gen. and Und.}}{\vspace{0.02in}}  \\
\pz\pz Janus-Pro-7B$_{\textit{\scriptsize (Baseline)}}$ & 38.63 &50.00  &43.33 &33.49& 50.00& 43.00&53.22 & 50.00 & 27.33& 16.81 & 50.00&  31.00&  31.40&   50.00 &  2.33& 37.73 &50.00 &10.33&35.21 &50.00 &26.22\\
\pz\pz\pz + \textit{SFT}&53.93 & 65.22 & 29.67 &38.63 & 53.85 & 34.00 & 73.41 & 54.62 & 10.85 & 23.73 & 26.67& 22.00 & 31.45 & 75.00 & 1.00 & 38.57 & 75.00 & 4.33 & 43.29 & 58.39 & 16.98 \\
\pz\pz\pz + \textit{C-SFT (10)}&56.38 & 66.67&28.33&39.86&64.52&33.67&73.77&52.14&12.20&24.87&38.46&21.67&31.44&75.00&2.33&38.78&70.00&3.33&44.18&61.13&16.92\\
\pz\pz\pz + \textit{C-SFT (4)} &55.95&50.00&28.33&39.23&60.00&32.67&74.67&52.73&10.85&23.42&26.67&23.00&31.38&75.00&0.33&38.49&77.27&7.67&43.86&56.94&17.14 \\
    \bottomrule
  \end{tabular}
\end{adjustbox}
\vspace{-0.1in}
\caption{Curriculum learning at different epochs consistently leads to better self-improvement, and we consistently apply it at a later epoch (epoch 10).}
\label{tab:cl-epoch}
\end{table*}


\subsection{Improvement with External Reward}
\label{app:Improvement with External Reward}
We construct post-training data using external Qwen2.5-VL-72B-Instruct. For Janus-Pro-7B with the SFT strategy, \Cref{tab:external} compares Qwen-based alignment with self-improvement. Qwen enables Janus-Pro-7B to achieve better generation and unification. Self-improvement yields slightly weaker alignment, likely due to Janus-Pro-7B's inferior image understanding capability compared to Qwen. Nevertheless, without introducing any external signals, the self-improvement method achieves results close to those obtained with Qwen-based alignment.
\begin{table*}[h] 
\centering
\setlength{\tabcolsep}{2.8pt}     
\renewcommand{\arraystretch}{1.05}

\begin{adjustbox}{max width=\textwidth, max height=\textheight, keepaspectratio}
  \begin{tabular}{l ccc |ccc |ccc |ccc |ccc |ccc |ccc}
    \toprule
    \multirow{2}{*}{Model} & \multicolumn{3}{c}{Texture} & \multicolumn{3}{c}{Shape} & \multicolumn{3}{c}{Color} & \multicolumn{3}{c}{Spatial} & \multicolumn{3}{c}{Non-Spatial} & \multicolumn{3}{c}{Complex} & \multicolumn{3}{c}{Overall} \\
    \cmidrule(lr){2-4}\cmidrule(lr){5-7}\cmidrule(lr){8-10}\cmidrule(lr){11-13}\cmidrule(lr){14-16}\cmidrule(lr){17-19}\cmidrule(lr){20-22}
    & Gen.$\uparrow$ & Und.$\uparrow$ & Non.$\downarrow$
    & Gen.$\uparrow$ & Und.$\uparrow$ & Non.$\downarrow$
    & Gen.$\uparrow$ & Und.$\uparrow$ & Non.$\downarrow$
    & Gen.$\uparrow$ & Und.$\uparrow$ & Non.$\downarrow$
    & Gen.$\uparrow$ & Und.$\uparrow$ & Non.$\downarrow$
    & Gen.$\uparrow$ & Und.$\uparrow$ & Non.$\downarrow$
    & Gen.$\uparrow$ & Und.$\uparrow$ & Non.$\downarrow$ \\
    \midrule
\multicolumn{10}{l}{\textit{Gen. and Und.}}{\vspace{0.02in}}  \\
\pz\pz Janus-Pro-7B$_{\textit{\scriptsize (Baseline)}}$ & 38.63 &50.00  &43.33 &33.49& 50.00& 43.00&53.22 & 50.00 & 27.33& 16.81 & 50.00&  31.00&  31.40&   50.00 &  2.33& 37.73 &50.00 &10.33&35.21 &50.00 &26.22\\
\pz\pz\pz + \textit{Self-improved SFT}&53.93 & 65.22 & 29.67 &38.63 & 53.85 & 34.00 & 73.41 & 54.62 & 10.85 & 23.73 & 26.67& 22.00 & 31.45 & 75.00 & 1.00 & 38.57 & 75.00 & 4.33 & 43.29 & 58.39 & 16.98 \\
\pz\pz\pz + \textit{Qwen-assisted SFT}&56.84& 56.00& 25.00 &41.53& 59.26& 37.33 & 76.18&49.63 & 11.86 &24.14&31.25 & 19.33 & 31.48&70.00 & 1.00 & 38.53 & 66.67& 5.33 & 44.78&55.47 &16.64\\
    \bottomrule
  \end{tabular}
\end{adjustbox}
\vspace{-0.1in}
\caption{Constructing post-training samples with Qwen also enhances the generation, understanding, and unification of MLLMs. Without any external rewards, the self-improvement method yields slightly lower performance and unification than Qwen-based MLLMs.}
\label{tab:external}
\end{table*}

\end{document}